\documentclass[10pt,twocolumn,letterpaper]{article}

\usepackage[pagenumbers]{cvpr} 

\usepackage{float}
\usepackage{lipsum}
\usepackage{wasysym}
\usepackage{multirow}
\usepackage{placeins}
\usepackage[dvipsnames]{xcolor}
\usepackage{bm}
\usepackage{titletoc}

\usepackage[resetlabels,labeled]{multibib}

\usepackage[accsupp]{axessibility}

\newcommand{\normal}{N}
\newcommand{\mnormal}{\text{\normal}}
\newcommand{\sensor}{S}
\newcommand{\msensor}{\text{\sensor}}

\newcommand{\fake}{F}
\newcommand{\mfake}{\text{\fake}}

\newcommand{\rend}{3D}
\newcommand{\mrend}{\text{\rend}}

\newcommand{\lL}{\mathcal{L}}
\newcommand\norm[1]{\left\lVert#1\right\rVert_{2}^{2}}

\newcommand{\mns}{\text{\normal\sensor}}
\newcommand{\msn}{\text{\sensor\normal}}
\newcommand{\mnsn}{\text{\normal\sensor\normal}}
\newcommand{\msns}{\text{\sensor\normal\sensor}}

\DeclareMathOperator*{\argmin}{arg\,min}

\newcites{Sup}{References - Supplementary}

\definecolor{cvprblue}{rgb}{0.21,0.49,0.74}
\usepackage[pagebackref,breaklinks,colorlinks,citecolor=cvprblue]{hyperref}

\def\paperID{16314} 
\def\confName{CVPR}
\def\confYear{2025}

\title{Electromyography-Informed Facial Expression Reconstruction for Physiological-Based Synthesis and Analysis}   
\author{
Tim Büchner\textsuperscript{\rm 1}
$\quad$
Christoph Anders\textsuperscript{\rm 2}
$\quad$
Orlando Guntinas-Lichius\textsuperscript{\rm 2}
$\quad$
Joachim Denzler\textsuperscript{\rm 1}\\
{\normalsize \textsuperscript{\rm 1}Computer Vision Group $\quad$ \textsuperscript{\rm 2}Jena University Hospital} \\
{\normalsize Friedrich Schiller University Jena, Germany}\\
{\tt\small \{tim.buechner, joachim.denzler\}@uni-jena.de}\\
{\tt\small \{christoph.anders, orlando.guntinas\}@med.uni-jena.de}
}

\begin{document}
\twocolumn[{%
    \maketitle
    \renewcommand\twocolumn[1][]{#1}
    \begin{center}
        \vspace{-0.20in}
        \captionsetup{type=figure}
        \includegraphics[width=0.96\textwidth]{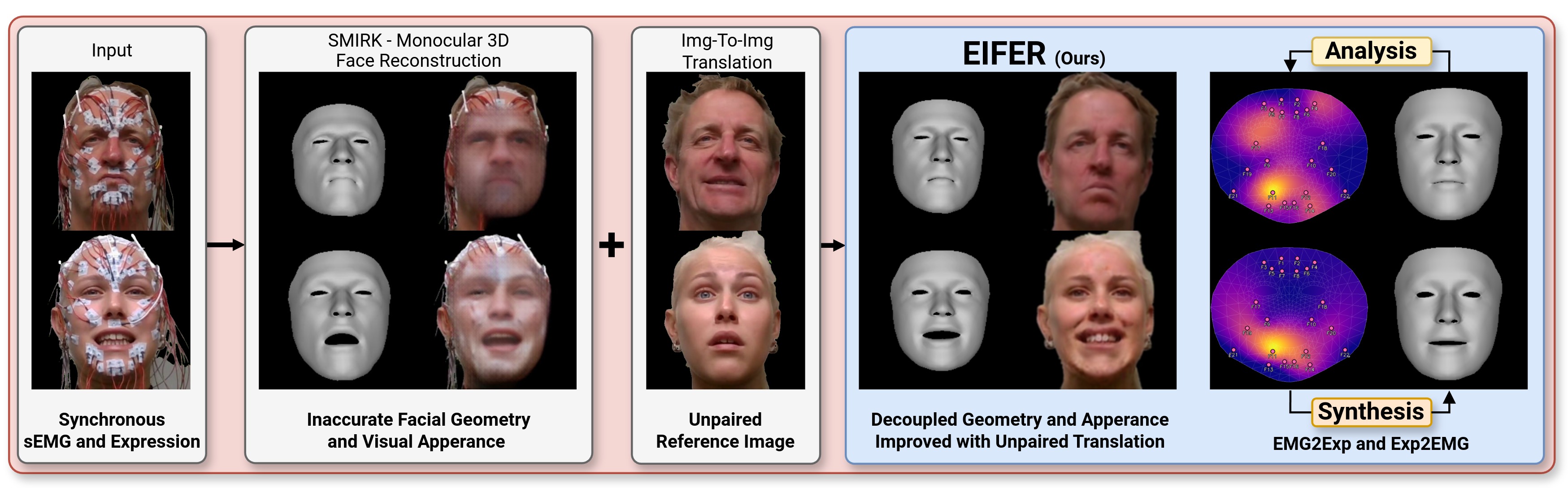}
        \caption{
            \textbf{Bridging the gap between mimics and muscles:}
            Our method \textbf{EIFER} utilizes neural unpaired image-to-image translation to decouple facial geometry and appearance for muscle-activity-based expression synthesis and electrode-free facial electromyography.
        }
        \label{fig:teaser}
    \end{center}
}]

\begin{abstract}
The relationship between muscle activity and resulting facial expressions is crucial for various fields, including psychology, medicine, and entertainment.
The synchronous recording of facial mimicry and muscular activity via surface electromyography (sEMG) provides a unique window into these complex dynamics.
Unfortunately, existing methods for facial analysis cannot handle electrode occlusion, rendering them ineffective.
Even with occlusion-free reference images of the same person, variations in expression intensity and execution are unmatchable.
Our electromyography-informed facial expression reconstruction (EIFER) approach is a novel method to restore faces under sEMG occlusion faithfully in an adversarial manner.
We decouple facial geometry and visual appearance (e.g., skin texture, lighting, electrodes) by combining a 3D Morphable Model (3DMM) with neural unpaired image-to-image translation via reference recordings.
Then, EIFER learns a bidirectional mapping between 3DMM expression parameters and muscle activity, establishing correspondence between the two domains. 
We validate the effectiveness of our approach through experiments on a dataset of synchronized sEMG recordings and facial mimicry, demonstrating faithful geometry and appearance reconstruction.
Further, we synthesize expressions based on muscle activity and how observed expressions can predict dynamic muscle activity.
Consequently, EIFER introduces a new paradigm for facial electromyography, which could be extended to other forms of multi-modal face recordings\footnote{Project page: \url{https://eifer-mam.github.io}}.
\end{abstract}    
\section{Introduction}
The relationship between muscle activity and facial expressions presents a complex challenge with significant implications for various application areas, including psychology~\cite{blasberg2023you, ekmanArgumentBasicEmotions1992, mao2023poster, elboudouriEmoNeXtAdaptedConvNeXt2023, psy1, psy2, psy3, psy4}, medicine~\cite{emg1, emg2, emg3, med1, med2, med3, med4, med5, med6, med7, funkWirelessHighresolutionSurface2024, schumann2010facial}, and animation~\cite{sculptor, mono3dsota, yangLearningGeneralizedPhysical2024, fermanFaceLiftSemisupervised3D2024, smirk, wagnerSoftDECAComputationallyEfficient2023, REALY, ruhlandDataDrivenApproachSynthesizing2017}.
Despite its importance, many questions about how muscle activity influences facial expressions and vice versa remain unanswered.
Synchronously recorded muscle activity via surface electromyography (sEMG) and facial expressions open a unique window into these dynamics.

The modeling of facial geometry, encompassing shape and expressions, with 3D Morphable Models (3DMMs) is a longstanding research area~\cite{3dmm, 3dmmpastpresentfuture, FLAME, bfm1, bfm2, ictfacekit, facescape1, facescape2, REALY}.
Combined with advances in monocular 3D face reconstruction~\cite{mono3dsota, deca, smirk, emoca, mostgan, sculptor}, 3DMMs are a solution to bridge the gap between mimics and muscles.
Thus, we could generate synthetic facial expressions from muscle activity and do electrode-free facial electromyography.

However, the electrodes affect the face reconstruction even for a current state-of-the-art monocular 3D face reconstruction method~\cite{smirk}, as seen in \Cref{fig:teaser}.
The reliance on occlusion-sensitive preprocessing and regularization terms~\cite{FLAMEidtMICA, unreasonableLPIPS, gecer2019ganfit, genovaUnsupervisedTraining3D2018, shang2020self, emoca, FOCUS} renders fine-tuning unfeasible for this data.
Further, the appearance models~\cite{bfm1, bfm2} used by these methods do not consider occlusion, which affects the photometric optimization~\cite{deca, emoca, FOCUS, deng2019accurate}.

\begin{figure*}[t]
    \centering
    \includegraphics[width=\textwidth]{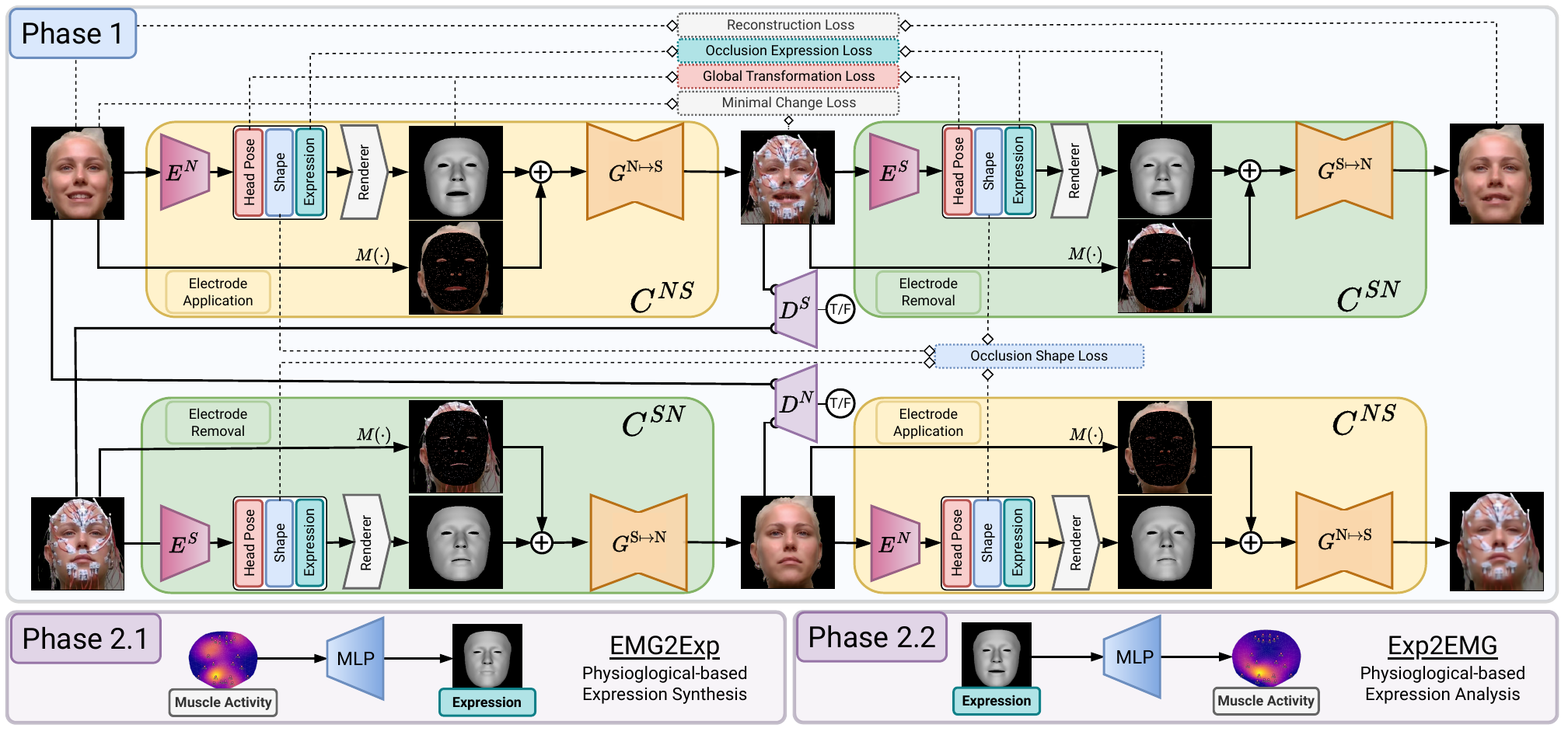}
    \caption{
        EIFER employs a double encoder-generator architecture in a CycleGAN-like framework~\cite{cyclegan} to reconstruct facial geometry and generate photorealistic appearances with artificially applied and removed sEMG electrodes during \textit{Phase One}.
        In \textit{Phase Two}, EIFER learns the bidirectional mapping between expressions and muscle activity, facilitating physiological-based synthesis and electrode-free facial electromyography.
        Full arrows denote information flow, while dashed arrows denote information flow by regularization terms.
    }
    \label{fig:model_architecture}
\end{figure*}

To overcome these limitations, we frame the appearance reconstruction as an adversarial problem~\cite{gecer2019ganfit, mostgan, buchner2023let}, leveraging unpaired occlusion-free reference images~\cite{cyclegan}.
With neural unpaired image-to-image translation via a CycleGAN structure~\cite{img2img, cyclegan}, we jointly train two models that place and remove sEMG electrodes from the appearance.
The double encoder-generator architecture is illustrated in \Cref{fig:model_architecture}.
This approach offers two key advantages:
\begin{itemize}
    \item \textbf{Adversarial Challenge:}
    We implicitly train the generators to produce photorealistic faces without requiring additional perceptual losses~\cite{vggloss,unreasonableLPIPS}.
    To avoid hallucinations, we employ a minimal change regularization~\cite{buchner2023improved} and multi-stage training such that the entire system converges to a meaningful solution~\cite{multistageAU}.
    \item \textbf{Cycle Consistency:}
    We use widely adopted pre-trained encoder networks to estimate the 3DMM parameters~\cite{smirk}.
    In a cycle, one encoder handles electrode-free faces while the other encoder replicates the same output for the electrode-covered version of the face.
    Knowing that a person's shape and expression should not change during a cycle, we constrain both encoders against each other. 
    We achieve this by replacing occlusion-sensitive regularization terms~\cite{FLAMEidtMICA, FLAME, deng2019accurate, gecer2019ganfit, genovaUnsupervisedTraining3D2018, shang2020self, emoca, smirk} with cycle-consistent self-supervised alternatives.
\end{itemize}
Our setup trains the encoders to handle sEMG occlusions, developing an electromyography-informed facial expression reconstruction (EIFER) approach.
Our model then learns the non-linear relationship between 3DMM expression space and measured muscle activity.

We evaluate EIFER's performance using synchronized sEMG recordings and facial mimicry, demonstrating its ability to handle occlusion and accurately reconstruct facial expressions, as shown in \Cref{fig:teaser}.
We also investigate the bidirectional mapping's capabilities for synthesis via muscle activity and facial electromyography from videos.

In summary, our contributions are:
1) A method for facial reconstruction under substantial sEMG occlusion.
2) Reframing the analysis-by-synthesis appearance reconstruction as an adversarial unpaired image-to-image translation task.
3) Building a correspondence between 3DMMs and muscle activity for physiological-based expression synthesis and electrode-free facial electromyography.

\section{Related Work}
3D Morphable Models (3DMMs) are widely used for facial expression analysis and synthesis~\cite{3dmm, 3dmmpastpresentfuture}.
Many 3DMMs~\cite{FLAME, bfm1, bfm2, facescape1, facescape2, ictfacekit, faceWarehouse} aim to split the face into shape and expression components~\cite{eggerIdentityExpressionAmbiguity3D2021, weihererApproximatingIntersectionsDifferences2024}.
However, the expressions are driven by the underlying facial muscle activity~\cite{emg1, emg2, emg3, med1, ekmanFacialActionCoding1978_paper, schumann2010facial, Medina_2022_CVPR}.
This link is not explicitly modeled despite this relationship, resulting in a gap.

Several methods attempt to close this gap either via the Facial Action Coding System~\cite{ekmanFacialActionCoding1978_paper, kuangAUAwareDynamic3D2024, multistageAU, OpenFace, shaoJAANetJointFacial2021}, physics simulations~\cite{mazza3DMechanicalModeling2011, ichimPhacePhysicsbasedFace2017, hungFrameworkGeneratingAnatomically2015} or geometry modeling~\cite{haberFaceFaceReal2001, kimFaceGPSComprehensiveTechnique2024, kahlerGeometrybasedMuscleModeling2001, wagnerSoftDECAComputationallyEfficient2023, sifakisAutomaticDeterminationFacial2005}.
Unfortunately, none of these compare with recorded sEMG muscle signals.
Therefore, we simultaneously capture mimicry and muscular activity.
This introduces occlusions during analysis, which we have to handle.

Modern 3DMM parameter estimators utilize monocular 3D face reconstruction~\cite{mono3dsota, FOCUS, mostgan, deca, emoca, smirk, shang2020self, changExpNetLandmarkFreeDeep2018, lugaresiMediaPipeFrameworkBuilding2019, guoFastAccurateStable2021, kimInverseFaceNetDeepMonocular2018, zhangAccurate3DFace2023, deng2019accurate, sculptor, baileyFastDeepFacial2020, FLAMEidtMICA, buechner2024facing}.
However, occlusions lead to inaccurate photometric reconstruction, as shown in \Cref{fig:teaser}.
This is likely due to the underlying analysis-by-synthesis assumptions, such as the used appearance model~\cite{bfm1, bfm2, ictfacekit, facescape1, faceWarehouse}.
They may not assume occlusions and, thus, diverge~\cite{ deng2019accurate, deca, emoca, smirk, guoFastAccurateStable2021, zhangAccurate3DFace2023}.

To solve this, recent approaches use implicit neural rendering to replace the appearance model~\cite{smirk, mostgan, doukasHead2HeadDeepFacial2021, doukasHeadGANOneshotNeural2021, buchner2023let, buchner2023improved}.
This offers flexibility to adapt to unseen occlusions and replaces traditional rendering techniques~\cite{oswald3drendering, deca, emoca, FOCUS, deng2019accurate, zhangAccurate3DFace2023}, promoting robust reconstruction~\cite{choiStarGANUnifiedGenerative2018, mostgan, ictfacekit, buchner2023let, buchner2023improved}.
Therefore, we apply neural rendering to handle the electrode occlusion.
Similarly to~\cite{smirk}, our renderer uses facial geometry and sparse color information from the input image.
Instead of minimizing the photometric difference between the input and reconstruction~\cite{deca, emoca, FOCUS, facescape1, faceWarehouse, ictfacekit, deng2019accurate, filntisisSPECTREVisualSpeechInformed2023, FLAMEidtMICA}, a discriminator distinguished between the generated and an unpaired occlusion-free reference image~\cite{goodfellow2014generative}.

During adversarial analysis-by-synthesis, a renderer removes electrodes and generates realistic faces~\cite{buchner2023let, buchner2023improved, mostgan, img2img, cyclegan, choiStarGANUnifiedGenerative2018, ictfacekit, instancenorm}.
However, this model might compensate for incorrect expressions, similar to~\cite{smirk, buchner2023improved}.
Therefore, we employ separate models for electrode removal and placing~\cite{cyclegan, buchner2023let}.
We introduce occlusion-robust cycle-consistent constraints to constrain both models~\cite{smirk, img2img, cyclegan, buchner2023improved}.

Lastly, we use expression parameters to solve downstream tasks~\cite{emoca, smirk, filntisisSPECTREVisualSpeechInformed2023}, specifically synthesizing expressions from muscular activity.
The inverse direction leads to a form of electrode-free facial electromyography.
Both directions integrate physiological information into 3DMMs, overcoming the existing gap towards muscular activity.

\section{Method: Electromyography-Informed Facial Expression Reconstruction (EIFER)}
EIFER follows analysis-by-synthesis reconstruction methods~\cite{deca, emoca, smirk, mostgan, filntisis2022visual, sculptor, FOCUS, shang2020self, zhangAccurate3DFace2023, ictfacekit, deng2019accurate} to decouple facial geometry and visual appearance to reconstruct expression under surface electromyography (sEMG) occlusion faithfully.
We replace occlusion-sensitive regularization terms used in photometric reconstruction by focusing on cycle-consistent self-supervision in a CycleGAN-like structure~\cite{cyclegan}.
Thus framing the appearance reconstruction as an adversarial problem, our design implicitly ensures realistic face generation without requiring additional training losses, such as perceptual~\cite{vggloss, unreasonableLPIPS, FLAME, deng2019accurate, gecer2019ganfit, genovaUnsupervisedTraining3D2018, shang2020self}, identity~\cite{FLAMEidtMICA}, or emotion losses~\cite{emoca}.
With synchronized sEMG and expressions, we then learn a bidirectional mapping, equipping the 3DMM with physiological capabilities.

\subsection{Model Architecture}
EIFER employs two encoder-generator pairs, illustrated in~\Cref{fig:model_architecture}.
One pair places the electrodes during a cycle while the other learns to remove them.
To ensure consistency, we constrain both pairs against each other.
We denote all variables with a superscript describing occlusion-free faces as (\textbf{\normal})ormal and with electrodes as (\textbf{\sensor})ensor.
\newline
\textbf{Face Embedding:}
We build upon the widely adopted FLAME face model~\cite{FLAME}, a mesh consisting of $n=5023$ vertices modifiable by shape $\beta$ and expression $\varphi$ parameters, with two additional blendshape parameters for eyelid closure $\varphi_{\text{eye}}$~\cite{FLAMEidtMICA}.
The model accounts for jaw movement $\varphi_{\text{jaw}}$ and head pose $\theta_{\text{pose}}$, representing rigid head transformations.
We predict the muscle activity via the expression triplet ($\varphi$, $\varphi_{\text{eye}}$, $\varphi_{\text{jaw}}$).
\newline
\textbf{Encoder Network:}
We use the recent SMIRK approach~\cite{smirk} by constructing an encoder model that consists of three sub-encoders, each utilizing a MobileNetV3 backbone~\cite{mobilenet3}, to process an input image $I$.
These sub-encoders are responsible for predicting the parameters for shape ($E_\beta$), expression (including eyelids and jaw pose, $E_\varphi$), and global transformations including camera position ($E_\theta$).
Then, we compute a monochrome render of the face ($I_\mrend$) using a differential mesh renderer to represent the facial geometry~\cite{ravi2020pytorch3d}, see \Cref{fig:model_architecture}.
\newline
\textbf{Generator Network:}
Advances in neural rendering~\cite{smirk, buchner2023let, buchner2023improved, shang2020self, gecer2019ganfit} led to new methods that replace traditional rendering techniques~\cite{deca,emoca,bfm1,bfm2}.
Particularly, image-to-image translation networks $G(\cdot)$~\cite{smirk} learn to combine rendered facial geometry with sparse color information from the original input face, as shown in \Cref{fig:model_architecture}.

The masking function $M(\cdot)$, introduced by~\cite{smirk}, samples these pixels within the 2D facial landmark hull, visualized in the $C^{\mns}$ module in \Cref{fig:model_architecture}.
To circumvent inaccurate landmarks on sEMG occluded faces, we use the rendered geometry as the sampling area.
We rely on the pre-trained encoder networks for alignment.
Notably, our problem is more complex than simple photometric reconstruction~\cite{deca, emoca, smirk, FOCUS, mono3dsota, liuDenseFaceAlignment2017}, as the generator must learn to ignore or produce pixels related to sEMG electrodes only via adversarial feedback.
We use a ResNet network as the backbone model~\cite{resnet, cyclegan, buchner2023improved, buchner2023let}.
This ensures a similar gradient flow as in~\cite{smirk, cyclegan}.
Our generator uses instance normalization to produce fine details~\cite{cyclegan,instancenorm}.
\newline
\textbf{Face Discriminators:}
We employ two neural discriminators~\cite{img2img, goodfellow2014generative}, $D^{\text{\normal}}(I^{\mnormal}, I^{\mnormal}_{\mfake})$ and $D^{\text{\sensor}}(I^{\msensor}, I^{\msensor}_{\mfake})$, to distinguish real and fake faces for the applied and removed sEMG electrodes, respectively.
This design implicitly enforces the generation of realistic faces and removes the additional need for perceptual losses to improve the visual quality~\cite{vggloss,unreasonableLPIPS}.
\newline
\textbf{Expression-Electromyography-Estimators:}
We use a six-layer MLP with ReLU activations~\cite{relupaper} to learn the mapping between facial mimicry and muscular activity and vice versa.
The final layers are \texttt{Tanh} for \textit{EMG2Exp} and a \texttt{ReLU}~\cite{relupaper} for \textit{Exp2EMG}, respectively, to accommodate non-negative sEMG values.
Only the 3DMM expression parameters are used, excluding identity information.

\subsection{Cycle-Consistent Self-Supervision}
Monocular 3D face reconstruction often relies on the occlusion-sensitive facial features to guide the reconstruction path~\cite{FLAMEidtMICA, unreasonableLPIPS, gecer2019ganfit, genovaUnsupervisedTraining3D2018, shang2020self, emoca, deca, smirk, mostgan, mono3dsota}.
This is compromised under sEMG occlusion, where the extraction is affected.
Moreover, when the encoder predicts incorrect expressions, the generator likely compensates for this error, which still leads to a solution to the adversarial problem~\cite{buchner2023improved, smirk, deca, FOCUS, mostgan, goodfellow2014generative, img2img}.

Therefore, we propose cycle-consistent self-supervision regularization terms insensitive to occlusion leveraging reference images during training.
Nonetheless, the expression extraction is still ill-posed.
Thus, we leverage a multi-stage training approach to guide the system to meaningful convergence.
For brevity, we denote the consecutive execution of an encoder $E^{\mnormal}(\cdot)$ generator $G^{\mnormal \rightarrow \msensor}(\cdot)$ pair as $C^{\mns}(\cdot)$, and cycle is $C^{\mnsn}(\cdot)$.
The same holds for the $C^{\msns}(\cdot)$ cycle.
\newline
\textbf{Cycle Consistency:}
To ensure correct reconstructions, we self-supervise both cycles with the following losses:

\textit{Reconstruction Loss}: After a full cycle, the reconstructed image shall resemble the input image:
\begin{equation*}
\lL_{\text{Reco}} = \norm{C^{\mnsn}(I^{\mnormal}) - I^{\mnormal}} + \norm{C^{\msns}(I^{\msensor}) - I^{\msensor}}~.
\end{equation*}

\textit{Identity Loss:}
To ensure that the encoder-generator module preserves the identity of an opposing input image, we introduce an identity loss~\cite{cyclegan, taigmanUnsupervisedCrossDomainImage2016, taigmanDeepfaceClosingGap2014}, $\lL_{\text{Idt}}$.
This loss minimizes changes when the image should not be altered:
\begin{equation*}
\lL_{\text{Idt}} = \norm{C^{\mns}(I^{\msensor}) - I^{\msensor}} + \norm{C^{\msn}(I^{\mnormal}) - I^{\mnormal}}~.
\end{equation*}
This prompts the generator to learn the sensor locations and stops the encoder from relying on default expressions~\cite{cyclegan}.

\textit{Minimal Change Loss:}
We introduce a minimal change loss term, $\lL_{\text{MC}}$~\cite{buchner2023improved}, to prevent the generators from introducing unwanted features or hallucinations:
\begin{equation*}
\lL_{\text{MC}} = {\norm{C^{\mns}(I^{\mnormal}) - I^{\mnormal}}} + \norm{C^{\msn}(I^{\msensor}) - I^{\msensor}}~.
\end{equation*}
\newline
\textbf{Face Geometry Consistency}
Relying solely on reconstruction losses does not guarantee that the encoder $E^{\msensor}$ entangles shape, expression, and pose parameters, impacting the bidirectional mapping between expression and muscle activity.
Nonetheless, the facial geometry should remain consistent during a full reconstruction cycle of artificial electrode removal and placing.
Hence, $E^{\msensor}$ should mirror $E^{\mnormal}$ behavior, but on the electrode-occluded face.
We enforce this behavior by freezing $E^{\mnormal}$ throughout training.
We assume sufficient pre-training and already regularized prediction capabilities to prevent extreme and unrealistic expressions~\cite{smirk}.

\textit{Occlusion Expression Loss:}
The expression encoder $E^\mnormal_{\varphi}$ and $E^{\msensor}_{\varphi}$ should predict the same for faces where the electrodes have been artificially placed or removed:
\begin{equation*}
\lL_{\text{OccExp}} 
= \norm{E^\mnormal_{\varphi}(I^{\mnormal}) - E^\msensor_{\varphi}(I^{\msensor}_\mfake)} 
+ \norm{E^\msensor_{\varphi}(I^{\msensor}) - E^\mnormal_{\varphi}(I^{\mnormal}_\mfake)}~.
\end{equation*}

\textit{Occlusion Shape Loss:}
The shown facial expression between \textit{unpaired} training images of a person might differ, but the shape should not~\cite{smirk,  3dmm, 3dmmpastpresentfuture, eggerIdentityExpressionAmbiguity3D2021, weihererApproximatingIntersectionsDifferences2024}.
Thus, we ensure shape consistency throughout both cycles:
\begin{equation*}
\lL_{\text{OccShp}} 
= \norm{E^\mnormal_{\beta}(I^{\mnormal}) - E^\msensor_{\beta}(I^{\msensor})} 
+ \norm{E^\msensor_{\beta}(I^{\msensor}_\mfake) - E^\mnormal_{\beta}(I^{\mnormal}_\mfake)}~.
\end{equation*}

\textit{Global Transformation Loss:}
By modifying the sampling area to the masking function $M(\cdot)$ based on the rendered face $I_{\mrend}$, we risk that the pose encoders offset the FLAME model such that the entire image plane is used for sampling.
Hence, the generator would compensate by coupling facial geometry and appearance again.  
To prevent this, we introduce an additional constraint on the rigid transformation of $E^{\msensor}_{\varphi}$, specifically on camera position, head pose, and barycentric landmarks defined on the FLAME model~\cite{deca, emoca, smirk, FLAME, FLAMEidtMICA}, within a full $C^{\mnsn}$ cycle:
\begin{equation*}
\lL_{\text{Lmk}} 
= \sum^{K}_{i=1}
    \norm{
        \mathbf{k}_{I^{\mnormal}_\mrend} -
        \mathbf{k}_{I^{\msensor}_{\mfake,\mrend}}
    }
+ 
\norm{
        \mathbf{k}_{I^{\msensor}_\mrend} -
        \mathbf{k}_{I^{\mnormal}_{\mfake,\mrend}},
    }
\text{and }
\end{equation*}
\begin{equation*}
\lL_{\text{RigTra}} 
= \norm{E^\mnormal_{\theta}(I^{\mnormal}) - E^\msensor_{\theta}(I^{\msensor}_\mfake)} 
+ \norm{E^\msensor_{\theta}(I^{\msensor}) - E^\mnormal_{\theta}(I^{\mnormal}_\mfake)}~.
\end{equation*}
\newline
\textbf{Full Objective:}
We define a joined loss function, $\lL_{\text{GEN}}$, which combines all \textbf{Cycle Reconstruction} and all \textbf{Face Geometry Consistency} loss terms.
Adopting a CycleGAN structure, we employ adversarial training to implicitly create highly realistic faces.
Specifically, we view the discriminators as classifiers (a real ($1$) and fake ($0$) classification), encouraging samples at the decision boundary using the \textit{least squares generative adversarial loss}~\cite{lsgan} $\lL_{\text{GAN}}$ as
\begin{align*}
\lL_{\text{GAN}}^{\msensor}(C^{\mns}, D^{\msensor}, I^{\msensor}, I^{\mnormal}) =
    &\frac{1}{2}\norm{D^{\msensor}(I^\msensor) - 1} +\\
    &\frac{1}{2}\norm{D^{\msensor}(C^{\msn}(I^\mnormal)) - 0}~,
\end{align*}
and same for $\lL^{\mnormal}_{\text{GAN}}$.
The generative parts of the model (encoder and generator jointly) and the discriminators train in alternating passes~\cite{cyclegan}, solving:
\begin{equation*}
    \bar{C}^{\mns}, \bar{C}^{\msn} = \argmin_{C^{\mns}, C^{\msn}} ~\max_{D^{\mnormal}, D^{\msensor}} \lL(C^{\mns}, C^{\msn}, D^{\mnormal}, D^{\msensor})~,
\end{equation*}
with
\begin{equation*}
\lL(C^{\mns}, C^{\msn}, D^{\mnormal}, D^{\msensor}) = \lL_{\text{GEN}} + \lL^{\mnormal}_{\text{GAN}} + \lL^{\msensor}_{\text{GAN}}~.
\end{equation*}
We weigh the loss terms during training with the following lambdas, obtained by a hyperparameter search: $\lambda_{\text{Reco}}=10,\lambda_{\text{Idt}}=1.5, \lambda_{\text{MC}}=0.5, \lambda_{\text{OccExp}}=1.0,\lambda_{\text{OccShp}}=0.1,\lambda_{\text{Lmk}}=2.5,\lambda_{\text{RigTra}}=0.1$

\subsection{Training EIFER - Phase 1}
Although the regularization terms prevent the encoders from collapsing into unrelated expressions, the task remains ill-posed.
To address this, we employ a multi-stage training procedure guiding each component to learn its correct task and ensuring the entire system converges to a solution.
\newline
\textbf{Stage 1 - sEMG Application and Removal:}
We first train the generators to produce realistic faces.
By freezing the encoder weights, the generator focuses on learning the relationship between geometry and pixels, retaining 50\% of pixels.
Although this approach temporarily disregards expression, the generators learn the electrode locations and solve the adversarial problem.
As a result, the generators develop the ability to produce realistic faces, a crucial foundation for the subsequent stages.
\newline
\textbf{Stage 2 - Estimating Occluded Expressions:}
The encoder $E^{\msensor}$ is unfrozen to adapt to electrode occlusion.
Further, we reduce the pixel amount to 10\%, guiding the generator towards the rendered facial geometry to reconstruct the face.
\newline
\textbf{Stage 3 - Final Decoupling}
We retain only 1\% of pixels, forcing the generator to rely on the rendered face geometry.

\subsection{Training EIFER - Phase 2}
The instance normalization in the generators requires a batch size of one~\cite{instancenorm}.
However, neither of the MLPs would converge during training.
This enforces a two-phase approach, shown in \Cref{fig:model_architecture}: training the encoders to handle sEMG occlusions and then learning the bidirectional mapping between expression parameters and muscle activity.

\section{Dataset - Mimics And Muscles}
\begin{figure}[t]
    \centering
    \includegraphics[width=\linewidth]{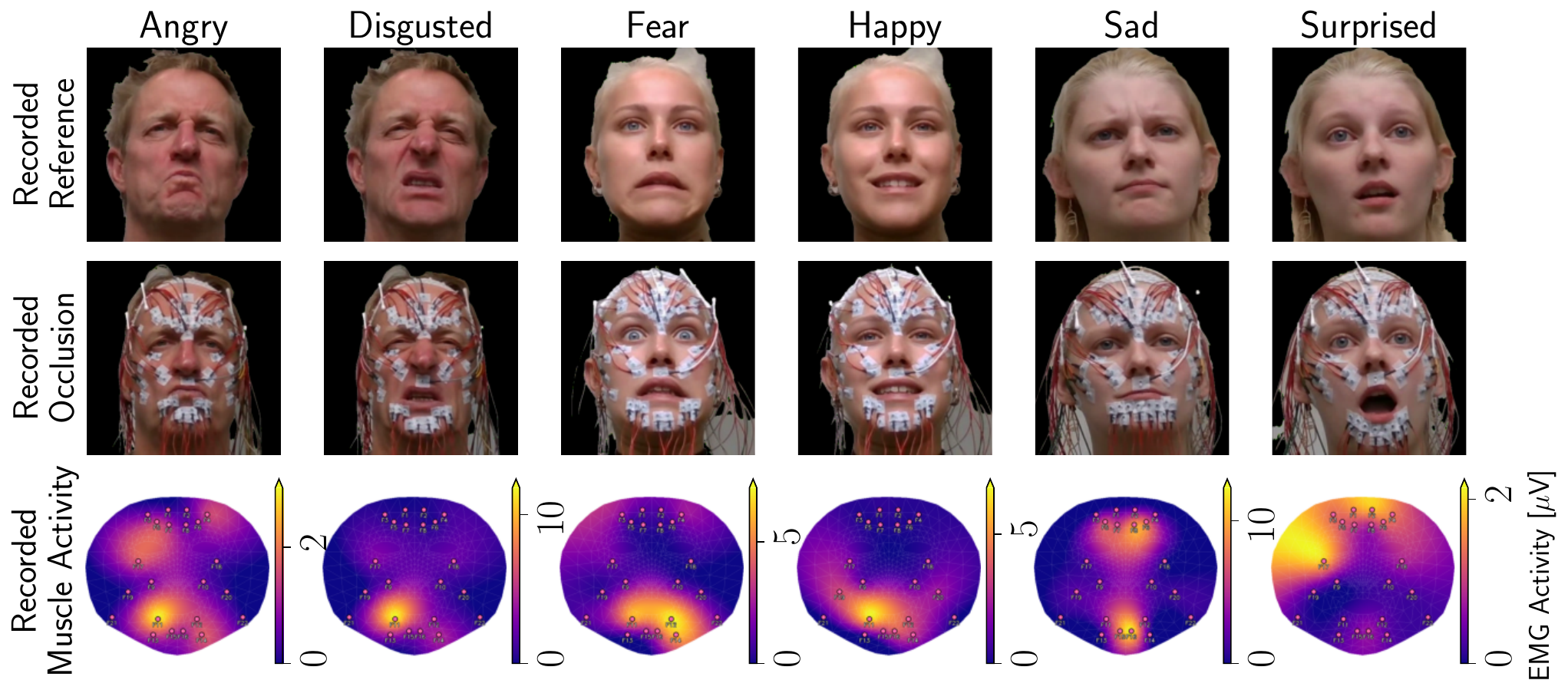}
    \caption{
        Three individuals mimic six basic emotions~\cite{ekmanArgumentBasicEmotions1992}, with synchronized sEMG heat-maps illustrating muscular activity~\cite{buchner2023using}.
        These images showcase varying expression intensities and executions, emphasizing the need for a robust sEMG occlusion method.
    }
    \label{fig:dataset}
\end{figure}

We recorded 36 participants (19 \venus, 17 \mars, age range: 18-67 years) without a history of any neurological disease to obtain synchronous facial expression and muscle activity.
Only beardless men were recruited to attach the surface electromyography (sEMG) electrodes to the face.
Participants performed a series of eleven facial functional movements and six emotional expressions repeated four times following an instruction video~\cite{schaede,ekmanArgumentBasicEmotions1992}.
Each individual was recorded three times per recording session (two weeks apart), twice with EMG and once without sEMG as reference.
Samples of three individuals are depicted in \Cref{fig:dataset}.

We used the Fridlund sEMG scheme~\cite{fridlund} to obtain muscle activation captured with a sampling rate of 4096/s~\cite{emg1,emg2,emg3}.
The facial movements were captured using a frontal-facing camera at 30 FPS.
The recordings are synchronized and processed according to established standards~\cite{emg1, emg2, emg3, xiaEMGBasedEstimationLimb2018, tsinganosDeepLearningEMGbased2018, phinyomarkEMGPatternRecognition2018, fridlund, kuramoto, med1, MODNet, zheng2022avatar}.
We extracted the downsampled linear envelope of the muscle activity signal, which is used to learn the correspondence to the 3DMM expression space.
Details on electrode locations, data preprocessing, and data set statistics can be found in the supplementary material.
\section{Experiments and Results}
We compare EIFER to recent state-of-the-art monocular 3D face reconstruction techniques, including DECA~\cite{deca}, EMOCAv2~\cite{emoca}, and SMIRK~\cite{smirk}, which utilize the FLAME model~\cite{FLAME}, and Deep3DFace~\cite{deng2019accurate} and FOCUS~\cite{FOCUS}, which employ the BFM model~\cite{bfm1,bfm2}.
We also compare with MC-CycleGAN~\cite{buchner2023let, buchner2023improved}, which does not rely on a face model.
To ensure a fair comparison, we train and fine-tune all models on a common 10\% frame subset of the reference recordings of the individuals, as shown in \Cref{fig:dataset}.
Hence, the models learn the individuals' characteristics, and we assume that any deviations in behavior are due to the occlusion.
Additional results, including ablation studies, training hyperparameters, and visualizations, are provided in the supplementary material.

\begin{figure}[t]
    \centering
    \includegraphics[width=\linewidth]{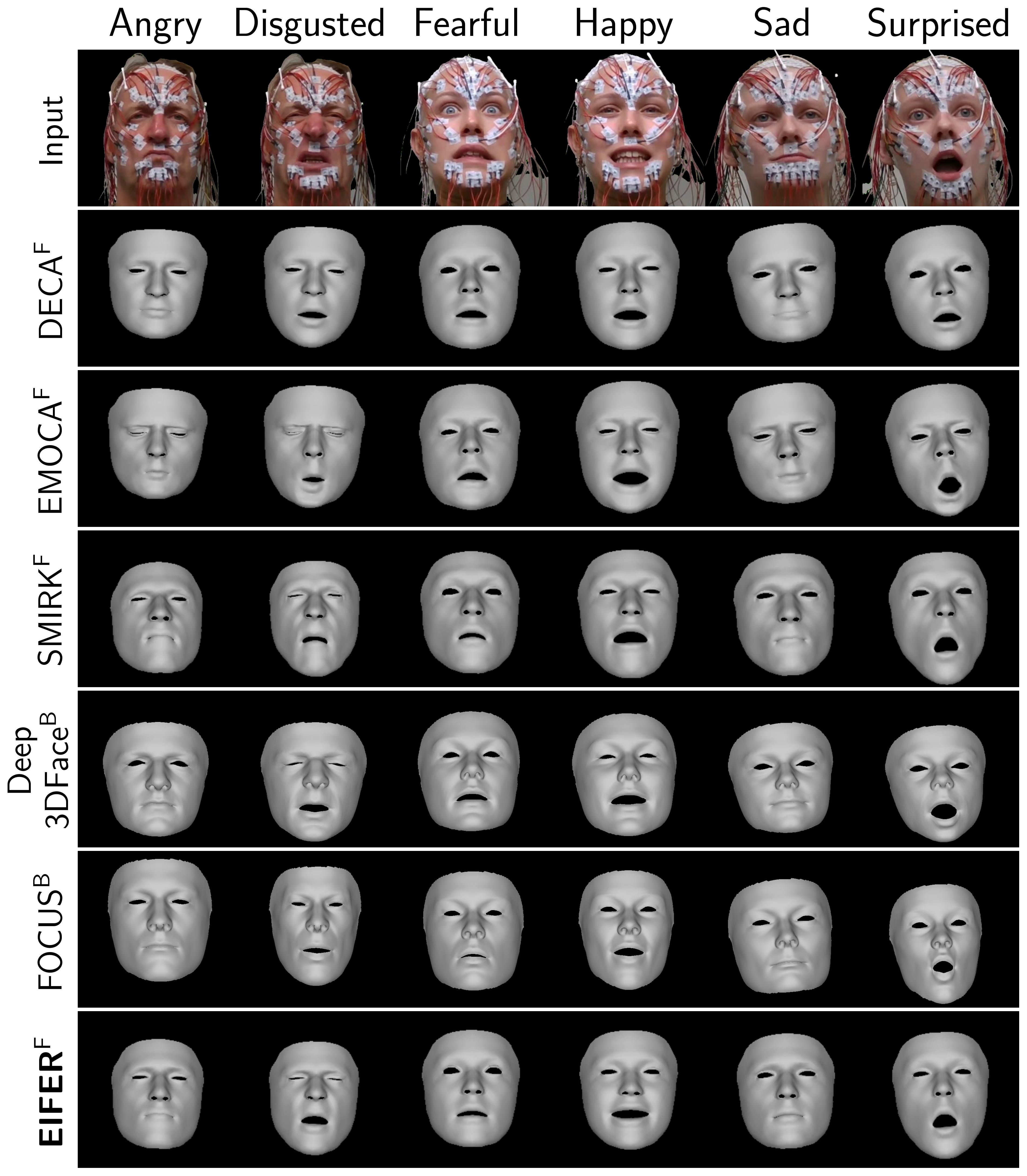}
    \caption{
        \textbf{Facial Geometry:}
        We visualize the estimated face geometry under sEMG occlusion for three individuals mimicking expressions.
        MC-CycleGAN has no face model and is thus omitted.
    }
    \label{fig:qualitative_shape}
\end{figure}

\subsection{Reconstruction Evaluation}
The reconstruction quality, using the occlusion-free recordings as a baseline, is evaluated on two factors: facial geometry and visual appearance of the restored faces.
\newline
\textbf{Qualitative:}
We visualize the estimated facial geometry in \Cref{fig:qualitative_shape}.
While most methods capture the general expression, they vary in intensity and alignment, as seen for \textit{happy}.
EIFER correctly maintains the shape parameters for the same individual with different expressions.
This indicates that other methods' face encoders struggle to disentangle shape and expression, which is crucial for learning expression-muscular activity mapping.

\begin{figure}[t]
    \centering
    \includegraphics[width=\linewidth]{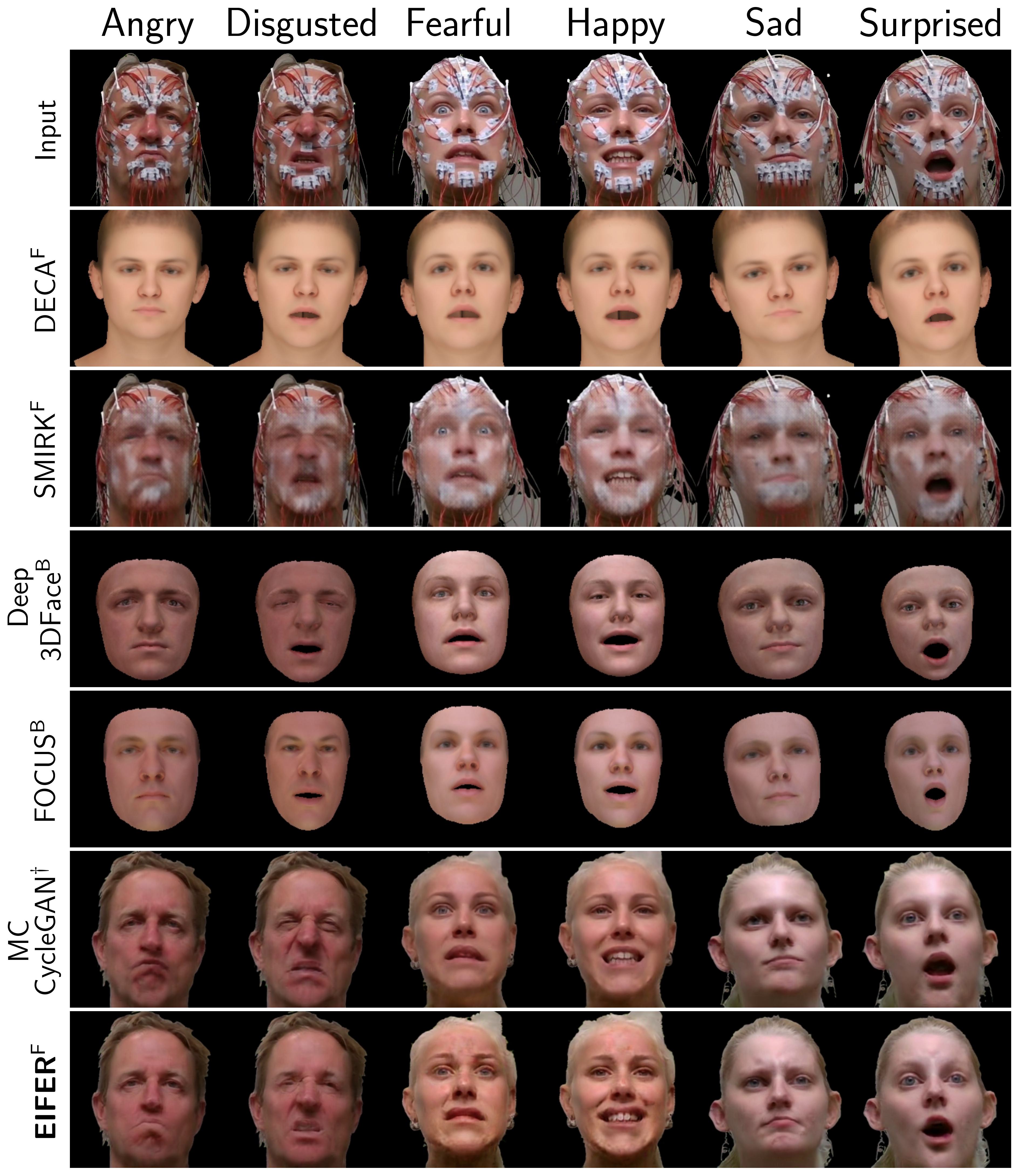}
    \caption{
        \textbf{Appearance reconstruction and electrode removal:}
        Among the state-of-the-art methods, only SMIRK fails the reconstruction.
        MC-CycleGAN and EIFER keep the nuanced features.
    }
    \label{fig:qualitative_recons}
\end{figure}

We further investigate the visual reconstructions, shown in \Cref{fig:qualitative_recons}.
Most existing methods inherently remove electrodes due to the used appearance model~\cite{bfm1, bfm2}.
However, only SMIRK, which forwards visual features to the neural generator, retains them.
Despite the same information flow, EIFER ignores artifacts by solving the adversarial task.

Deep3DFace achieves consistent results in extracting facial texture under occlusion across various expressions for the same individual.
In contrast, FOCUS, which relies on the same appearance model~\cite{bfm1, bfm2}, fails this task, indicating that electrodes significantly impact its performance.

MC-CycleGAN and EIFER, both trained adversarially, create photorealistic reconstructions.
EIFER achieves similar results to MC-CycleGAN with only 1\% of the pixel information, albeit with slightly degraded quality around the mouth.
Thus, EIFER combines the strengths of SMIRKs' encoding and MC-CycleGAN reconstruction.
\begin{figure*}[t]
    \centering
    \includegraphics[width=0.96\linewidth]{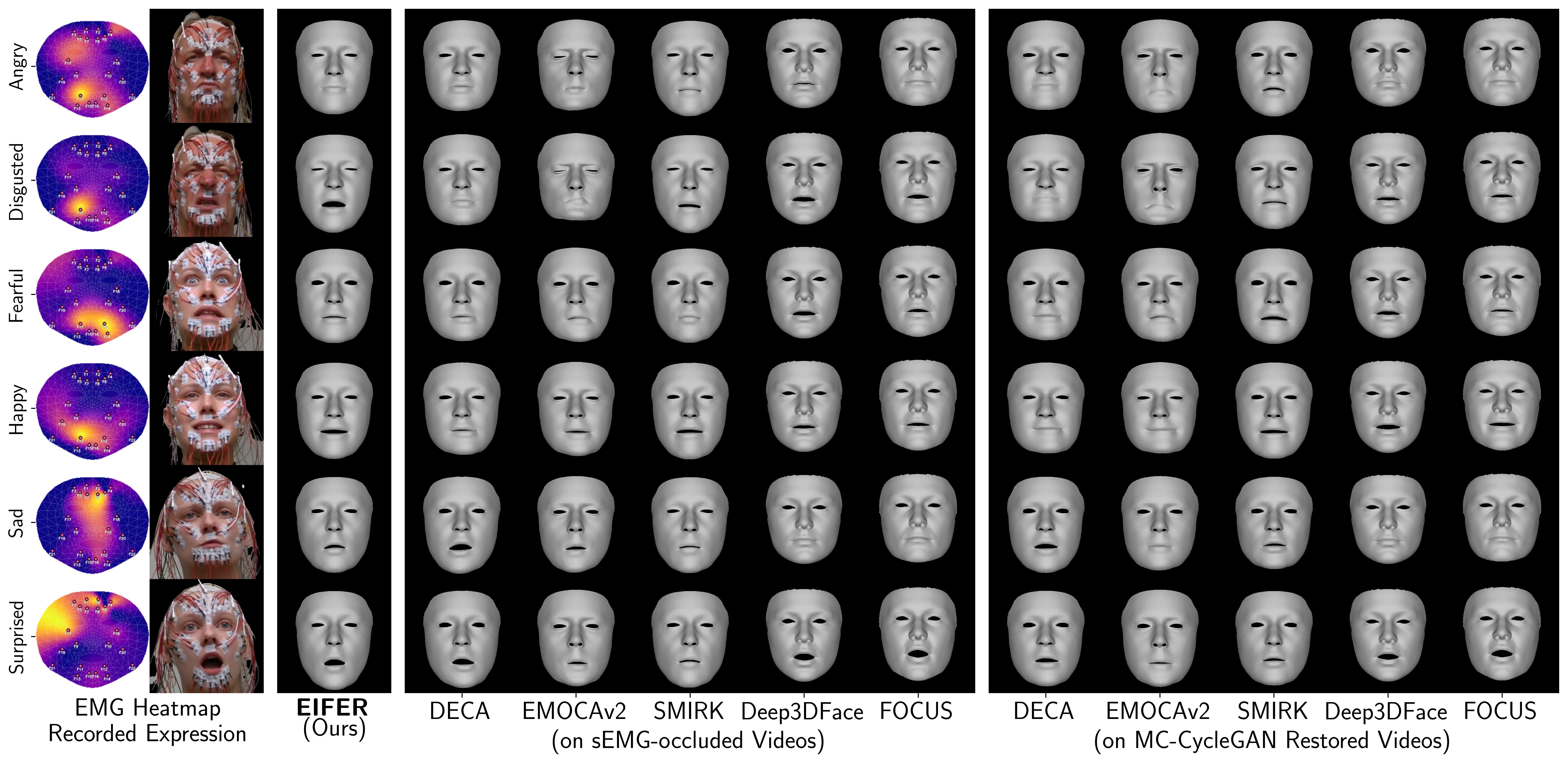}
    \caption{
        \textbf{Comparison of Synthesized Facial Expressions:}
        Using all expression encoder models, we synthesize various expressions (on a shape-free face) estimated from recorded muscle activity.
        For a fair comparison, we evaluate all on the electrode-occluded and the by MC-CycleGAN~\cite{buchner2023let, buchner2023improved} restored recordings.
        EIFER achieves comparable performance on occluded recordings, whereas other methods struggle to produce accurate results even on occlusion-free faces.
        Video reconstructions are provided in the supplementary material.
    }
    \label{fig:emg2exp}
\end{figure*}
\begin{table}[t]
	\centering
	\resizebox{\columnwidth}{!}{%
		\begin{tabular}{l|cccc|r}
			\toprule
			Method                                 & SSIM ($\uparrow$)                         & GMSD ($\downarrow$)                       & PSNR ($\uparrow$)                          & MDSI ($\downarrow$)                       & FID ($\downarrow$)                          \\
			\midrule
			Baseline N-N                           & 0.86{\footnotesize $\pm$0.07}             & 0.12{\footnotesize $\pm$0.04}             & 27.95{\footnotesize $\pm$3.71}             & 0.34{\footnotesize $\pm$0.04}             & 7.41{\footnotesize $\pm$ 3.72}              \\
			Baseline N-S                           & 0.39{\footnotesize $\pm$0.05}             & 0.33{\footnotesize $\pm$0.01}             & 13.69{\footnotesize $\pm$1.27}             & 0.62{\footnotesize $\pm$0.02}             & 285.42{\footnotesize $\pm$38.18}            \\
			\midrule
			DECA\textsuperscript{F}                & 0.53{\footnotesize $\pm$0.04}             & 0.29{\footnotesize $\pm$0.01}             & 12.43{\footnotesize $\pm$0.65}             & 0.46{\footnotesize $\pm$0.01}             & 165.24{\footnotesize $\pm$31.80}            \\
			SMIRK\textsuperscript{F}               & 0.47{\footnotesize $\pm$0.06}             & 0.31{\footnotesize $\pm$0.02}             & 14.45{\footnotesize $\pm$1.41}             & 0.58{\footnotesize $\pm$0.02}             & 275.80{\footnotesize $\pm$46.38}            \\
			Deep3DFace\textsuperscript{B}          & 0.48{\footnotesize $\pm$0.05}             & 0.31{\footnotesize $\pm$0.01}             & 14.42{\footnotesize $\pm$1.39}             & 0.58{\footnotesize $\pm$0.03}             & 219.28{\footnotesize $\pm$43.29}            \\
			FOCUS\textsuperscript{B}               & 0.46{\footnotesize $\pm$0.05}             & 0.32{\footnotesize $\pm$0.02}             & 13.95{\footnotesize $\pm$1.35}             & 0.58{\footnotesize $\pm$0.03}             & 227.71{\footnotesize $\pm$50.21}            \\
			MC-CycleGAN\textsuperscript{$\dagger$} & 0.66{\footnotesize $\pm$0.08}             & \textbf{0.24{\footnotesize $\pm$0.03}} & \textbf{19.38{\footnotesize $\pm$2.39}} & 0.45{\footnotesize $\pm$0.02}             & 54.39{\footnotesize $\pm$24.32}             \\
			\midrule
			EIFER\textsuperscript{F}               & \textbf{0.66{\footnotesize $\pm$0.09}} & \textbf{0.24{\footnotesize $\pm$0.03}} & 19.42{\footnotesize $\pm$2.57}             & \textbf{0.44{\footnotesize $\pm$0.03}} & \textbf{52.56{\footnotesize $\pm$27.75}} \\
			\bottomrule
		\end{tabular}%
	}
	\caption{
		\textbf{Reconstruction quality}:
		We evaluate the reference recordings with the reconstructions, with upper and lower limits established by comparing reference recordings to themselves and sEMG-occluded recordings.
		We \textbf{mark} the best performance per metric and denote the underlying 3DMM face models with \textsuperscript{B} for BFM~\cite{bfm1,bfm2}, \textsuperscript{F} for FLAME~\cite{FLAME}, and \textsuperscript{$\dagger$} for no model.
	}
	\label{tab:qualitative}
\end{table}

\newline
\textbf{Quantitative:}
We report the appearance reconstruction with the occlusion-free baselines in \Cref{tab:qualitative}.
We use the metrics: Structure Similarity Index (SSIM)~\cite{ssim}, Gradient Magnitude Similarity Deviation (GMSD)~\cite{xue2013gradient}, Peak Signal-To-Noise Ratio (PSNR), Mean Deviation Similarity Index (MDSI)~\cite{nafchi2016mean}, and Frechet Inception Distance (FID)~\cite{heusel2017gans}.

We observe that all methods exceed the lower limit set by occluded recordings.
Notably, state-of-the-art methods perform similarly to MC-CycleGAN and EIFER for similarity measures.
Regardless of traditional or neural rendering, MC-CycleGAN and EIFER outperform simple photometric reconstruction in the FID metric.
This highlights the strength of adversarial training for face generation.
More downstream tasks for an objective evaluation, such as expression classification, are provided in the supplementary.

\subsection{Synthesis of Expressions - EMG2Exp}
We investigate synthesizing facial expressions from muscular activity, establishing a correspondence between muscle activity and 3D Morphable Model (3DMM) expression parameters.
To evaluate the effectiveness of EIFER, we additionally train the same MLPs architecture, \textit{EMG2Exp}, with expression extracted from state-of-the-art (SOTA) methods on two datasets: the sEMG occluded recordings and MC-CycleGAN restored recordings, see \Cref{fig:qualitative_recons}.
The latter ensures a fair comparison of the generalized SOTA model performance such that the underlying 3DMMs are comparable. 
We employ the same optimization strategy as EIFER's second training phase.
However, due to differences in expression space ranges between FLAME~\cite{FLAME, emoca, smirk} and BFM~\cite{bfm1, bfm2, FOCUS, deng2019accurate}, the final loss term is not directly comparable.
Given the challenges of defining a clear ground truth for facial expressions, we use visual evaluation to assess our method's performance.
We show the synthesized expressions on a shape-free face in \Cref{fig:emg2exp}.

EIFER accurately synthesizes facial expressions from muscle activity, matching the recording frame.
In contrast, SOTA models are significantly impacted by sEMG occlusion, leading to inaccurate expression synthesis.
This is likely due to the inherent inaccuracies in extracted expressions, which limits the range of synthesis.
As expected, restored visual appearance tests yield better results, but DECA and EMOCAv2 still struggle to synthesize correct expressions.
Notably, only SMIRK's occlusion-free variant accurately synthesizes the \textit{fearful} expression, while SMIRK, Deep3DFace, and FOCUS produce visually similar results to EIFER.
However, EIFER works directly with sEMG occluded faces, eliminating the removal step.

Our analysis reveals current limitations in synthesizing expressions based on muscle activity, particularly concerning expression intensity.
The \textit{masseter} muscle's minimal activity during mouth-opening poses a challenge, as it requires only slight activation to maintain an open mouth~\cite{emg1,emg3}.
This limitation affects the synthesis of \textit{disgusted} and \textit{surprised} expressions.
Additionally, voluntary eye blinking is not fully captured by any method, as it is not accounted for by the \textit{Fridlund} sEMG schematic~\cite{fridlund}.
Video examples illustrating these limitations are provided in the supplementary material.
This highlights the importance of individual differences in muscle activation patterns in future research.

We present the potential of synthesizing expressions from muscular activity for future research in physiological-based animations and sEMG-based facial motion capture.

\subsection{Analysis of Muscular Activity - Exp2EMG}
Learning the mapping from facial expression to muscle activity proposes electrode-free facial electromyography.
For a fair comparison, we train the same EIFER MLP architecture on the expression parameters extracted by the other state-of-the-art methods.
We follow the same training protocol for \textit{EMG2Exp} and compared two datasets: sEMG occluded and MC-CycleGAN restored recordings.

Each model's performance is evaluated in \Cref{tab:exp2emg} based on five-fold cross-validation.
We observe that all models learn the envelope shape (measured by Spearmans $\rho$ ~\cite{spearmanProofMeasurementAssociation1904}) but differ in predictive strength (measured by mean squared error, root mean square, and symmetric mean absolute percentage error~\cite{armstrongLongrangeForecastingCrystal1985}).
Building on the base SMIRK encoder model, which already achieves high performance, EIFER effectively handles sEMG occlusion without requiring prior electrode removal by MC-CycleGAN.
Both BFM and FLAME face models perform similarly for the restored recordings, except EMOCAv2~\cite{emoca}, which exhibits difficulties in mapping likely due to overestimated expression parameters.
This suggests that the trained facial encoder is more crucial than the employed face model.
\begin{table}[t]
	\centering
	\resizebox{\columnwidth}{!}{%
		\begin{tabular}{ll|rrrc}
			\toprule
			 & Model                         & MSE ($\downarrow$)                          & RMS ($\downarrow$)                         & SMAPE ($\downarrow$)                        & Spearman $\rho$ ($\uparrow$)              \\
			\midrule
			\multirow{5}{*}{\rotatebox[origin=c]{90}{Occluded}}
			 & DECA\textsuperscript{F}       & 10.86{\footnotesize $\pm$75.44}             & 1.75{\footnotesize $\pm$ 2.79}             & 70.65{\footnotesize $\pm$36.58}             & 0.37{\footnotesize $\pm$0.19}             \\
			 & EMOCAv2\textsuperscript{F}    & 22.13{\footnotesize $\pm$81.00}             & 3.23{\footnotesize $\pm$ 4.96}             & 77.52{\footnotesize $\pm$26.48}             & 0.34{\footnotesize $\pm$0.18}             \\
			 & SMIRK\textsuperscript{F}      & 6.79{\footnotesize $\pm$59.74}              & 1.62{\footnotesize $\pm$ 2.04}             & 75.15{\footnotesize $\pm$34.81}             & 0.36{\footnotesize $\pm$0.19}             \\
			 & Deep3DFace\textsuperscript{B} & 12.34{\footnotesize $\pm$74.08}             & 5.86{\footnotesize $\pm$22.88}             & 72.82{\footnotesize $\pm$37.52}             & 0.35{\footnotesize $\pm$0.18}             \\
			 & FOCUS\textsuperscript{B}      & 14.20{\footnotesize $\pm$75.37}             & 5.94{\footnotesize $\pm$22.47}             & 77.21{\footnotesize $\pm$36.14}             & 0.34{\footnotesize $\pm$0.18}             \\
			\midrule
			\multirow{5}{*}{\rotatebox[origin=c]{90}{Restored}}
			 & DECA\textsuperscript{F}       & 6.52{\footnotesize $\pm$ 59.82}             & 1.58{\footnotesize $\pm$ 2.82}             & 70.99{\footnotesize $\pm$35.20}             & 0.37{\footnotesize $\pm$0.20}             \\
			 & EMOCAv2\textsuperscript{F}    & 28.82{\footnotesize $\pm$121.55}            & 3.96{\footnotesize $\pm$ 8.25}             & 82.67{\footnotesize $\pm$32.30}             & 0.34{\footnotesize $\pm$0.18}             \\
			 & SMIRK\textsuperscript{F}      & 6.44{\footnotesize $\pm$ 30.37}             & 1.69{\footnotesize $\pm$ 1.90}             & 74.07{\footnotesize $\pm$36.66}             & 0.36{\footnotesize $\pm$0.19}             \\
			 & Deep3DFace\textsuperscript{B} & 8.36{\footnotesize $\pm$ 68.11}             & 1.64{\footnotesize $\pm$ 2.38}             & 69.17{\footnotesize $\pm$33.92}             & 0.36{\footnotesize $\pm$0.19}             \\
			 & FOCUS\textsuperscript{B}      & 9.73{\footnotesize $\pm$ 80.42}             & 5.70{\footnotesize $\pm$23.15}             & 75.22{\footnotesize $\pm$37.33}             & 0.35{\footnotesize $\pm$0.19}             \\
			\midrule
			 & EIFER\textsuperscript{F}      & \textbf{5.09{\footnotesize $\pm$ 33.97}} & \textbf{1.45{\footnotesize $\pm$ 1.73}} & \textbf{67.71{\footnotesize $\pm$32.54}} & \textbf{0.38{\footnotesize $\pm$0.19}} \\
			\bottomrule
		\end{tabular}%
	}
	\caption{
		\textbf{Muscle Activity Prediction Metrics.}
		We measure the difference between the linear envelopes of ground truth and the sEMG prediction over a five-fold cross-validation.
		We denote the face models with \textsuperscript{B} for BFM~\cite{bfm1,bfm2} and \textsuperscript{F} for FLAME~\cite{FLAME}.
	}
	\label{tab:exp2emg}
\end{table}

\begin{figure}[t]
    \centering
    \includegraphics[width=\linewidth]{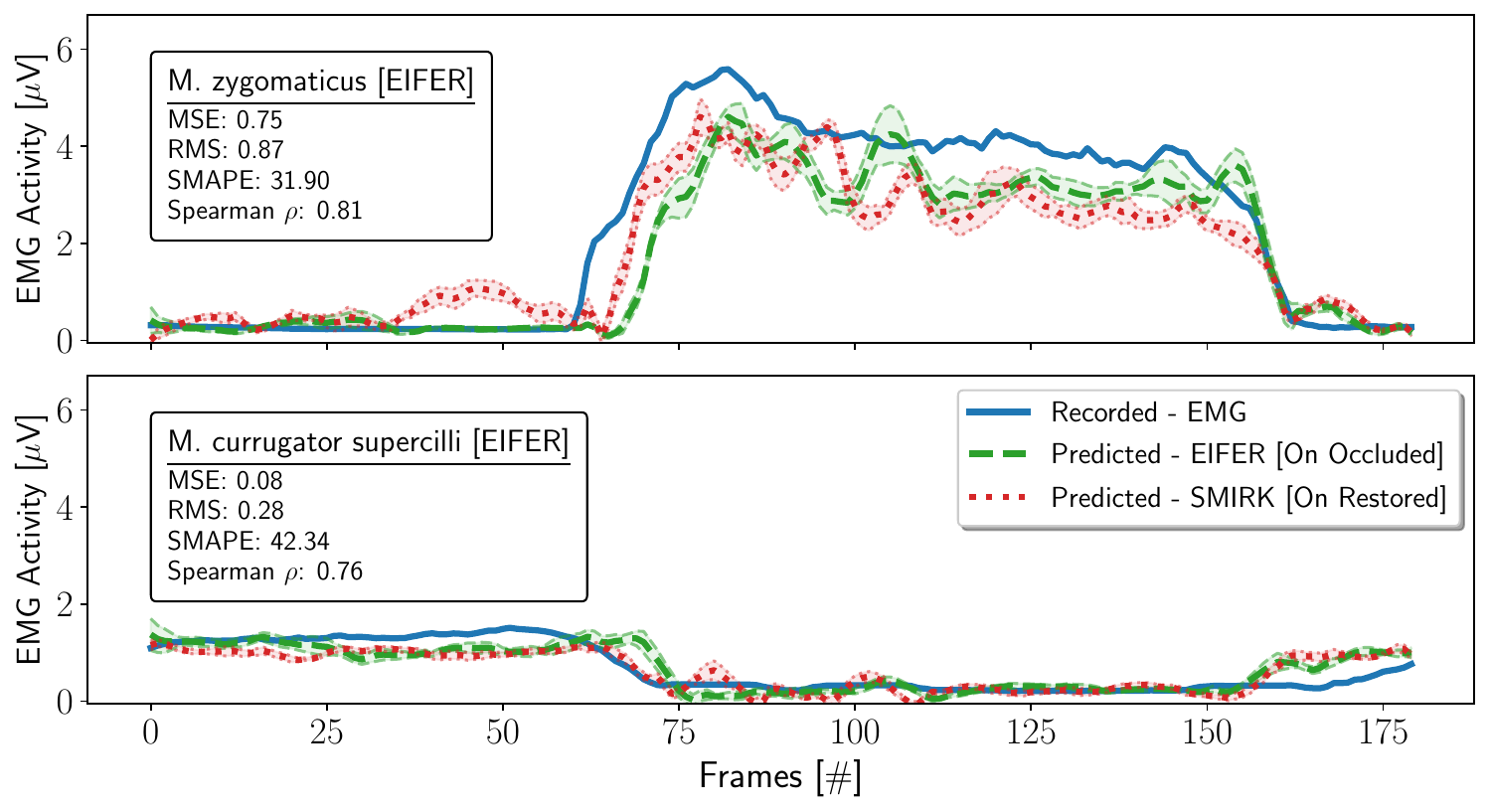}
    \caption{
        \textbf{Estimated Muscle Activity:}
        We show \textit{M. zygomaticus} and \textit{M. corrugator supercilii} during the \textit{happy} expression.
        Both models restore the temporal aspect of muscle activity, though EIFER achieves this on occluded recordings.
    }
    \label{fig:exp2emg}
\end{figure}

We highlight two muscles, one active (\textit{M. zygomaticus}) and one inactive (\textit{M. corrugator supercilii}), during the \textit{happy} expression in \Cref{fig:exp2emg}.
We focus on EIFER (on occluded recordings) and SMIRK (on restored recordings) since both have shown promising results for both geometry estimation (see \Cref{fig:qualitative_shape}) and expression synthesis (see \Cref{fig:emg2exp}).
Both models underestimate the amplitude and fail to capture the initial activity surge, likely due to the electromechanical delay between muscle activity and visible movement~\cite{xiaEMGBasedEstimationLimb2018,med1}. 
However, we rule out synchronization issues as the offset is handled correctly.
Static snapshots of the muscle activity in the form of topological heatmaps are provided in the supplementary material.
Our results demonstrate that the 3DMM expression space of FLAME~\cite{FLAME} and BFM~\cite{bfm1,bfm2} link to muscle activity and vice versa.

\section{Social Impact and Limitations}
EIFER presents a new paradigm for facial electromyography, estimating muscle signals from expressions in a data-driven manner.
An in-depth comparison with methods, such as the Facial Action Coding System~\cite{ekmanFacialActionCoding1978_paper}, remains open and is outside this study's scope.
Our results' generalizability is uncertain due to the small sample size ($N=36$).
Thus, we provide both \textit{Exp2EMG} and \textit{EMG2Exp} models for FLAME~\cite{FLAME} and BFM~\cite{bfm1, bfm2} to encourage further scrutiny.
We rely upon both 3DMMs disentanglement of shape and expression and the face encoder capabilities for establishing this correspondence~\cite{eggerIdentityExpressionAmbiguity3D2021, weihererApproximatingIntersectionsDifferences2024}.
For a more detailed discussion, please see the supplementary material.

\section{Conclusion}
Our work presents a novel approach to physiological-based face synthesis and analysis, electromyography-informed facial expression reconstruction (EIFER), which addresses the challenge of occlusion-sensitive regulation terms by leveraging unpaired references. 
Our findings have significant implications for multi-modal facial analysis, and we believe that the concepts of EIFER can be extended to other occlusion forms.
We demonstrate that EIFER handles occlusion through extensive experiments and faithfully reconstructs facial geometry with nuanced visual reconstruction.
EIFER estimates robust 3DMM parameters, synthesizes expressions, and predicts muscle activity using dynamic facial expressions. 
With these promising results, we plan to explore temporal aspects and muscle regularization during EIFER training with bootstrapping.
\newline
{\footnotesize
\textbf{Ethics Approval}
All participants gave written consent.
The Jena University Hospital ethics committee approved the study (No. 2019-1539).
\newline
\textbf{Acknowledgement}
Funded by Deutsche Forschungsgemeinschaft (DFG - German Research Foundation) project 427899908 BRIDGING THE GAP: MIMICS AND MUSCLES (DE 735/15-1 and GU 463/12-1).
}

{
    \small
    \bibliographystyle{ieeenat_fullname}
    \bibliography{main}
}

\clearpage
\maketitlesupplementary
\setcounter{section}{0}
\renewcommand{\thesection}{\Alph{section}}
\renewcommand\thesubsection{\thesection.\arabic{subsection}}
\section*{Table of Contents}
\startcontents
\printcontents{ }{1}{}

\section{Implementation and Model Details}
We provide an overview of the model architectures and experimental setups used in EIFER to facilitate re-implementation. This, combined with the publicly available source code\footnote{Project page: \url{https://eifer-mam.github.io}}, allows for a deeper understanding of EIFER's inner workings and suggests that the model architecture has a minor impact on the overall training pipeline.

EIFER is composed of three primary model components, which are duplicated for both the $C^\msn$ and $C^\mns$ cycles. Notably, during the evaluation of EMG2Exp and Exp2EMG, the $C^\msn$ cycle plays a crucial role. However, it is essential to recognize that the $C^\msn$ cycle cannot be trained in isolation from the other component, as the two cycles are interconnected and interdependent.

All models are implemented in PyTorch~\protect\citeSup{pytorch_supRef}, and we utilize PyTorch3D~\protect\citeSup{ravi2020pytorch3d_supRef} for rendering the FLAME~\protect\citeSup{FLAME_supRef} mesh to disentangle facial geometry from appearance.

\subsection{Encoder and Face Model}
We adopt the triple encoder structure from SMIRK~\protect\citeSup{smirk_supRef} and utilize MobileNetV3 as the backbone network.
This allows us to initialize EIFER with pre-trained SMIRK models, providing several benefits.

\begin{figure}[h]
    \centering
    \includegraphics[width=\linewidth]{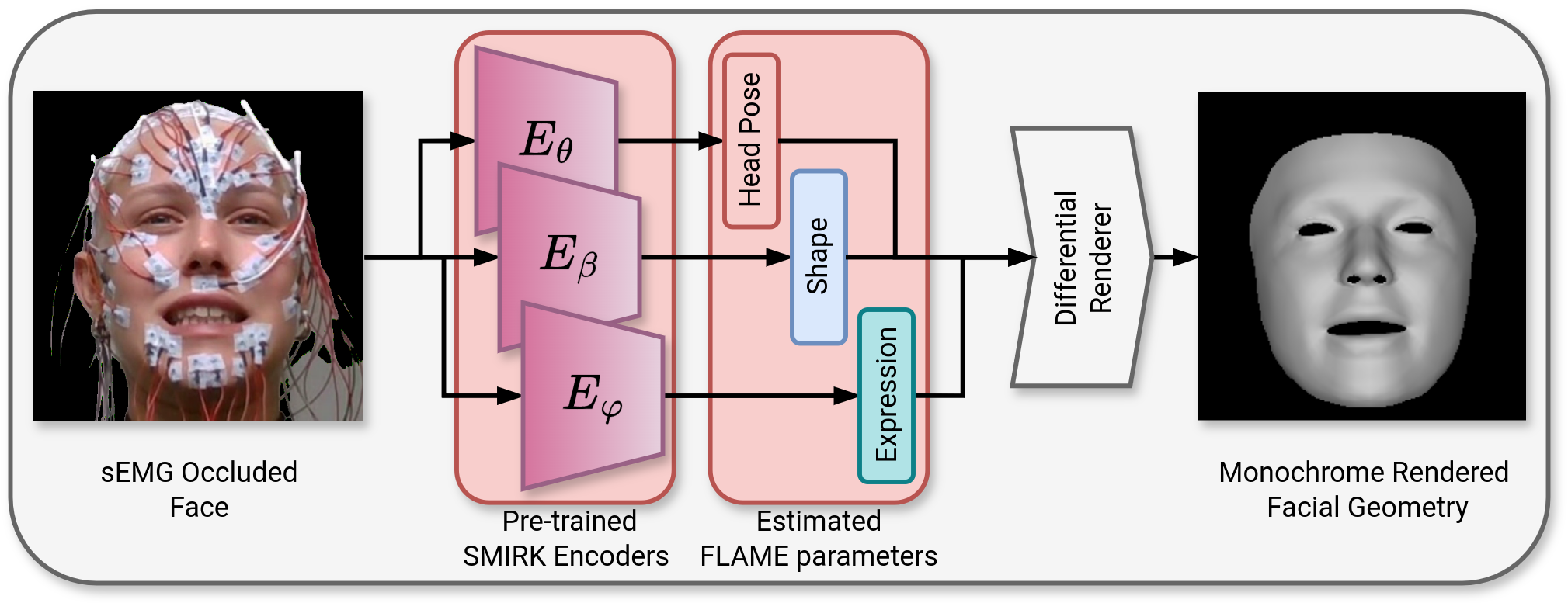}
    \caption{
       \textbf{EIFER Encoder Architecture}
       We utilize the tripe encoder setup of SMIRK~\protect\citeSup{smirk_supRef} to predict the FLAME parameters~\protect\citeSup{FLAME_supRef}.
       Therefore, each sub-encoder can be used independently of the given task.
       In our case, EIFER updates the pre-trained models to handle sEMG occlusion.
    }
    \label{fig:sup:encoder}
\end{figure}

Firstly, the pre-trained models are assumed to be robust to rough alignment without facial landmarks, as demonstrated in the ablation studies of ~\protect\citeSup{smirk_supRef}.
Secondly, we assume accurate facial feature extraction for non-sEMG occluded faces, enabling the other encoder to mimic the correct one under occlusion.
Lastly, this initialization ensures comparability with existing SMIRK results, as the updated model parameters are robust to sEMG occlusion.
The model architecture is illustrated in \Cref{fig:sup:encoder}.

We employ the intermediate FLAME~\protect\citeSup{FLAME_supRef} face representation, comprising $300$ shape and $50$ expression components to utilize pre-trained weights.
Additionally, we include three components for jaw movement and two blendshapes for the eyelids ~\protect\citeSup{FLAMEidtMICA_supRef}. The position sub-encoder estimates the head rotation and position, modeled by camera parameters.

We use these shape and expression parameters to construct the 3D FLAME mesh. 
A differential renderer~\protect\citeSup{ravi2020pytorch3d_supRef} then generates a monochromatic render of the frontal face view.
This rendered face contains essential facial geometry information following the same denomination as in ~\protect\citeSup{smirk_supRef, FOCUS_supRef}.
The generator model must restore the face correctly from this rendered face.

\subsection{Generator Model}
The generator model aims to reconstruct the input face faithfully.
Unlike traditional rendering approaches~\protect\citeSup{deca_supRef, emoca_supRef, FOCUS_supRef, deng2019accurate_supRef}, we employ implicit neural rendering~\protect\citeSup{smirk_supRef, buchner2023let_supRef, buchner2023improved_supRef} for its robustness.

To disentangle facial geometry and appearance, we use image-to-image translation techniques.
However, the input face image contains both geometry and appearance information.
To address this, we use the rendered face image, computed by the encoder networks, as the primary driver. 
Additionally, we forward random pixel information from the input face to the generator, similar to ~\protect\citeSup{smirk_supRef}, to recreate skin texture and lighting conditions.

The generator models take geometry and random appearance pixels as input, effectively functioning as an image-to-image translation network or style transfer model.
Unlike traditional rendering approaches that rely on an appearance model~\protect\citeSup{bfm1_supRef, bfm2_supRef, FOCUS_supRef, deng2019accurate_supRef}, we are not constrained by explicit assumptions, allowing us to adapt the generator models to our specific requirements.

To train the generator to ignore sEMG electrodes, we employ an unpaired reference image with a different expression and a discriminator.
This setup has two benefits: (1) the model learns to ignore pixels describing sEMG electrodes, and (2) the generated faces must be photorealistic to convince the discriminator, eliminating the need for additional perceptual losses.

However, this adversarial problem poses challenges, such as generative models creating incorrect features or hallucinating wrong expressions.
We refer the reader to the main paper for details on regularization terms that address these issues.

Unlike recent works~\protect\citeSup{smirk_supRef, img2img_supRef, cyclegan_supRef}, we use a ResNet~\protect\citeSup{resnet_supRef} as the backbone architecture for our generator models.
Although this differs from the typical Unet architecture, it allows for a similar gradient flow.

We modify the architecture to replace Conv2DTranspose layers with a single Upsample and Conv2D layer, eliminating the pixelated output and checkerboard patterns in SMIRK (see visualization in the main paper).
This improves the overall quality of the generated images.

We employ instance normalization as the primary activation function throughout the network~\protect\citeSup{instancenorm_supRef}, which enhances the reconstruction quality and information flow in the optimization problem.
However, instance normalization requires a batch size of one to avoid mirroring the behavior of standard batch normalization~\protect\citeSup{instancenorm_supRef}.

We adopt the multi-phase approach outlined in the main paper to address this limitation, as parallel-trained models like EMG2Exp cannot converge with small batch size. This approach ensures stable training and convergence.

\begin{figure}[h]
    \centering
    \includegraphics[width=\linewidth]{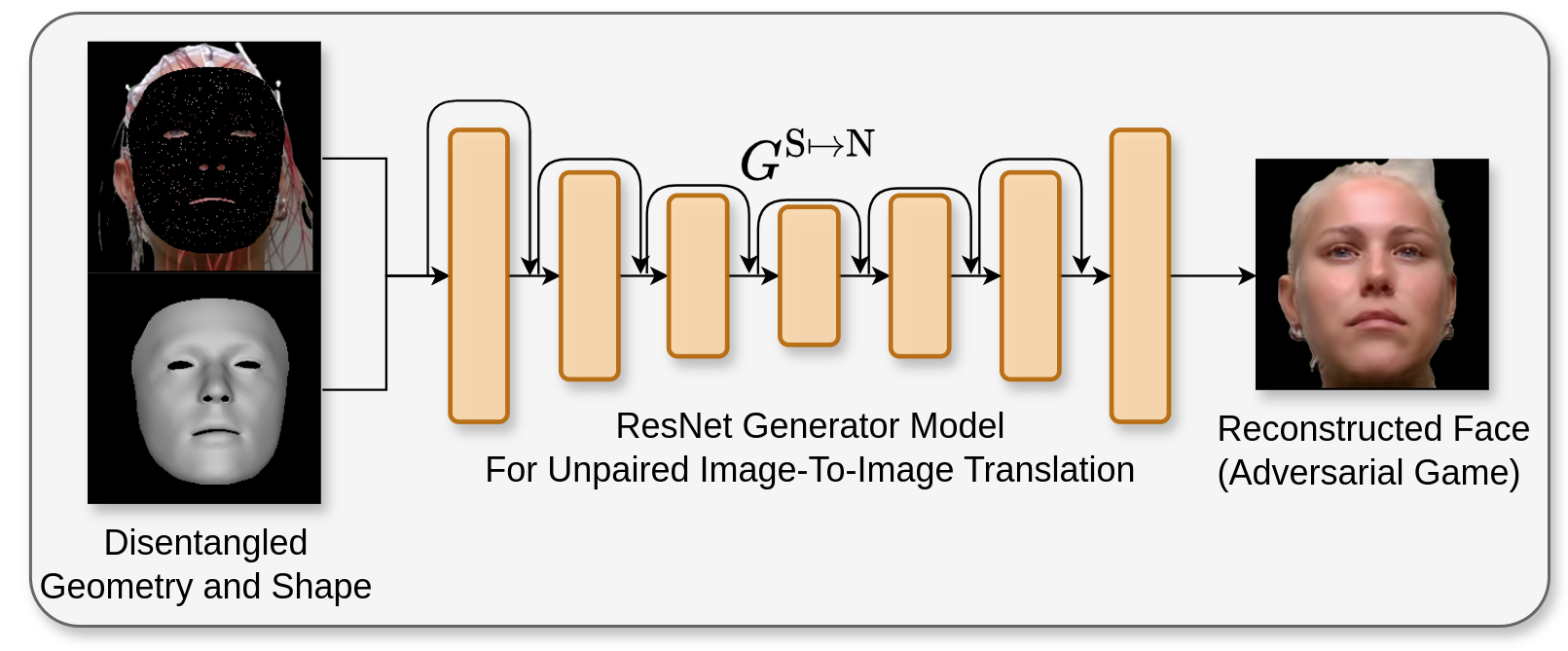}
    \caption{
        \textbf{EIFER Generator Architecture:}
        Our EIFER generator architecture is based on the ResNet~\protect\citeSup{resnet_supRef} backbone, which serves as the neural generator for restoring faces. 
        We incorporate skip connections to facilitate information flow, similar to the U-Net~\protect\citeSup{ronneberger2015u_supRef, cyclegan_supRef} architecture employed in SMIRK~\protect\citeSup{smirk_supRef}.
        However, we introduce two modifications: (1) we utilize instance normalization~\protect\citeSup{instancenorm_supRef} to generate nuanced details, and (2) we replace the Conv2DTranspose layers with a combination of upsampling and convolutional layers to eliminate the checkerboard patterns.
        }
    \label{fig:app:generator}
\end{figure}

Our ResNet Generator, shown in \Cref{fig:app:generator}, consists of 9 residual blocks with a feature depth of 64, similar to the parameter amount of the original UNet in SMIRK~\protect\citeSup{smirk_supRef}.

\subsection{Updated Masking Function}
The masking function, originally proposed in~\protect\citeSup{smirk_supRef}, selects a random pixel to represent facial appearance. However, this function relies on computing facial landmarks in the input images to define a suitable sampling area.
Unfortunately, this is not feasible under sEMG occlusion, as demonstrated in \Cref{fig:app:landmarks}.

We reformulate the sampling area based on the rendered FLAME face model to address this limitation.
This is made possible by the sufficient pre-training of the encoder models, allowing us to tackle this complex problem without requiring a retrained sEMG occlusion-robust facial landmarking model.

\begin{figure}[h]
    \centering
    \includegraphics[width=\linewidth]{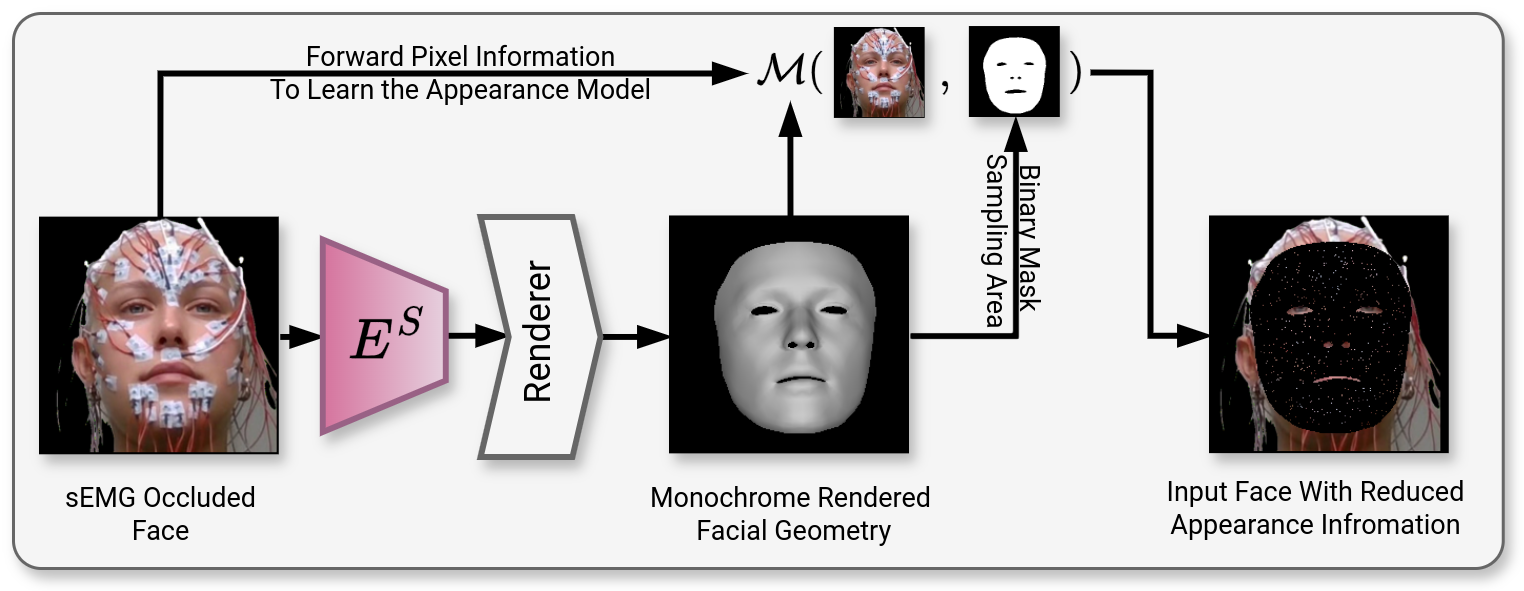}
    \caption{
        \textbf{Update Information Flow For the Pixel Masking:}
        As we cannot rely on the facial landmarks convex hull as a sampling area, we utilize the monochrome rendered facial geometry instead for the masking function $\mathcal{M}(\cdot)$~\protect\citeSup{smirk_supRef}.
        This has the advantage that we can utilize the learned alignment capabilities of SMIRK~\protect\citeSup{smirk_supRef}.
        The selected sampling area covers the facial area well. 
    }
    \label{fig:app:masking}
\end{figure}

As a result, EIFER implicitly becomes a robust facial landmarking tool under occlusion, as shown in the ablation studies. We illustrate the information flow of the updated masking function in \Cref{fig:app:masking}.

\subsection{Discriminators}
We employ a simple yet effective discriminator model inspired by previous works~\protect\citeSup{goodfellow2014generative_supRef, cyclegan_supRef, buchner2023improved_supRef}, distinguishing between generated faces and their unpaired reference images.
Specifically, we compare generated faces with removed and applied sEMG electrodes, ensuring that the generator produces realistic faces consistent with the input data.

To train the generator, we utilize the least square GAN loss~\protect\citeSup{lsgan_supRef}, which encourages the generator to produce more realistic faces by enforcing generation near the decision boundary.
This loss function helps to stabilize the training process and improve the overall quality of the generated faces.
Consequently, the problem of distinguishing between real and fake faces is now reduced to a two-class decision problem.
Our discriminator network consists of a 3-layer convolutional neural network with two output neurons, which classify input images as real or fake.

\subsection{Multi-Stage Training}
We adopt a multi-stage training approach for the two encoder-generator pairs and the two-phase training protocol to overcome the batch size limitation.
This approach is critical due to the challenging nature of our problem, where facial features are obstructed by electrodes, making expression extraction difficult.

Inspired by previous works~\protect\citeSup{buchner2023let_supRef, buchner2023improved_supRef}, we employ a two-stage training strategy. In the first stage, we train the entire architecture with frozen encoders and provide more appearance pixels to the generator.
This allows the model to learn to disregard the correct facial expression and focus on generating faces that can fool the discriminators while implicitly encoding facial geometry in the appearance.

In the subsequent stages, we enforce the disentanglement of geometry and appearance by (1) enabling the encoder on sEMG occluded faces to update its weights and (2) gradually reducing the available appearance information.
As a result, the model is forced to rely on the estimated facial geometry to restore correct faces over time.

Combining this multi-stage approach with the regularization terms introduced in the main paper ensures that the encoders correctly compute shape, expression, and position. This approach is crucial for achieving convergence, as it would otherwise require significantly more training effort.

\subsection{EMG2Exp And Exp2EMG Architecture}
We utilize simple multi-layer perceptrons (MLPs) to learn the non-linear relationship between the input data and the desired output for both our EMG2Exp (Synthesis) and Exp2EMG (Analysis) networks.
These MLPscapture the complex relationships between the electromyography (EMG) signals and the corresponding facial expressions and vice versa.
By employing MLPs, we effectively model the non-linear interactions between the input and output data.

As previously discussed, these models are trained in the second phase of EIFER, as the first phase requires a batch size of one.
Therefore, we could not guarantee convergence of the training.
By training them separately in the second phase, we can ensure that they learn the complex relationships between the input and output data effectively.

\begin{figure}[h]
    \centering
    \includegraphics[width=\linewidth]{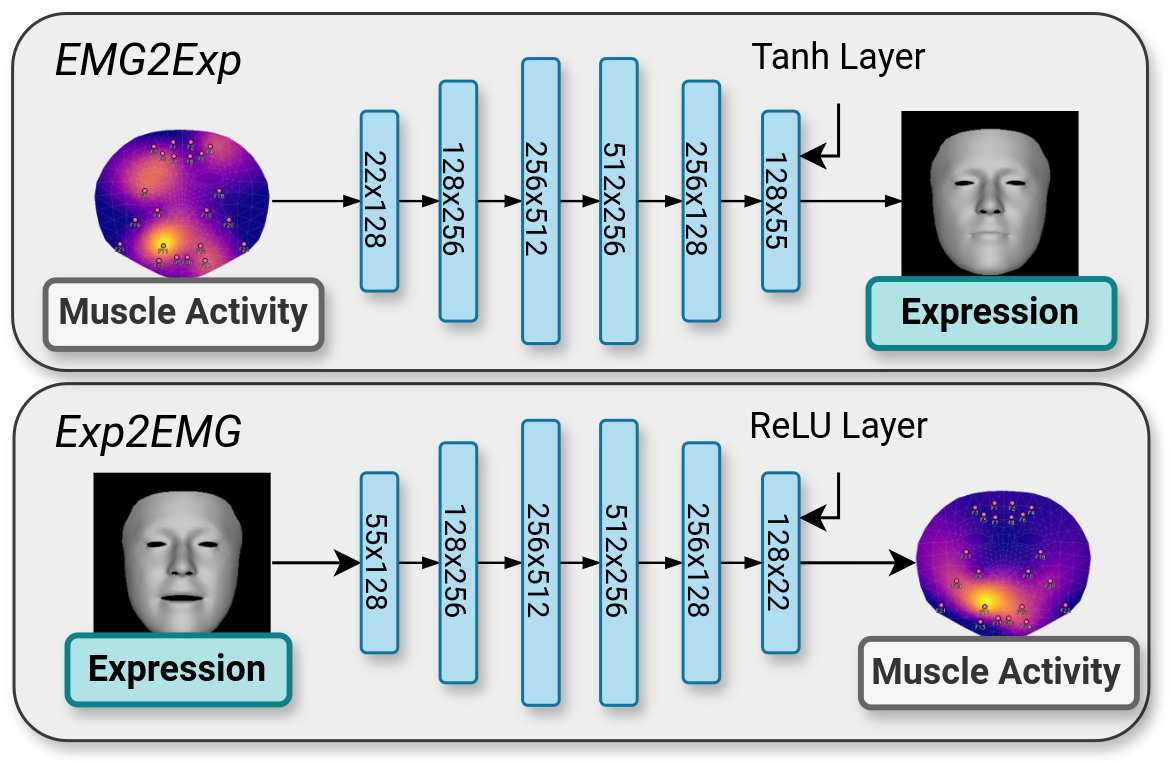}
    \caption{
        \textbf{\textit{EMG2Exp} and \textit{Exp2EMG} Architectures:}
        The simple MLP architecture learns the non-linear mapping between facial expression and muscle activity.
        Thus, the models learn the correspondence between these two domains.
    }
    \label{fig:app:emg2exp}
\end{figure}

We provide a detailed illustration of both the \textit{EMG2Exp} and the \textit{Exp2EMG} models in \Cref{fig:app:emg2exp}. In terms of architecture, we employ a simple yet effective design, utilizing \texttt{ReLU} activations~\protect\citeSup{relupaper_supRef} for all intermediate layers. This choice of activation function allows the models to learn non-linear relationships between the input and output data.

The final layer of each model is designed to accommodate the specific requirements of the output data. For the \textit{EMG2Exp} model, we use a \texttt{Tanh} activation function, which allows the model to produce output values in the range of -1 to 1 (the typical ranges for the 3DMM expression space), suitable for representing facial expressions. 
In contrast, the \textit{Exp2EMG} model uses a \texttt{ReLU} activation function in the final layer, as the sEMG signals are non-negative and require a non-negative output range.

During our experiments, we explored various expression encoder models, including DECA~\protect\citeSup{deca_supRef}, EMOCAv2~\protect\citeSup{emoca_supRef}, FOCUS~\protect\citeSup{FOCUS_supRef}, and Deep3DFace~\protect\citeSup{shang2020self_supRef}.
To accommodate the unique characteristics of each model, we adapted the input and output dimensions of our architecture accordingly, taking into account each model's specific expression parameter dimensions.
This allowed us to effectively integrate these different expression encoder models into our framework and evaluate their performance in our experiments.
Therefore, we compare the expression independently of the model architecture, gaining more insights into their underlying 3DMM and encoder capabilities instead.

\section{Dataset - Mimics And Muscles}
We created a custom dataset that simultaneously captures facial mimicry and muscle activity, bridging the gap between these two aspects.
To our knowledge, this is the first dataset of its kind.

This dataset provides new insights into the complex dynamics between facial expressions and muscle activity.
We provide a detailed description of the recording setup, recording scheme, surface electromyography schemes, data processing, and general data statistics to facilitate a deeper understanding of the dataset.

\subsection{Recording Setup}
In our experimental setup, a set of participants was instructed by an instruction video~\protect\citeSup{schaede_supRef} to perform different facial movements.
Each movement was repeated three times; thus, we can compare the repetitions against each other.
Each movement task varied in time, ranging from 10 to 30 seconds.
First, the following eleven facial movements were performed in that order:

\begin{enumerate}
    \item Face-At-Rest
    \item Forehead-Raise
    \item Eye-Gentle
    \item Eye-Tight
    \item Smile-Closed
    \item Smile-Open
    \item Nose-Wrinkler
    \item Cheeks Blow
    \item Lip-Pucker
    \item Snarl
    \item Depress-Lip
\end{enumerate}

Afterward, the participants had to mimic the six basic emotions~\protect\citeSup{ekmanArgumentBasicEmotions1992_supRef} four times in total random order.
The participants were shown faces to recreate.
This ensured that no memory effect of previous repetitions could set in.
Each expression was shown for three seconds, followed by a three-second interval for repetition.
At 4.5 seconds, we assume the height peak during the expression.

We repeated the experiment twice with sEMG electrodes attached to measure muscle activity and once without electrodes as a reference.
The duplicate sEMG measurement was conducted to ensure the reliability of the sEMG results.
Additionally, we repeated the entire experiment two weeks later to account for potential changes in muscle activity and minimize inaccuracies that may arise from the participants' daily state.
This allowed us to capture a more comprehensive and accurate representation of the participants' muscle activity over time.

Our participants were recorded with a frontal-facing Intel RealSence Depth Camera D415 (Intel Corporation, Santa Clara, California, U.S.) at $1280 \times 720$ resolution.
Unfortunately, the obtained 3D information was unreliable and inaccurate in supporting the monocular 3D facial reconstruction but suitable enough for foreground and background separation.

We employ the same data collection setup as in~\protect\citeSup{emg1_supRef, emg2_supRef, emg3_supRef}.
To minimize skin impedance, all participants thoroughly cleaned their faces with non-refatting medical soap.
The electromyography recording setup used reusable surface electrodes (Ag–Ag–Cl discs, diameter: 4 mm, DESS052606, GVB-geliMED, Bad Segeberg, Germany) to measure muscle activity.
Reference electrodes (H93SG, Kendall, Germany) were bilaterally attached to the mastoid bone to provide a stable reference point.
The muscle signals were amplified using sEMG amplifiers (ToEM16G, gain 100, frequency range 10–1,861 Hz, DeMeTec, Langgöns, Germany).
Then they converted with an analog to digital converter (Tom, resolution: 5.96 nV/Bit, sampling rate: 4096/s, cutoff frequency: 2048 Hz, DeMeTec, Langgöns, Germany).
The digitized data were then sampled using ATISArec (GJB Stentechnik, Ilmenau, Germany).

\begin{figure}[h]
    \centering
    \includegraphics[width=\linewidth]{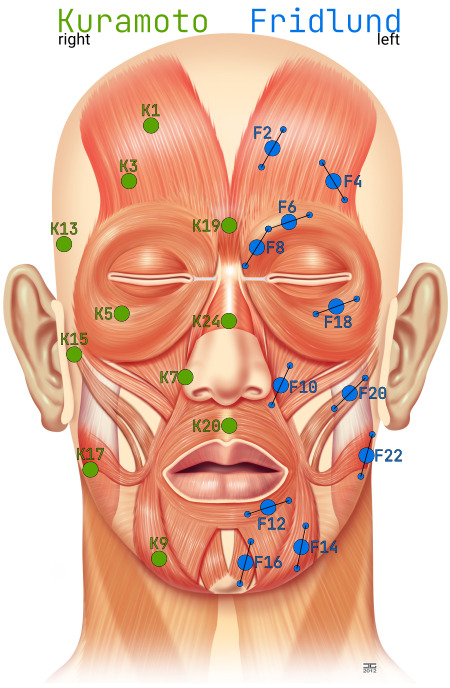}
    \caption{
        \textbf{sEMG Electrode Locations:}
        We highlight both surface electromyography schemes on their corresponding anatomical locations.
        Both Fridlund~\protect\citeSup{fridlund_supRef} (\textbf{F}, blue) and Kuramoto~\protect\citeSup{kuramoto_supRef} (\textbf{K}, red) are attached to both face sides, but drawn here only on one for clarity.
        Please note that Fridlund is a bipolar scheme (denoted by the two smaller dots per electrode), and Kuramoto is monopolar by using \textbf{K24} as reference.
    }
    \label{fig:app:electrodes}
\end{figure}

Our experimental setup allowed us to simultaneously record both the Fridlund~\protect\citeSup{fridlund_supRef} and Kuramoto~\protect\citeSup{kuramoto_supRef, emg2_supRef, emg1_supRef} surface electromyography (sEMG) schemes. 
However, it is essential to note that the Kuramoto scheme provides regional information on muscle activity, whereas the Fridlund scheme offers more precise activation data. The electrode locations are illustrated in \Cref{fig:app:electrodes}, and our medical partners ensured accurate anatomic placement.
A detailed description of the electrode channels is provided in \Cref{tab:app:electrodes}, which reveals that some Fridlund electrodes overlap with Kuramoto electrodes in specific locations.
For a more comprehensive understanding of the sEMG schemes and electrode placement, we refer the reader to previous studies~\protect\citeSup{emg1_supRef, emg2_supRef, emg3_supRef}.

While our primary focus is on the facial muscles responsible for expressions, we also recorded the activity of the \textit{M. masseter}, a digestive muscle, and the \textit{M. temporalis}.
The facial muscles have been linked to specific facial movements through the Facial Action Coding System (FACS)~\protect\citeSup{ekmanFacialActionCoding1978_paper_supRef}, which is also included in \Cref{tab:app:electrodes}. 
Notably, since we directly recorded facial expressions and muscle activity, we can bypass the intermediate Action Unit (AU) proxy variable in our approach.
This unique aspect of our study offers benefits for improving and investigating the established FACS, providing new insights into the relationship between facial muscles and expressions.

\begin{table}[h]
    \centering
    \resizebox{\columnwidth}{!}{%
        \begin{tabular}{lllll}
	\toprule
	Fridlund       & Kuramoto   & Muscle                     & Action Unit & Movement             \\
	\midrule
	F1, F2         & K1, K2     & medialer frontalis         & AU1         & inner brow raiser    \\
	F3, F4         & K3, K4     & lateraler frontalis        & AU2         & outer brow raiser    \\
	F5, F6, F7, F8 & K19        & glabellae                  & AU4         & brow lowerer         \\
	                &            & depressor supercilii       &             &                      \\
	                &            & corrugator supercilii      &             &                      \\
	F17, F18       & K5, K6     & orbicularis oculi          & AU6         & cheek raiser         \\
	F9, F10        & K7, K8     & levator labii superioris   & AU9         & nose wrinkler        \\
	F9, F10        & K7, K8     & levator labii superioris   & AU10        & upper lip raiser     \\
	F19, F20       & -          & zygomaticus minor          & AU11        & nasolabial deepener  \\
	F19, F20       & (K15, K16) & zygomaticus major          & AU12        & lip corner puller    \\
	F13, F14       & -          & depressor anguli oris      & AU15        & lip corner depressor \\
	F15, F16       & K9, K10    & mentalis                   & AU17        & chin raiser          \\
	F11, F12       & (K20)      & philtrum, orbicularis oris & AU22        & lip funneler         \\
	F11, F12       & (K20)      & orbicularis oris           & AU23        & lip tightener        \\
	F11, F12       & (K20)      & philtrum, orbicularis oris & AU24        & lip pressor          \\
	F21, F22       & K17, K18   & masseter                   & AU26        & jaw drop             \\
	F11, F12       & -          & philtrum, orbicularis oris & AU28        & lip suck             \\
	-              & K13, K14   & temporalis                 & -           & -                    \\
	\bottomrule
\end{tabular}
    }
    \caption{
        \textbf{Electrode Channels and Muscles:}
        With our two sEMG electrode schemes, we capture the majority of facial muscles.
        We also included the according action units~\protect\citeSup{ekmanFacialActionCoding1978_paper_supRef}, providing insights into further research in the future.
        Please note channel names surrounded by brackets are just roughly attributable to the muscles, and \textbf{K11} and \textbf{K12} do not exist in the Kuramoto scheme~\protect\citeSup{kuramoto_supRef, emg1_supRef, emg2_supRef, emg3_supRef}.
    }
    \label{tab:app:electrodes}
\end{table}

\subsection{Participant Cohort}
We recruited 36 participants (19 \venus, 17 \mars, age range: 18-67 years) without a history of any neurological disease to obtain synchronous facial expression and muscle activity for this study.
We specifically selected beardless male participants to ensure the accurate application of surface electromyography (sEMG) electrodes.
Although our sample size is limited and may not represent an entire population, we aimed to achieve a balanced distribution of male and female participants across various age ranges.
While the generalizability of our findings to a broader population remains uncertain, we expect the results to be consistent within this cohort, providing a reliable basis for further investigation.

To account for potential occlusions caused by the surface electromyography (sEMG) electrodes on key facial features, we recorded the same participants without electrode occlusion.
This additional recording protocol allowed us to establish a reference dataset, which serves as a baseline for evaluating the accuracy of shape and expression reconstruction.
By comparing the reconstructed results with the unoccluded recordings, we can assess the effectiveness of our approach in capturing the nuances of facial expressions despite electrode placement.

\subsection{Video Preprocessing}
\begin{figure}[h]
    \centering
    \begin{subfigure}[b]{0.49\linewidth}
         \centering
         \includegraphics[width=\textwidth]{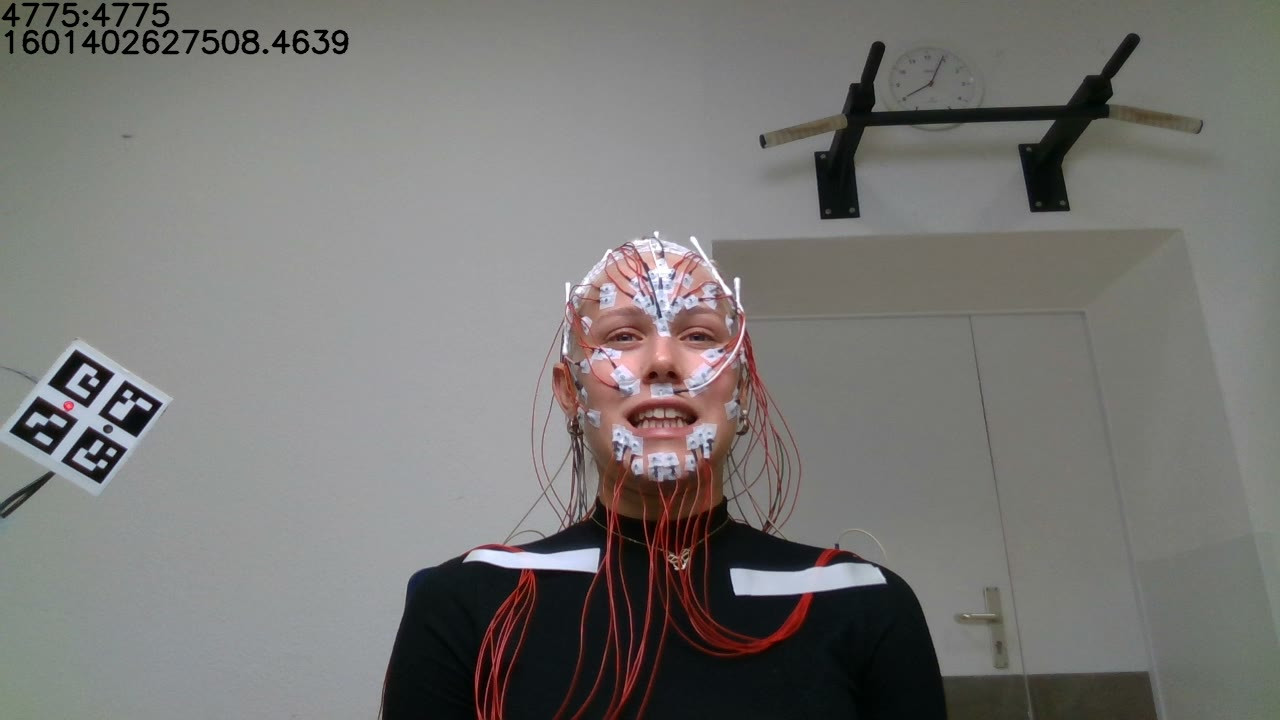}
         \caption{Raw Input Video of Intel RealSense}
     \end{subfigure}
     \hfill
     \begin{subfigure}[b]{0.49\linewidth}
         \centering
         \includegraphics[width=\textwidth]{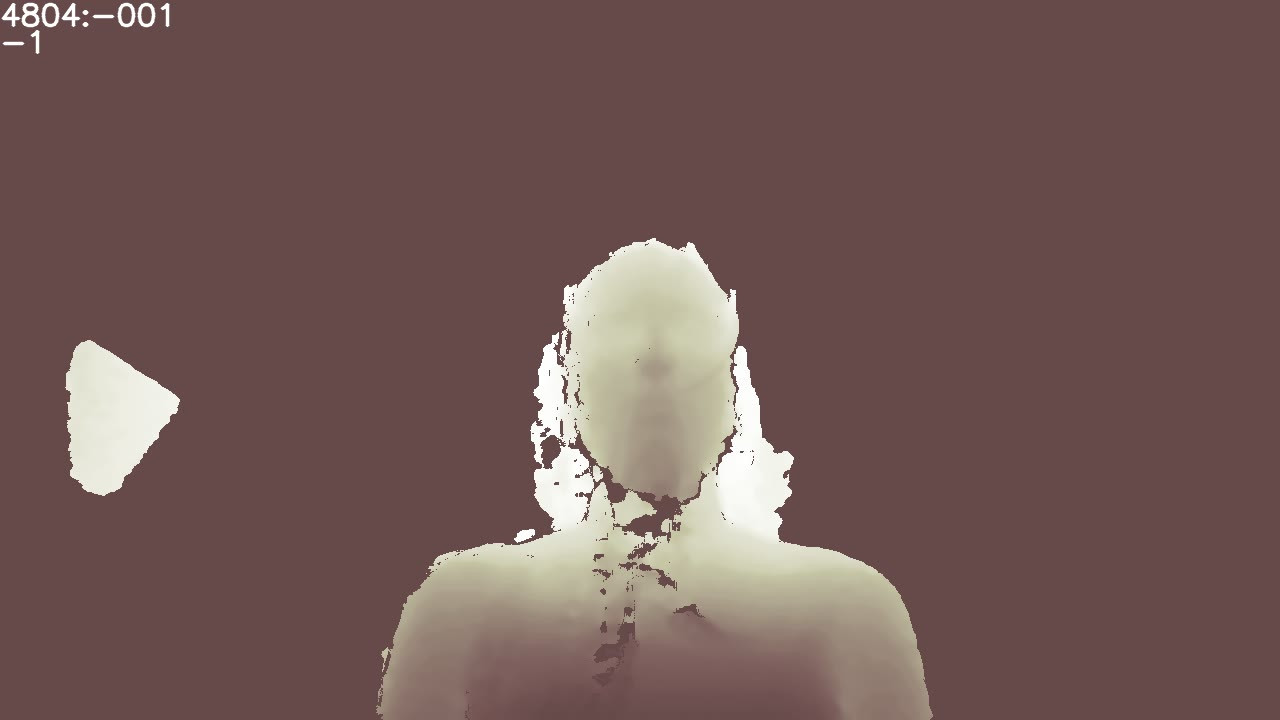}
         \caption{Visualize Depth Map of the Intel RealSense}
     \end{subfigure}
     \newline
     \begin{subfigure}[t]{0.3\linewidth}
         \centering
         \includegraphics[width=\textwidth]{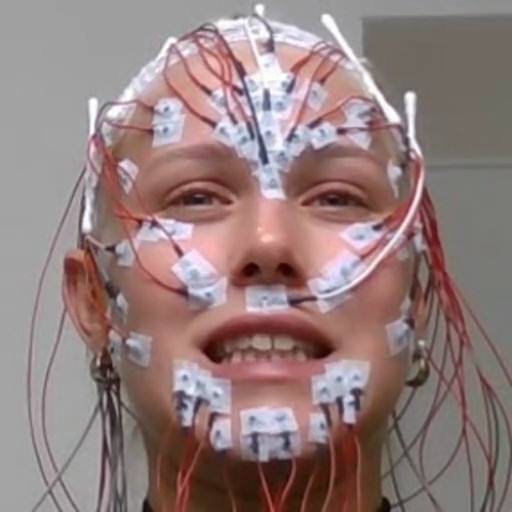}
         \caption{Estimated Face Bouding Box via BlazeFace~\protect\citeSup{bazarevsky2019blazeface_supRef}}
     \end{subfigure}
     \hfill
     \begin{subfigure}[t]{0.3\linewidth}
         \centering
         \includegraphics[width=\textwidth]{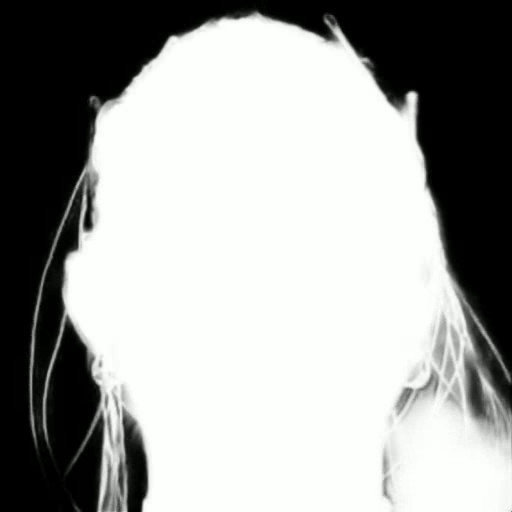}
         \caption{Foreground masking via depth map and MODNet~\protect\citeSup{MODNet_supRef}}
     \end{subfigure}
     \hfill
     \begin{subfigure}[t]{0.3\linewidth}
         \centering
         \includegraphics[width=\textwidth]{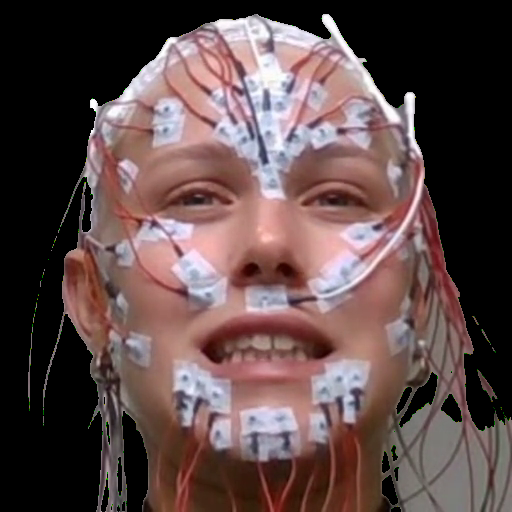}
         \caption{Cropped and extracted face for the training}
     \end{subfigure}
    \caption{
        \textbf{Video Preprocessing:}
        We illustrate the preprocessing steps of the recorded facial videos for EIFER training, where we remove the background to facilitate facial expression extraction by the encoder models.
        Please note that you can see the Aruco markers on the left side of the raw input frame, which is used for the synchronization. 
    }
    \label{fig:app:recording}
\end{figure}

We provide a visualization of the original recording captured by the Intel RealSense camera, showcasing both RGB and depth data, in \Cref{fig:app:recording} for a representative participant. 
To focus the model's attention on the most relevant facial regions, we employed the BlazeFace model~\protect\citeSup{bazarevsky2019blazeface_supRef} to compute the facial bounding box.
However, not every frame yielded a valid bounding box, likely due to minor face orientation or unaccountable lighting changes.
To address this, we interpolated missing bounding boxes using the position from the previous frame, assuming minimal participant movement due to the attached electrodes hindering a lot of movement.
Additionally, Aruco markers placed on the left side of the frame facilitated synchronization across different data streams.

Following the extraction of bounding boxes, we leveraged rough depth information to segment the face from the background, thereby mitigating potential influences from external factors, such as people in the background.
Subsequently, we applied a matting estimation technique using MODNet~\protect\citeSup{MODNet_supRef} to refine the segmentation results.
Please note that the cables around the shoulder and neck area still make this segmentation challenging and might introduce artifacts.
The outcome of this process is illustrated in \Cref{fig:app:recording}, also with artifacts above the right shoulder area.
In conjunction with the recorded muscle activity data, these preprocessed frames were then utilized to train the EIFER model.

\subsection{Electromyography Signal Preprocessing}
\begin{figure}[h]
    \centering
    \includegraphics[width=\linewidth]{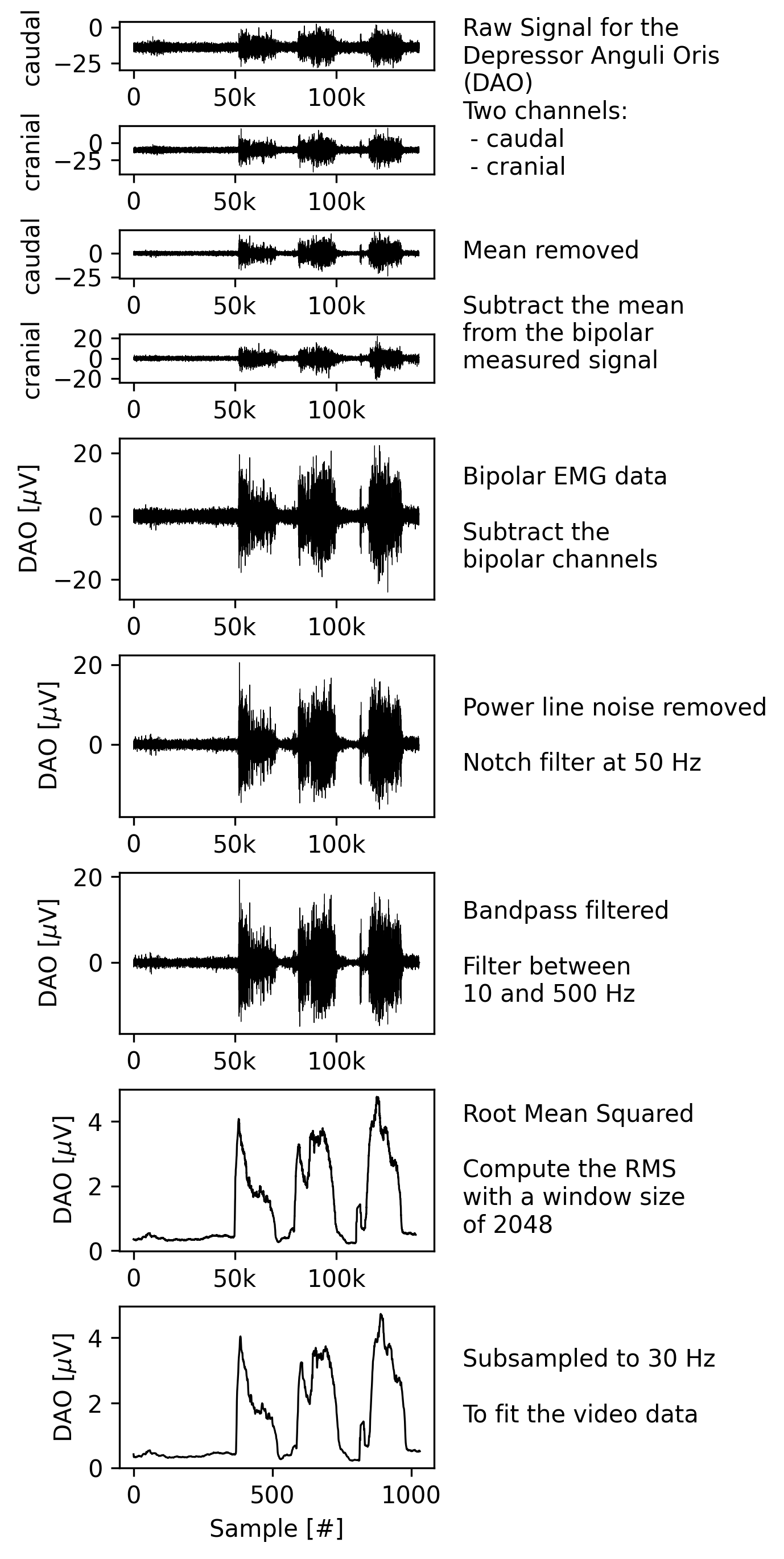}
    \caption{
        \textbf{Muscle Signal Preprocessing:}
        Illustration of the data processing pipeline for a single sEMG measurement of the \textit{depressor anguli oris} muscle during the "Smiling" movement.
        Note the variability in the linear envelope of the measured muscle activity, even for repeated instances of the same movement.
    }
    \label{fig:app:data-processing}
\end{figure}

We adhere to established protocols for processing the recorded electromyography (EMG) signals, as described in previous studies~\protect\citeSup{emg1_supRef, emg2_supRef, emg3_supRef, funkWirelessHighresolutionSurface2024_supRef, xiaEMGBasedEstimationLimb2018_supRef}. 
Specifically, we focus on the Fridlund scheme~\protect\citeSup{fridlund_supRef} due to its direct association with the corresponding muscles.
As illustrated in \Cref{fig:app:data-processing}, our processing pipeline is uniformly applied to all sEMG recordings, including those for the \textit{M. depressor anguli oris} (F19, F20) during the functional movement of \textit{smiling}. The resulting signal exhibits the three repetitions of the movement.
Notably, we refrain from normalizing the data during this preprocessing step, intentionally delaying normalization until the training phase to preserve participant-specific characteristics and avoid loss of information.

Unlike most research that typically operates on high-resolution sEMG (HR-sEMG) signals at 4096 Hz, we need to synchronize our signal with the recorded video at 30 frames per second (FPS).
We employ a Fast Fourier Transform (FFT)-based downsampling approach~\protect\citeSup{2020SciPy-NMeth_supRef}, carefully ensuring that the essential frequency features are preserved. 
As the example demonstrates, the downsampling operation effectively maintains the signal's overall shape and fine-grained nuances.
By successfully recording muscle activity and facial expression, we can explore the relationship between these two modalities, enabling a more comprehensive understanding of the underlying mechanisms.

\subsection{Synchronization}
We implemented an automated triggering system that simultaneously initiated both data streams to ensure precise synchronization between the video recording and surface electromyography (sEMG) signals.
Additionally, we incorporated visual and sEMG-based synchronization triggers, which were repeated twice to guarantee accurate alignment; see \Cref{fig:app:recording} for the Aruco markers.
This dual-triggering approach allowed us to align the video sections with the corresponding sEMG signals confidently.
However, despite this rigorous synchronization protocol, some recordings still exhibited low confidence levels, necessitating their exclusion from the dataset.
To provide transparency and account for these variations, we report the number of suitable recording snippets employed during training and evaluation for each facial movement in \Cref{tab:app:sync}.
This information promotes a more nuanced understanding of the dataset's composition and the reliability of our results.

\begin{table}[h]
    \centering
    \begin{tabular}{lccc}
	\toprule
	Facial Movement & Total Recordings & Usable & Failed \\
	\midrule
	Face-At-Rest    & 141              & 105    & 36     \\
	Forehead-Raise  & 141              & 106    & 35     \\
	Eye-Gentle      & 141              & 106    & 35     \\
	Eye-Tight       & 141              & 106    & 35     \\
	Cheeks-Blow     & 141              & 106    & 35     \\
	Smile-Closed    & 141              & 105    & 36     \\
	Smile-Open      & 141              & 106    & 35     \\
	Nose-Wrinkler   & 141              & 107    & 34     \\
	Lip-Pucker      & 141              & 106    & 35     \\
	Snarl           & 141              & 106    & 35     \\
	Depress-Lip     & 141              & 105    & 36     \\
	\midrule
	angry           & 560              & 528    & 32     \\
	disgusted       & 560              & 528    & 32     \\
	fearful         & 560              & 528    & 32     \\
	happy           & 560              & 528    & 32     \\
	sad             & 560              & 528    & 32     \\
	surprised       & 560              & 528    & 32     \\
	\midrule
	$\sum$          & 4911             & 4332   & 579    \\
	\bottomrule
\end{tabular}
    \caption{
        \textbf{Synchronization Results:}
        We show how many recordings (at 30 FPS) of the synchronized facial expression and muscle activity are available for training and evaluation.
        Please note that the occlusion-free reference recording can be used fully.
    }
    \label{tab:app:sync}
\end{table}

\subsection{Limitations}
Our dataset is subject to several limitations that warrant consideration.
Firstly, the facial expressions mimicked by participants may not accurately reflect natural, evoked expressions, as noted in previous studies~\protect\citeSup{emoca_supRef, ekmanArgumentBasicEmotions1992_supRef, affectnet_supRef}.
However, this limitation does not compromise our ability to predict muscle activity from expressions and vice versa, as the measured facial muscle activity and recorded facial expressions still exhibit a strong correlation.
This alignment ensures we can investigate the relationship between muscle activity and facial expressions despite the potential differences between mimicked and natural expressions.

\begin{figure}[h]
    \centering
    \includegraphics[width=\linewidth]{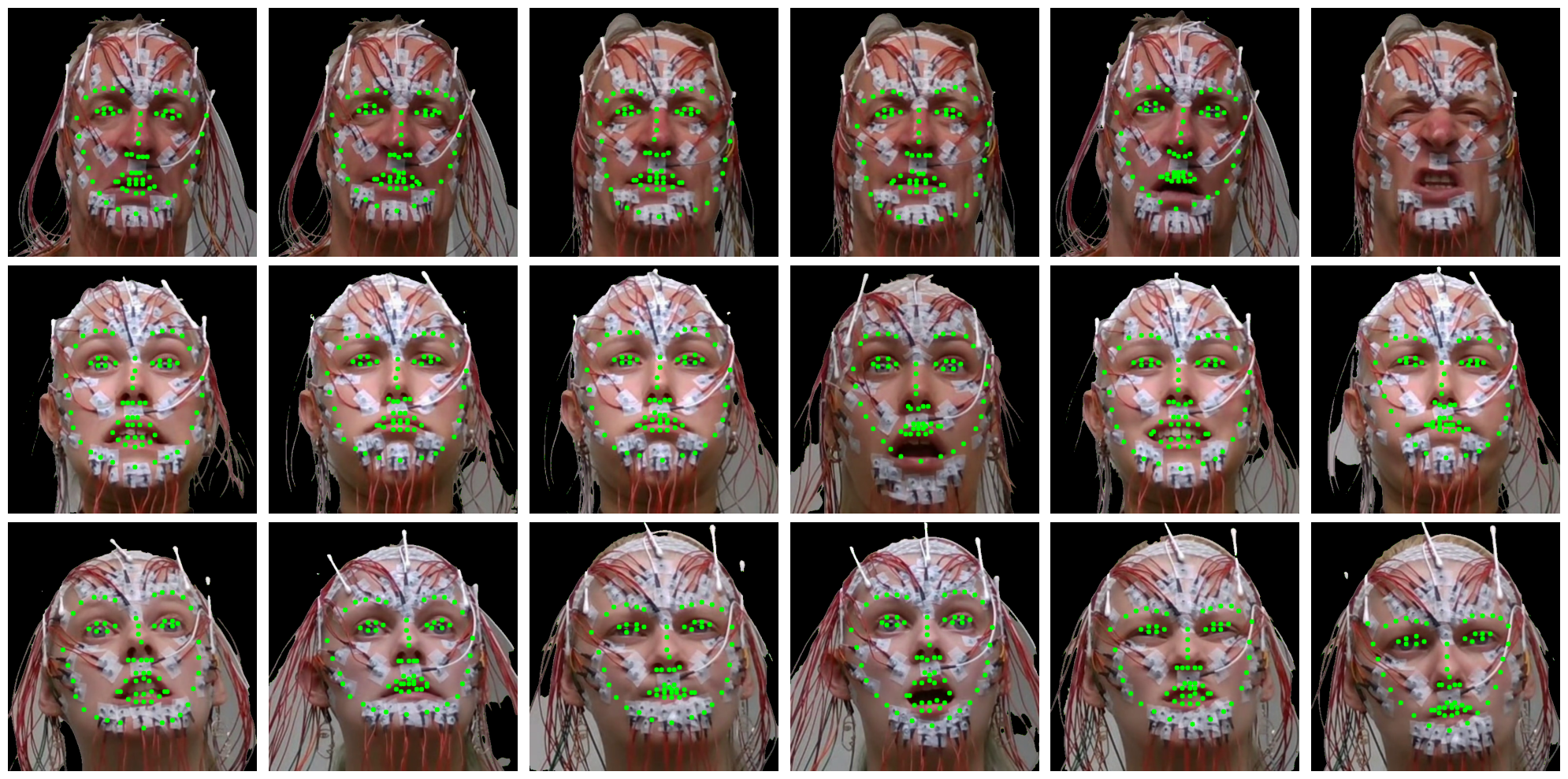}
    \caption{
        \textbf{Examples of Occluded Facial Expressions with Predicted Landmarks:}
        We present several examples of different facial expressions from three study participants.
        In addition to the original images, we overlay the predicted landmarks, as obtained using the methods described in~\protect\citeSup{liuDenseFaceAlignment2017_supRef, bazarevsky2019blazeface_supRef}.
        Notably, the predicted landmarks, if at all, exhibit inaccuracies, particularly in the outer regions of the face.
        This is of concern, as the outer part of the face is used to compute the convex hull for the masking function in SMIRK~\protect\citeSup{smirk_supRef}, and the strong offset observed in this region may impact the accuracy of the masking process.
    }
    \label{fig:app:landmarks}
\end{figure}

Secondly, the surface electromyography (sEMG) electrodes used in our study introduce significant occlusion, which poses a challenge for feature extraction as illustrated in \Cref{fig:app:landmarks}.
Existing methods are not trained on such data, and we cannot determine the potential bias these methods may introduce into our model~\protect\citeSup{buchner2024facing_supRef, buchner2024power_supRef}. 
This highlights the need for more robust facial feature extraction to handle occlusions and ensure accurate predictions effectively.

Further, our recording setup is limited in capturing characteristic comprehensive muscle activity measurements.
Specifically, certain muscle activities, such as the voluntary evoked eyelid closure, are controlled by the palpebral part of the \textit{M. orbicularis oculi}, are not accounted for in the Fridlund scheme~\protect\citeSup{fridlund_supRef}. 
This omission is because the Fridlund scheme focuses on a specific set of facial muscles and the palpebral part of the \textit{M. orbicularis oculi} is not included in this set.
Additional measurements or specialized electrodes would be required to capture this activity, as discussed in previous studies~\protect\citeSup{steiner2024electro_supRef}.
This limitation highlights the need for more research concerning recording setups that can capture a broader range of muscle activities, enabling a more complete understanding of the complex relationships between facial muscles and expressions~\protect\citeSup{funkWirelessHighresolutionSurface2024_supRef}.

Our dataset has limitations, including its size and focus on functional movements, which may restrict the generalizability of our findings and the model's effectiveness in handling complex movements or subtle variations in facial expressions.
Additionally, the impact of extreme facial expressions or diseases like facial palsy on our approach is unclear and warrants further investigation.

Our dataset was recorded within a medical study in Germany, subject to strict data privacy regulations. 
As a result, we are limited in the number of faces we can display and the participants who agreed to share their data for new research databases. 
However, we will publish our trained models, \textit{EMG2Exp} and \textit{Exp2EMG}, which do not contain person-identifiable information, ensuring compliance with data protection regulations~\protect\citeSup{eggerIdentityExpressionAmbiguity3D2021_supRef}.

\section{Experimental Setup}
We compare EIFER to several state-of-the-art monocular 3D face reconstruction methods. 
Our comparison includes three models that employ the FLAME 3DMM~\protect\citeSup{FLAME_supRef}: DECA~\protect\citeSup{deca_supRef}, EMOCAv2~\protect\citeSup{emoca_supRef}, and SMIRK~\protect\citeSup{smirk_supRef}.
We also evaluate two models that use the BaselFaceModel~\protect\citeSup{bfm1_supRef, bfm2_supRef}: Deep3DFace~\protect\citeSup{shang2020self_supRef} and FOCUS~\protect\citeSup{FOCUS_supRef}.
Additionally, we compare MC-CycleGAN~\protect\citeSup{buchner2023let_supRef, buchner2023improved_supRef}, which does not rely on a face model and implicitly learns the reconstruction.
All models are available as PyTorch~\protect\citeSup{pytorch_supRef} implementations.

However, none of these methods were trained or tested on faces with sEMG electrodes attached. 
Moreover, our 36 participants were not part of their training data, making their faces completely unseen.

To ensure a fair comparison, we fine-tune all models on a common subset of occlusion-free reference recordings (10\% of available frames).
This approach has two benefits: First, it adapts the models to our data without occlusion, eliminating the need to account for their in-the-wild performance.
Therefore, we assume they will perform best.
Second, when applying the models to the sEMG-occluded faces of the same individuals, any behavior change can be attributed to the electrodes.
This allows us to assess the models' invariance to this type of occlusion.

In contrast, EIFER trains on the same subset of occlusion-free faces (and uses the occluded faces, also 10\% of available frames) as a reference to guide the reconstruction via adversarial challenge. 
As a result, all models have seen the same occlusion-free faces, making the comparison on the remaining 90\% of frames fair.

We report the training hyperparameters for the first phase of EIFER, which focuses on expression reconstruction under sEMG occlusion.

We employ two AdamW~\protect\citeSup{loshchilovDecoupledWeightDecay2019_supRef} optimizers to train the encoder-generator pairs and discriminators independently.
Both optimizers use a learning rate of $2\cdot 10^{-4}$ and a weight decay of $10^{-3}$.
A cosine annealing learning rate scheduler adapts the learning rate during training.
Again, we can only employ a batch size of one to facilitate the strength of instance normalization. 

EIFER is trained for 20 epochs, divided into three stages: 10 epochs for the first, 5 for the second, and 5 for the last.
We use 80\% of the 10\% available frames for training and 20\% for validation. Note that the reported results in the main paper are on the 90\% unseen frames.

During training, EIFER receives the triplet $(I^{\mnormal}, I^{\msensor})$, where $I^{\mnormal} \in \mathbb{R}^{224\times 224\times 3}$ is a color image of the occlusion-free face and $I^{\msensor} \in \mathbb{R}^{224\times 224\times 3}$ is a color image of the sEMG-occluded face.
We apply random data augmentations to the frames, including random cropping, sharpening, and horizontal and vertical flipping.

During the second phase of EIFER, we train \textit{EMG2Exp} and \textit{Exp2EMG} using the following hyperparameters.
We employ the Adam optimizer~\protect\citeSup{kingmaAdamMethodStochastic2017_supRef} with a learning rate of $10^{-3}$ and no additional learning rate scheduling or early-stopping.
We use a batch size of $512$ and train for 200 epochs. All results in the main paper are reported on a five-fold cross-validation.

Both models are trained on the tuple $(A, \varphi)$, where $A \in \mathbb{R}^{22}$ represents the 22 measured muscle signals using the Fridlund sEMG scheme.
We normalize the muscle signals $A$ by the maximum measured muscle activity \textit{for each participant.}
This normalization accounts for individual intensity and muscle strength differences, allowing for a more comparable analysis across participants.
Please note that this maximum value has been used to restore the reconstructed activity during the $Exp2EMG$ predictions.
$\varphi$ denotes the 3DMM expression space parameters.
The dimension of $\varphi$ varies across models:
\begin{itemize}
    \item For EIFER and SMIRK~\protect\citeSup{smirk_supRef}, $\varphi \in \mathbb{R}^{55}$ (50 expressions, two eyelids, three jaw).
    \item For DECA~\protect\citeSup{deca_supRef} and EMOCAv2~\protect\citeSup{emoca_supRef}, $\varphi \in \mathbb{R}^{53}$ (50 expression parameters and three jaw).
    \item For FOCUS, $\varphi \in \mathbb{R}^{100}$.
    \item For Deep3DFace, $\varphi \in \mathbb{R}^{64}$.
\end{itemize}
Please note that FLAME~\protect\citeSup{FLAME_supRef} models the jaw movement intentionally separate, and BFM~\protect\citeSup{bfm1_supRef, bfm2_supRef} models this implicitly via the expression space. 
This allows us to compare the expression space differences between FLAME~\protect\citeSup{FLAME_supRef} and BaselFaceModel~\protect\citeSup{bfm1_supRef, bfm2_supRef}.

\section{Visualizations And Videos}
We provide additional visual examples for each main experiment, including videos to highlight our approach's dynamic aspects and highlight our methods' advantages.

\subsection{Isolated Shape Visualization}
In our experiments, we observed that the same individual was reconstructed with varying facial geometries.
To investigate this, we analyzed the shape parameters of both FLAME~\cite{FLAME_supRef} and BFM~\cite{bfm1_supRef, bfm2_supRef} under neutral expressions, excluding camera or pose parameters.
Notably, EMOCAv2~\cite{emoca_supRef} employs the same encoder as DECA~\cite{deca_supRef}, resulting in identical shape parameter estimates.
Our analysis revealed that all models, except EIFER, exhibited differences in geometry for the same individual.
This discrepancy may explain why the expression parameters, compensating for the visual reconstruction, potentially affect the quality of muscle activation predictions in our later experiments.

\begin{figure}
    \centering
    \includegraphics[width=1.0\linewidth]{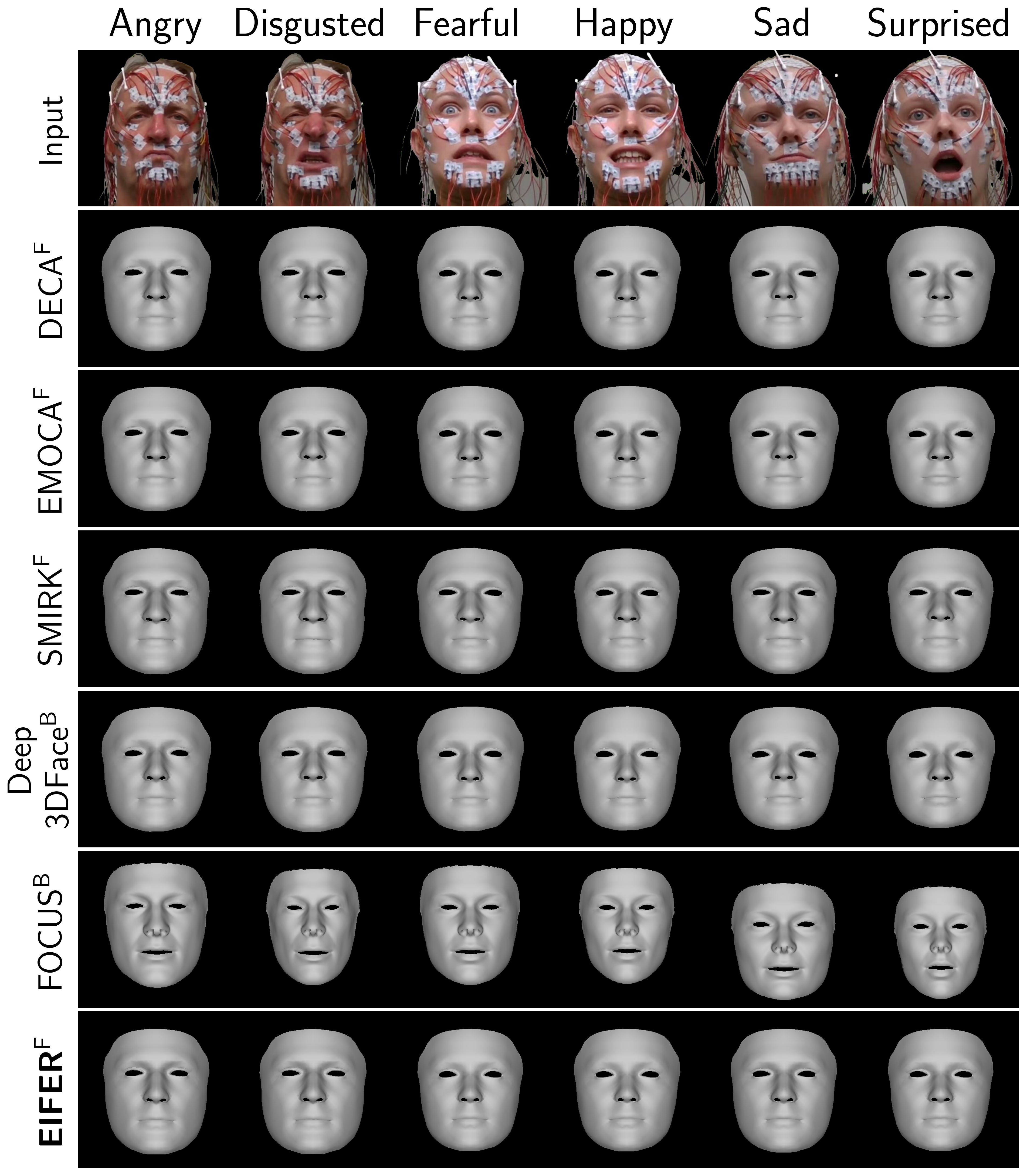}
    \caption{
        Isolated \textit{shape} parameters of the facial reconstruction.
        Many models have slightly different shape geometries for the same individual, indicating that the encoder might use the expression space to substitute the reconstruction.
    }
    \label{fig:eiferidt-only}
\end{figure}

\subsection{Reconstruction}
\begin{figure*}[h]
    \centering
    \includegraphics[width=\linewidth]{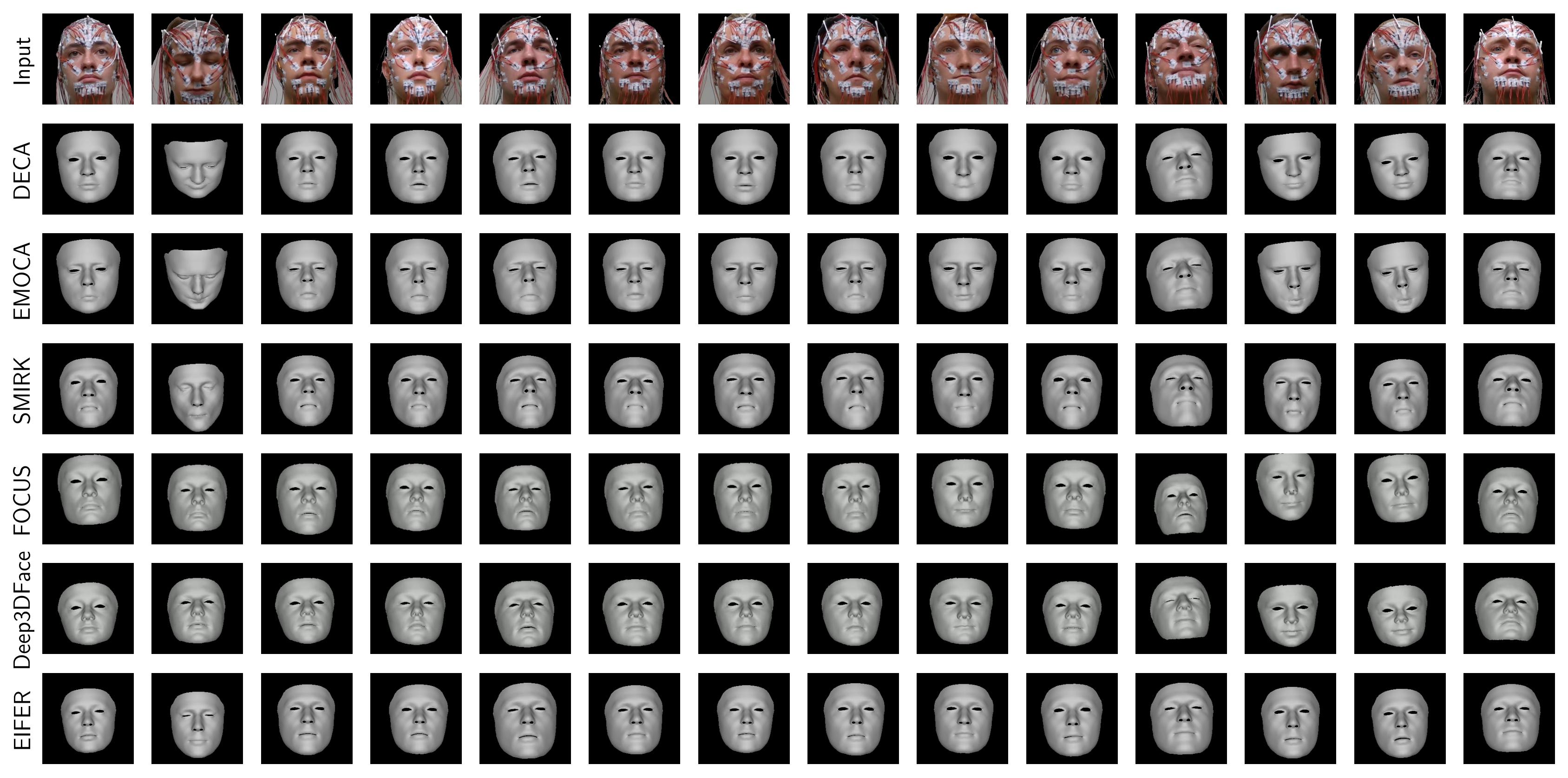}
    \caption{
        Facial Geometry Reconstruction during \textbf{Face-At-Rest}
    }
    \label{fig:app:syn:shape:face-at-rest}
\end{figure*}

\begin{figure*}[h]
    \centering
    \includegraphics[width=\linewidth]{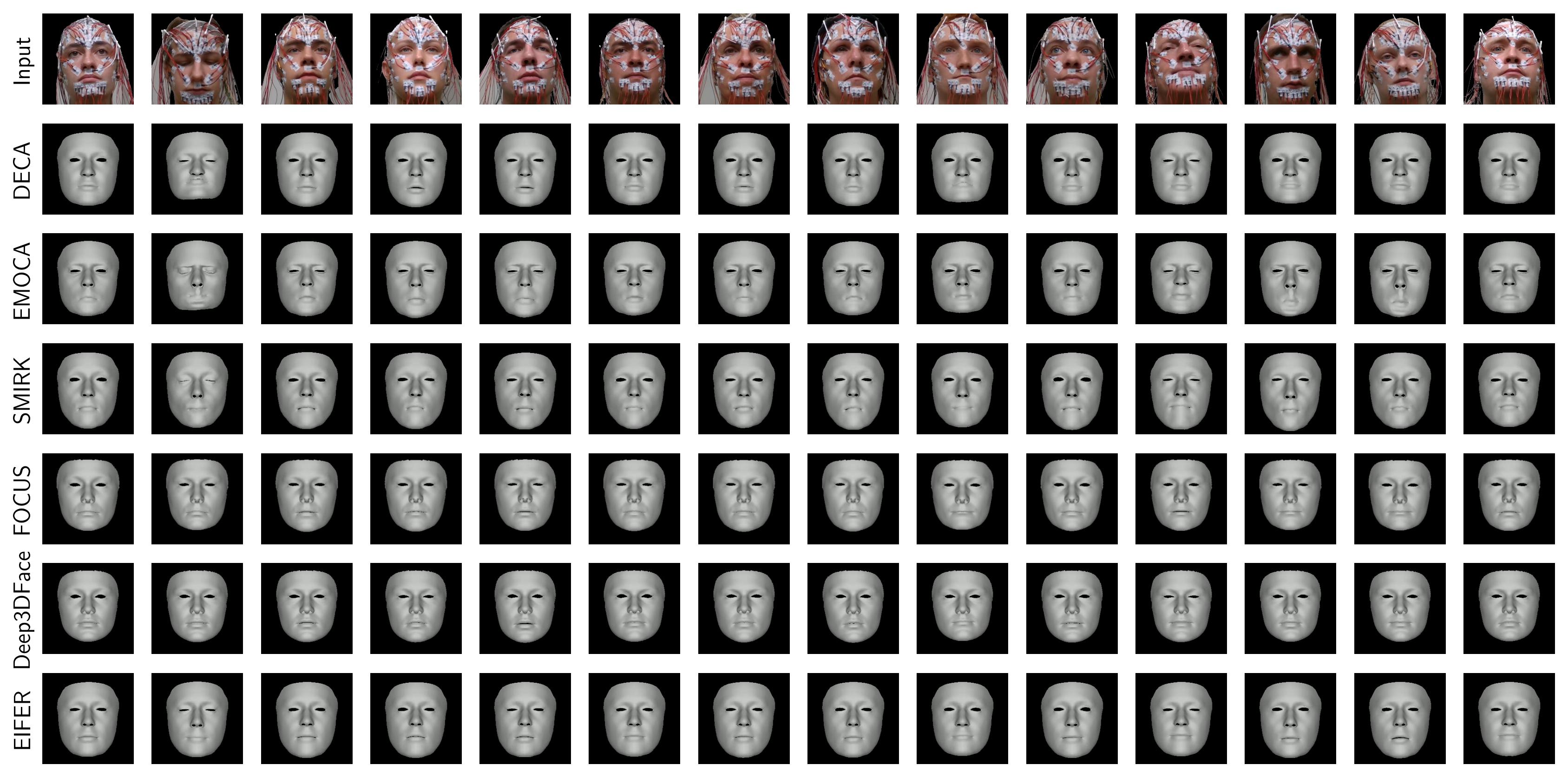}
    \caption{
        Isolated Facial Expression Reconstruction during \textbf{Face-At-Rest}
    }
    \label{fig:app:syn:emo:face-at-rest}
\end{figure*}

\begin{figure*}[h]
    \centering
    \includegraphics[width=\linewidth]{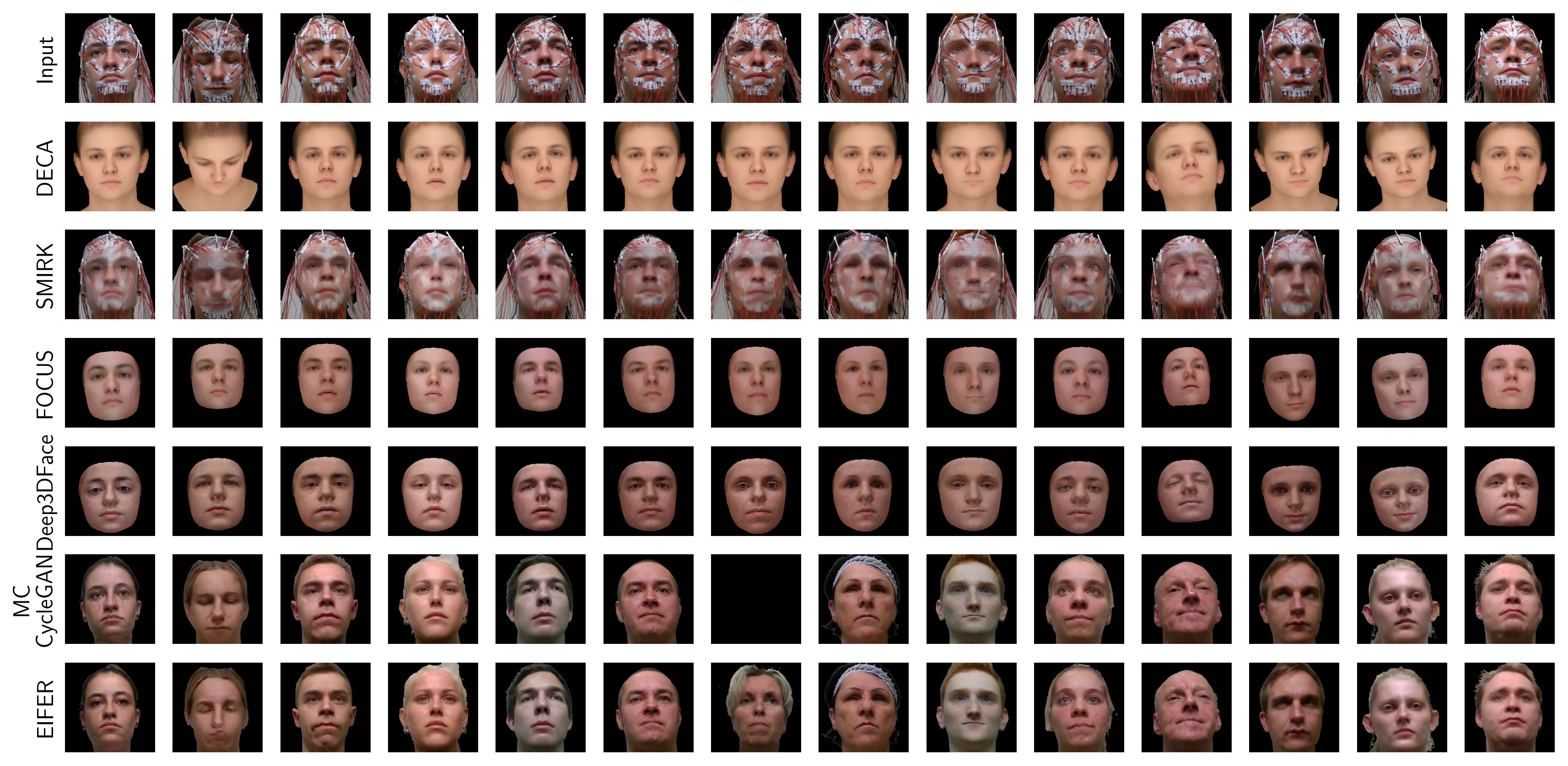}
    \caption{
        Facial Geometry Reconstruction during \textbf{Face-At-Rest}
    }
    \label{fig:app:syn:reco:face-at-rest}
\end{figure*}

\begin{figure*}[h]
    \centering
    \includegraphics[width=\linewidth]{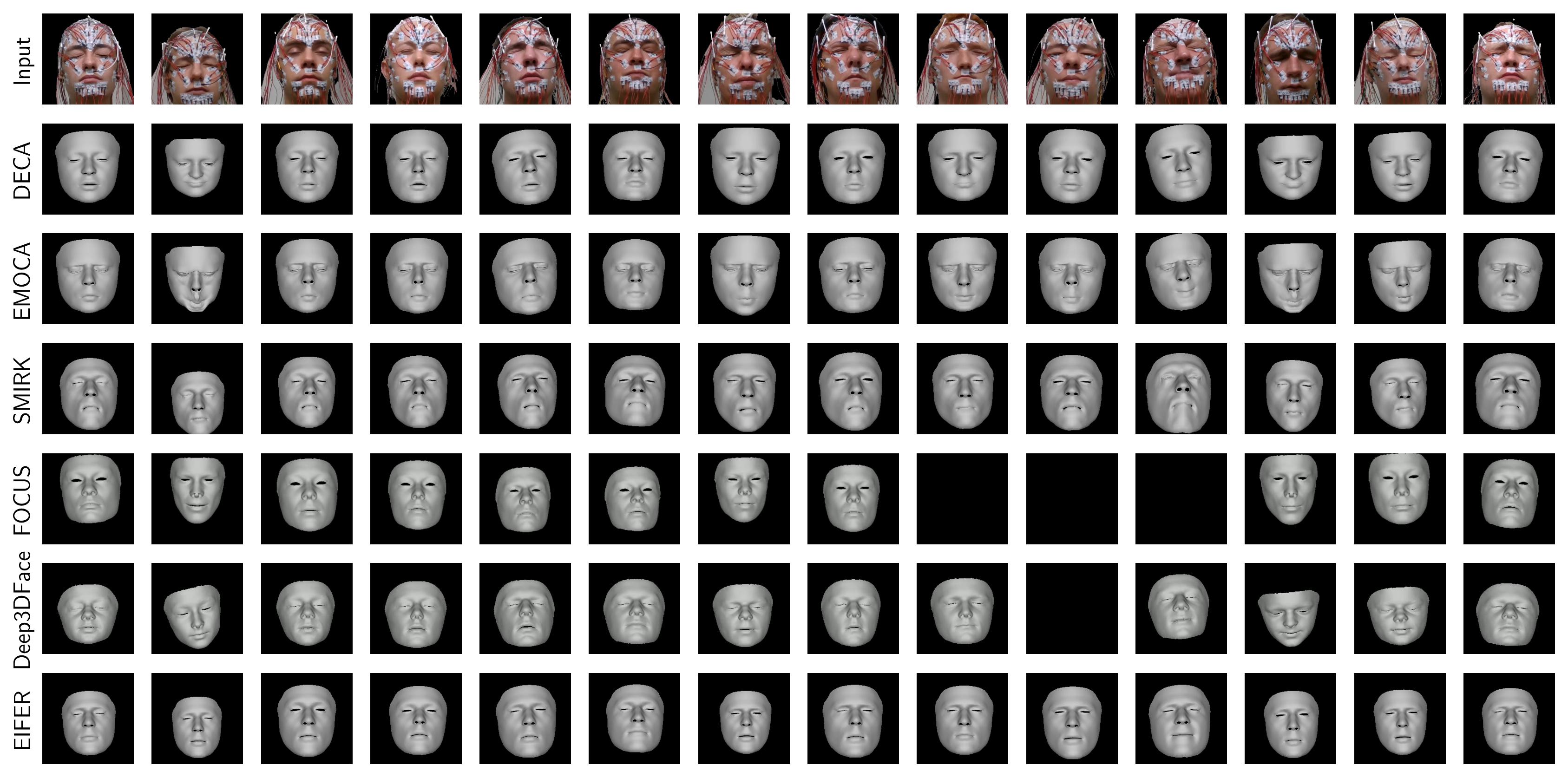}
    \caption{
        Facial Geometry Reconstruction during \textbf{Eye-Tight}
    }
    \label{fig:app:syn:shape:eye-tight}
\end{figure*}

\begin{figure*}[h]
    \centering
    \includegraphics[width=\linewidth]{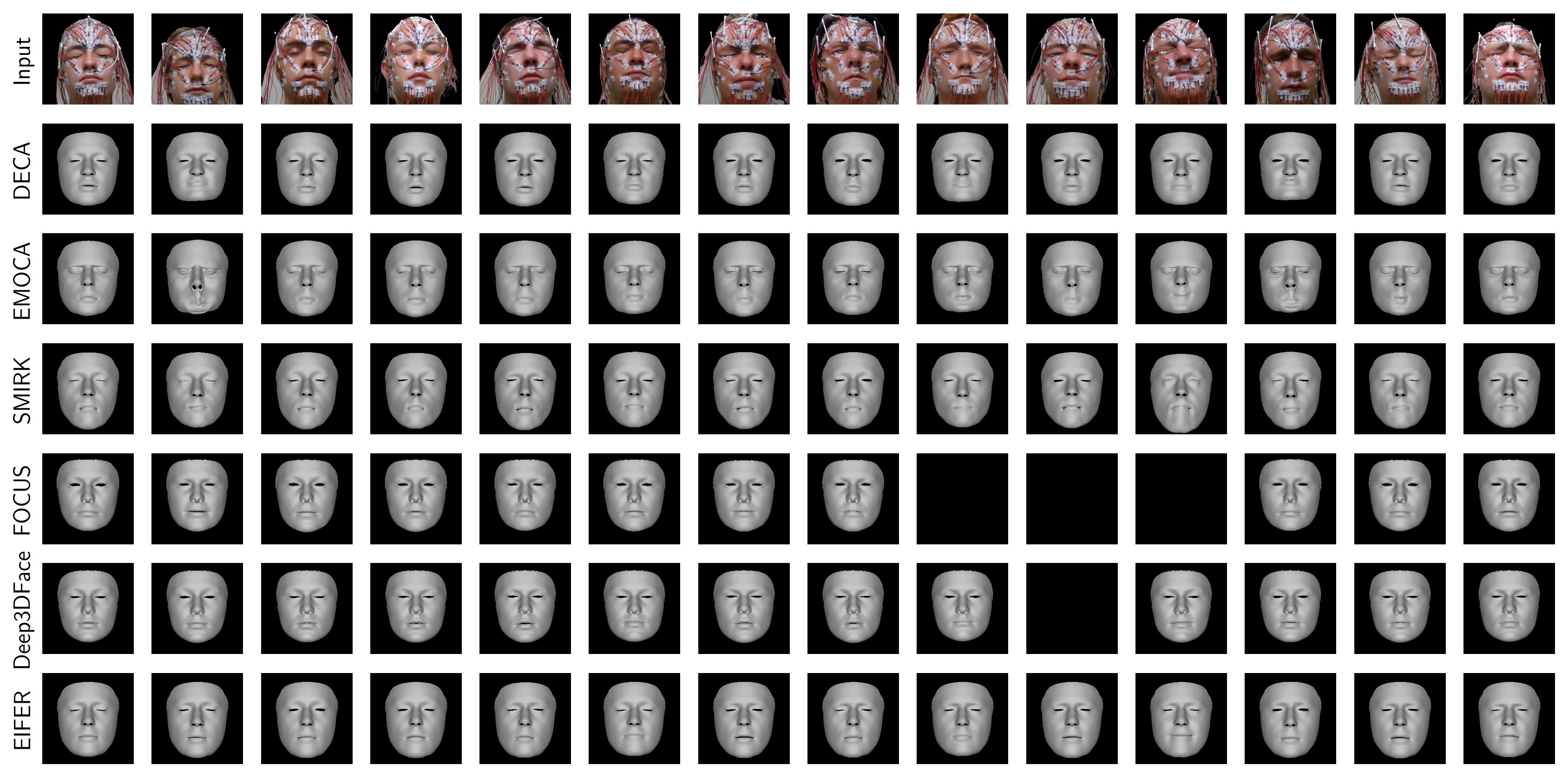}
    \caption{
        Isolated Facial Expression Reconstruction during \textbf{Eye-Tight}
    }
    \label{fig:app:syn:emo:eye-tight}
\end{figure*}

\begin{figure*}[h]
    \centering
    \includegraphics[width=\linewidth]{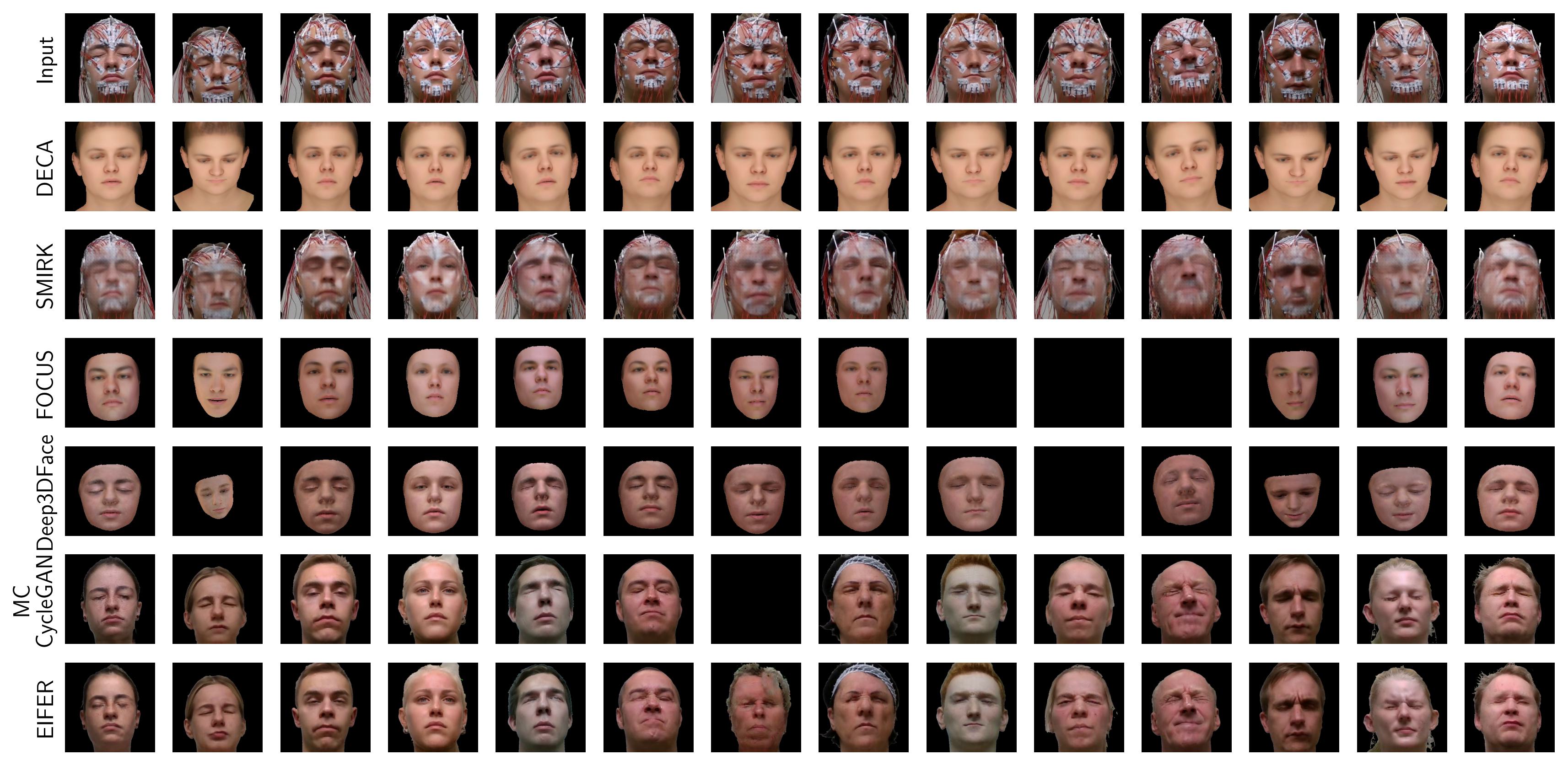}
    \caption{
        Facial Geometry Reconstruction during \textbf{Eye-Tight}
    }
    \label{fig:app:syn:reco:eye-tight}
\end{figure*}

\begin{figure*}[h]
    \centering
    \includegraphics[width=\linewidth]{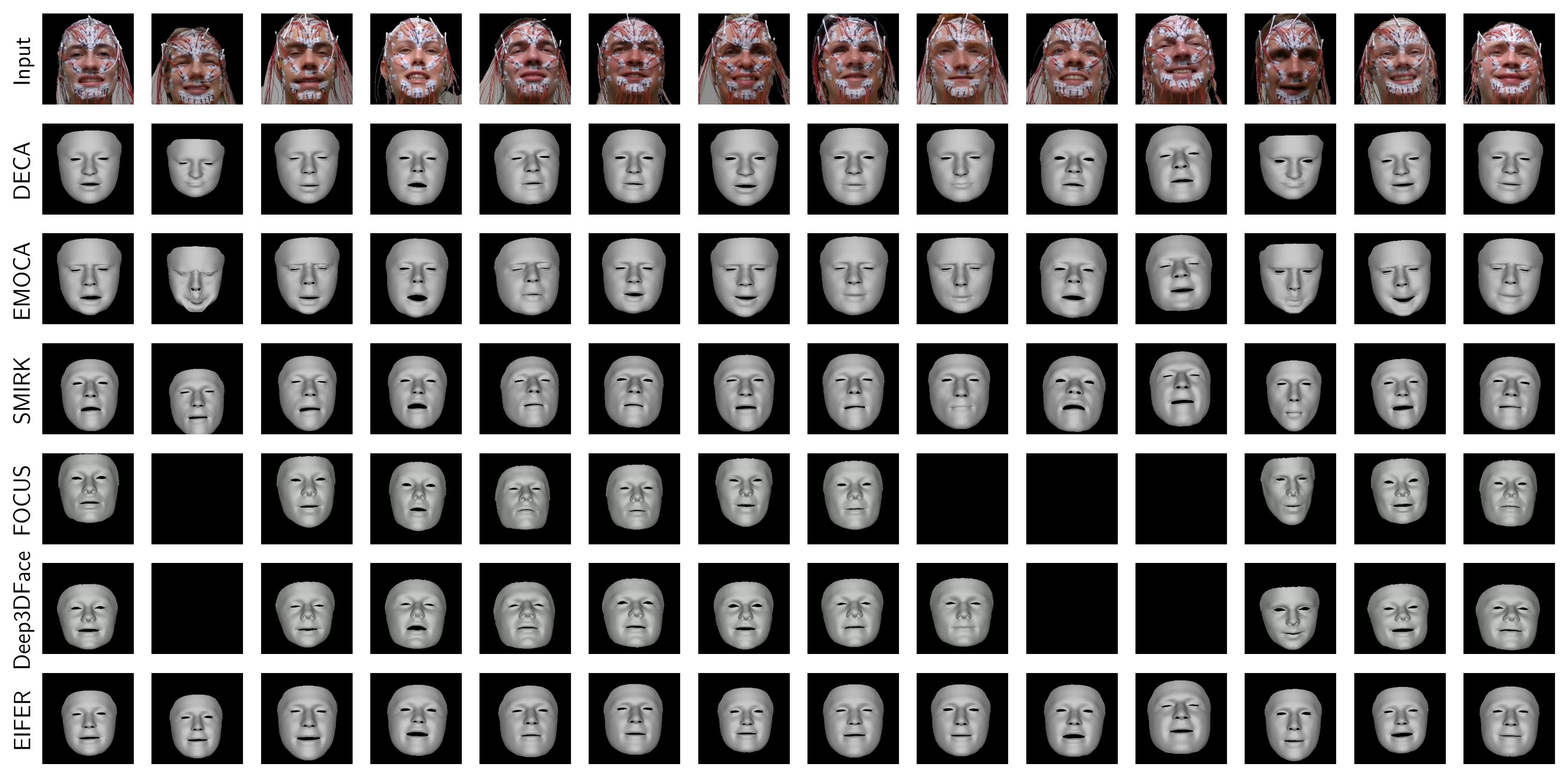}
    \caption{
        Facial Geometry Reconstruction during \textbf{Smile-Open}
    }
    \label{fig:app:syn:shape:smile-open}
\end{figure*}

\begin{figure*}[h]
    \centering
    \includegraphics[width=\linewidth]{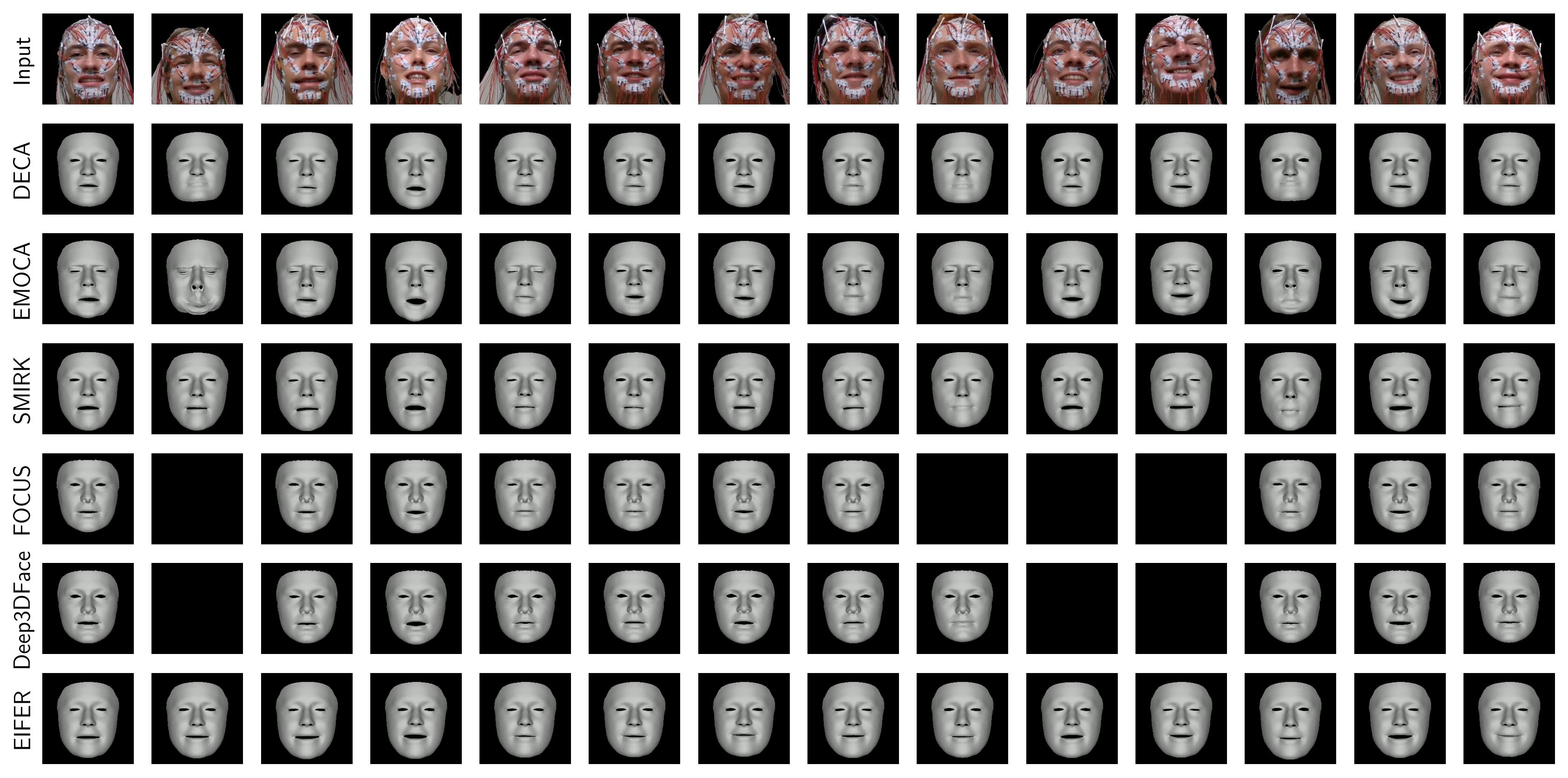}
    \caption{
        Isolated Facial Expression Reconstruction during \textbf{Smile-Open}
    }
    \label{fig:app:syn:emo:smile-open}
\end{figure*}

\begin{figure*}[h]
    \centering
    \includegraphics[width=\linewidth]{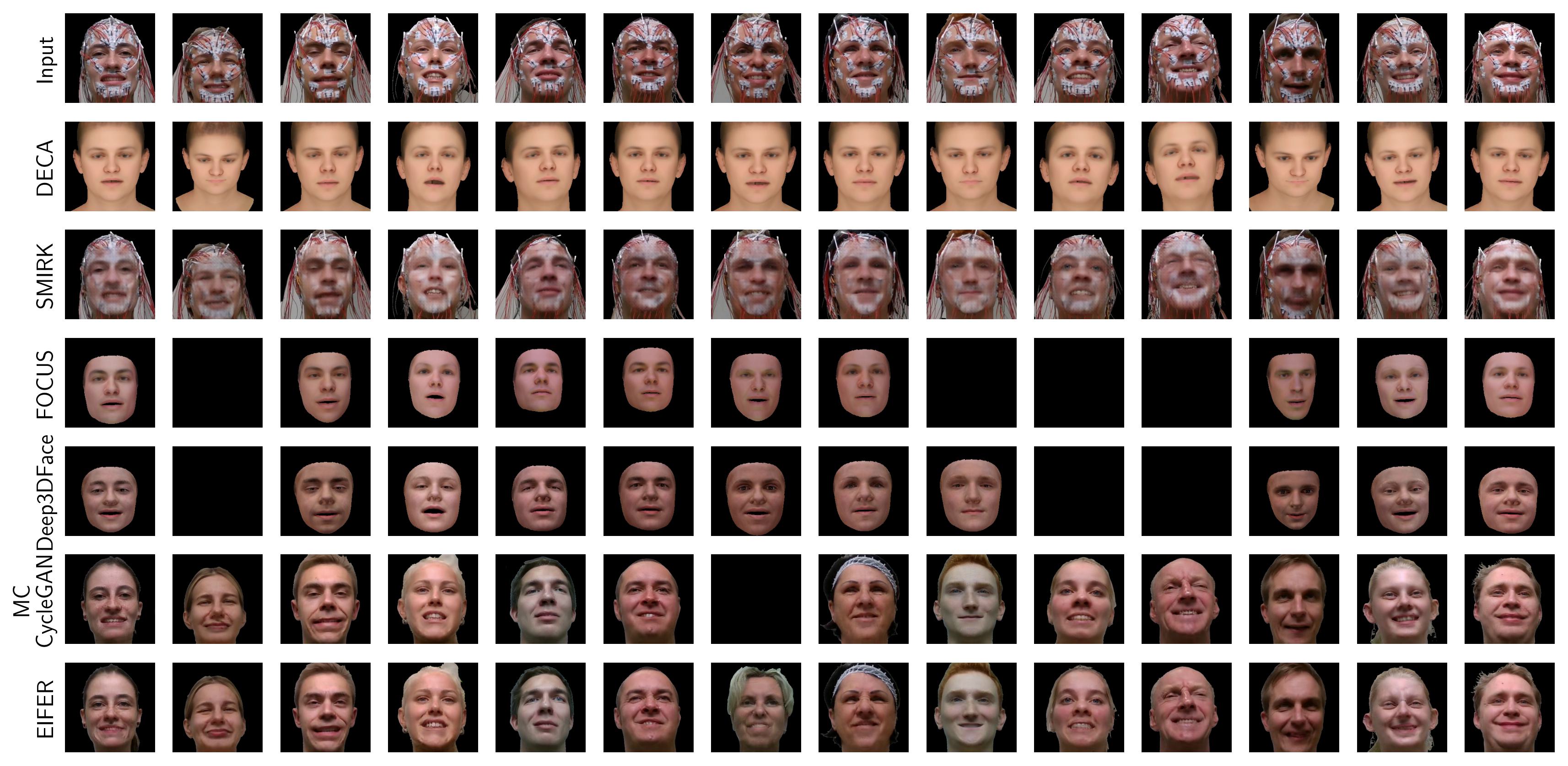}
    \caption{
        Facial Geometrie Reconstruction during \textbf{Smile-Open}
    }
    \label{fig:app:syn:reco:smile-open}
\end{figure*}

\begin{figure*}[h]
    \centering
    \includegraphics[width=\linewidth]{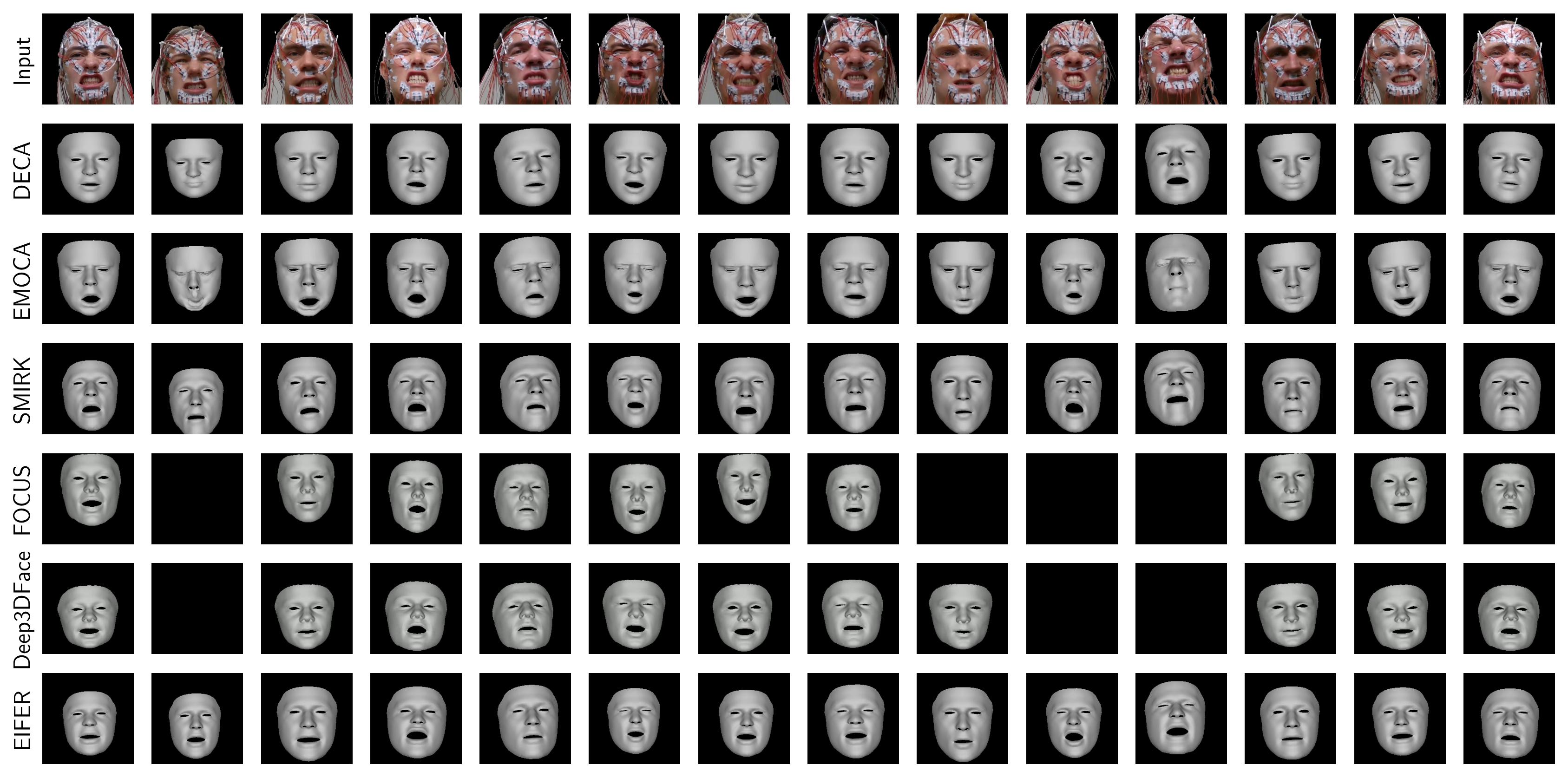}
    \caption{
        Facial Geometry Reconstruction during \textbf{Snarl}
    }
    \label{fig:app:syn:shape:snarl}
\end{figure*}

\begin{figure*}[h]
    \centering
    \includegraphics[width=\linewidth]{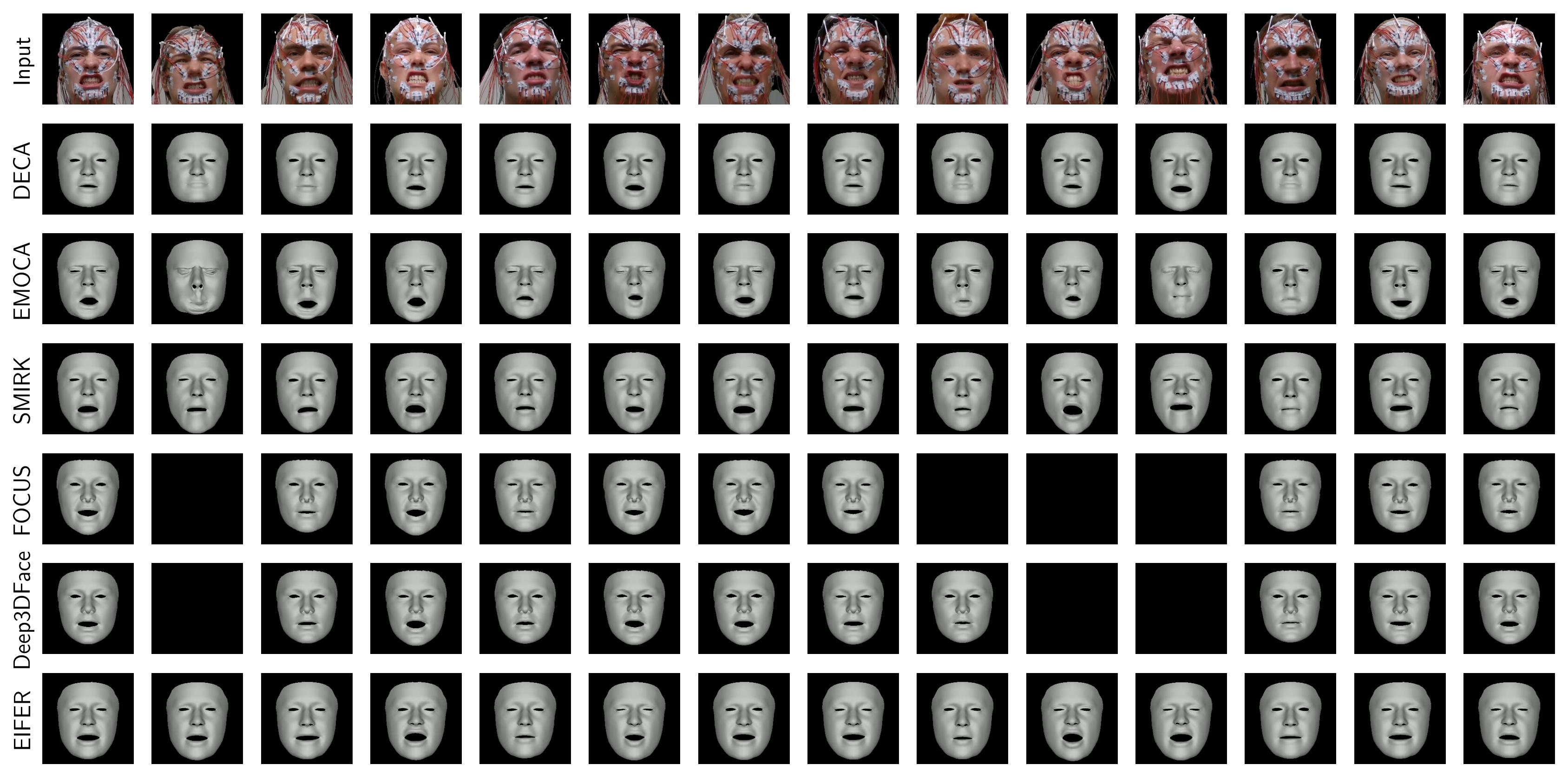}
    \caption{
        Isolated Facial Expression Reconstruction during \textbf{Snarl}
    }
    \label{fig:app:syn:emo:snarl}
\end{figure*}

\begin{figure*}[h]
    \centering
    \includegraphics[width=\linewidth]{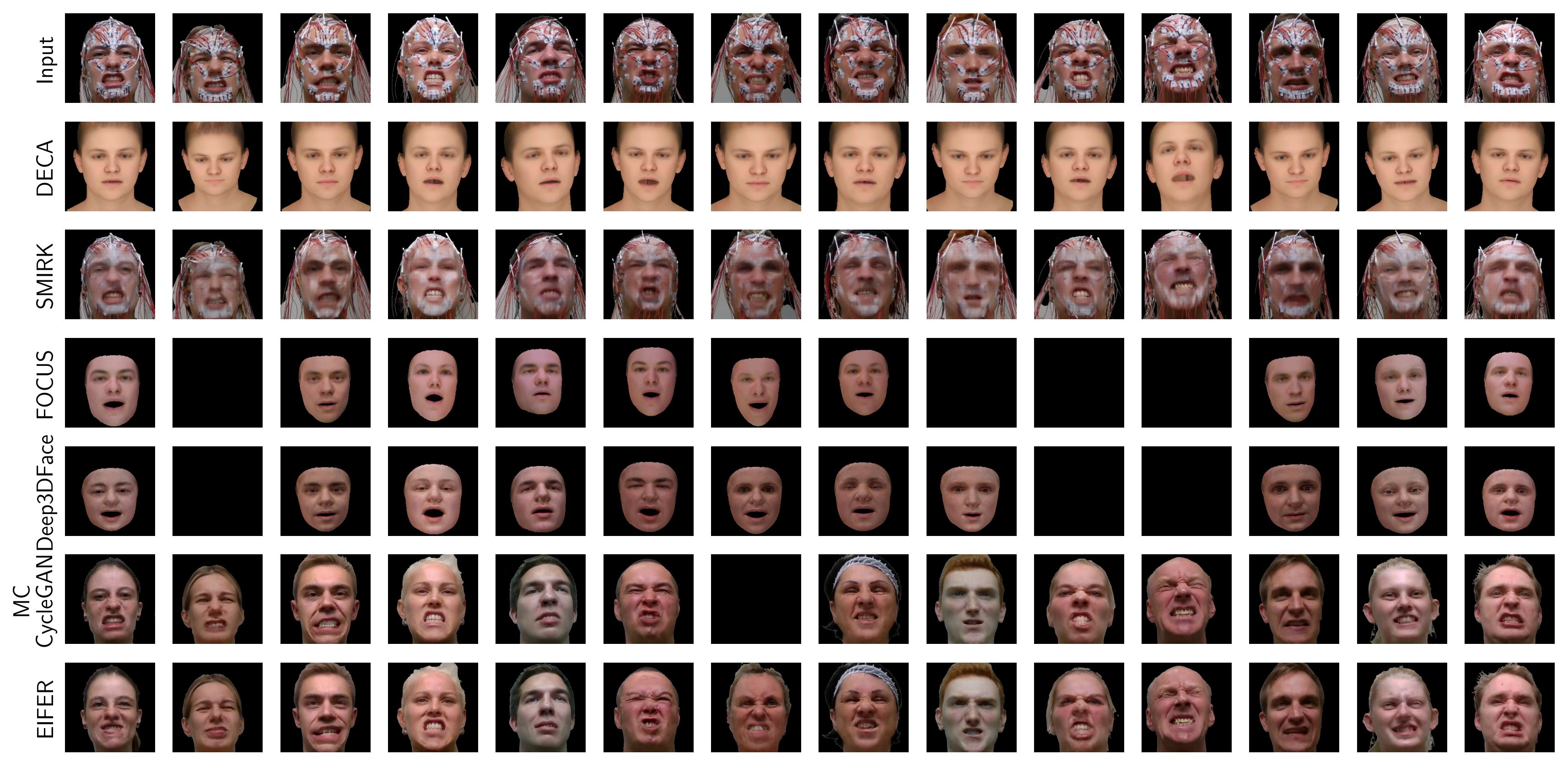}
    \caption{
        Facial Geometry Reconstruction during \textbf{Snarl}
    }
    \label{fig:app:syn:reco:snarl}
\end{figure*}

\begin{figure*}[h]
    \centering
    \includegraphics[width=\linewidth]{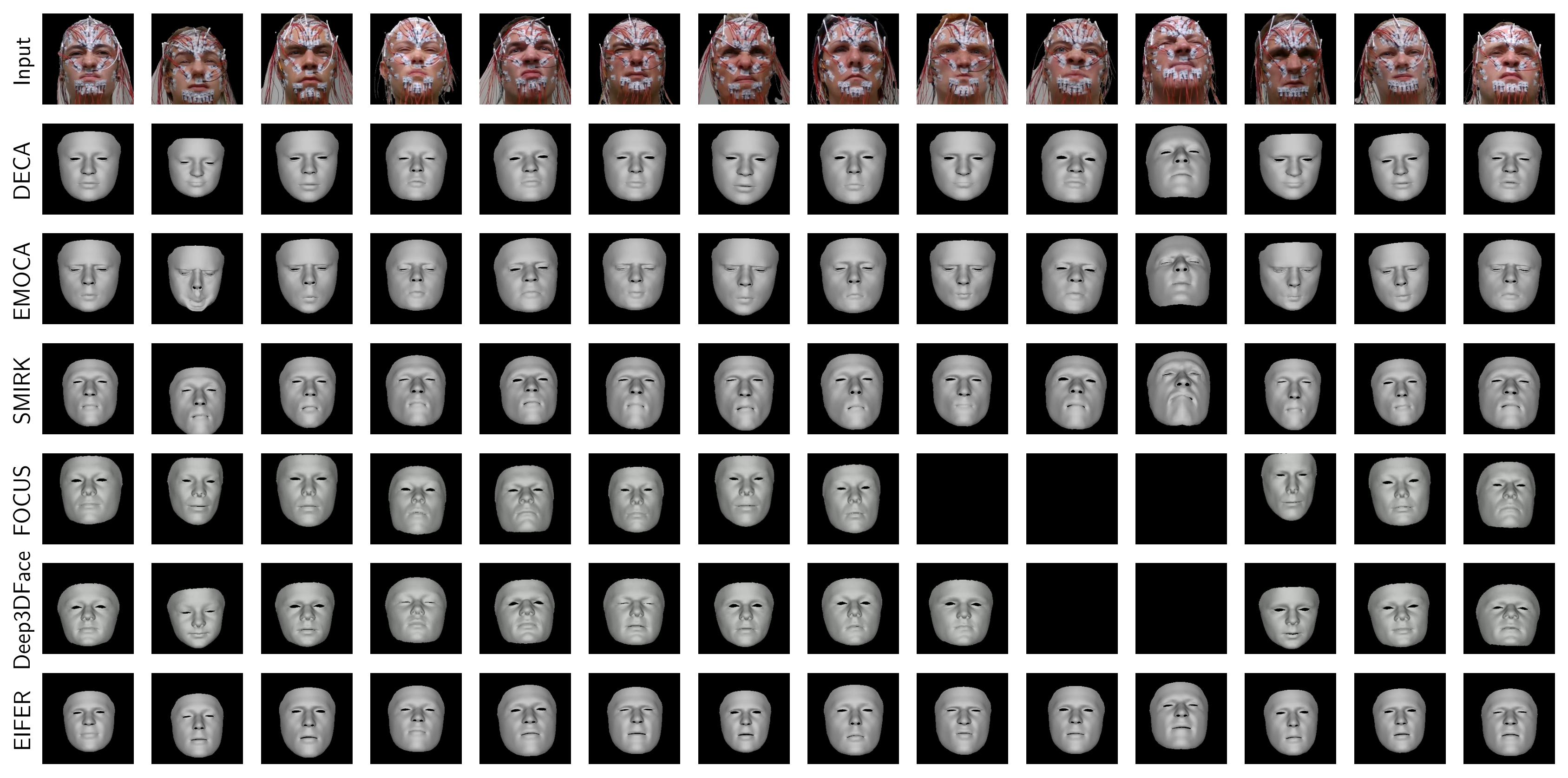}
    \caption{
        Facial Geometry Reconstruction during \textbf{Nose-Wrinkler}
    }
    \label{fig:app:syn:shape:nose}
\end{figure*}

\begin{figure*}[h]
    \centering
    \includegraphics[width=\linewidth]{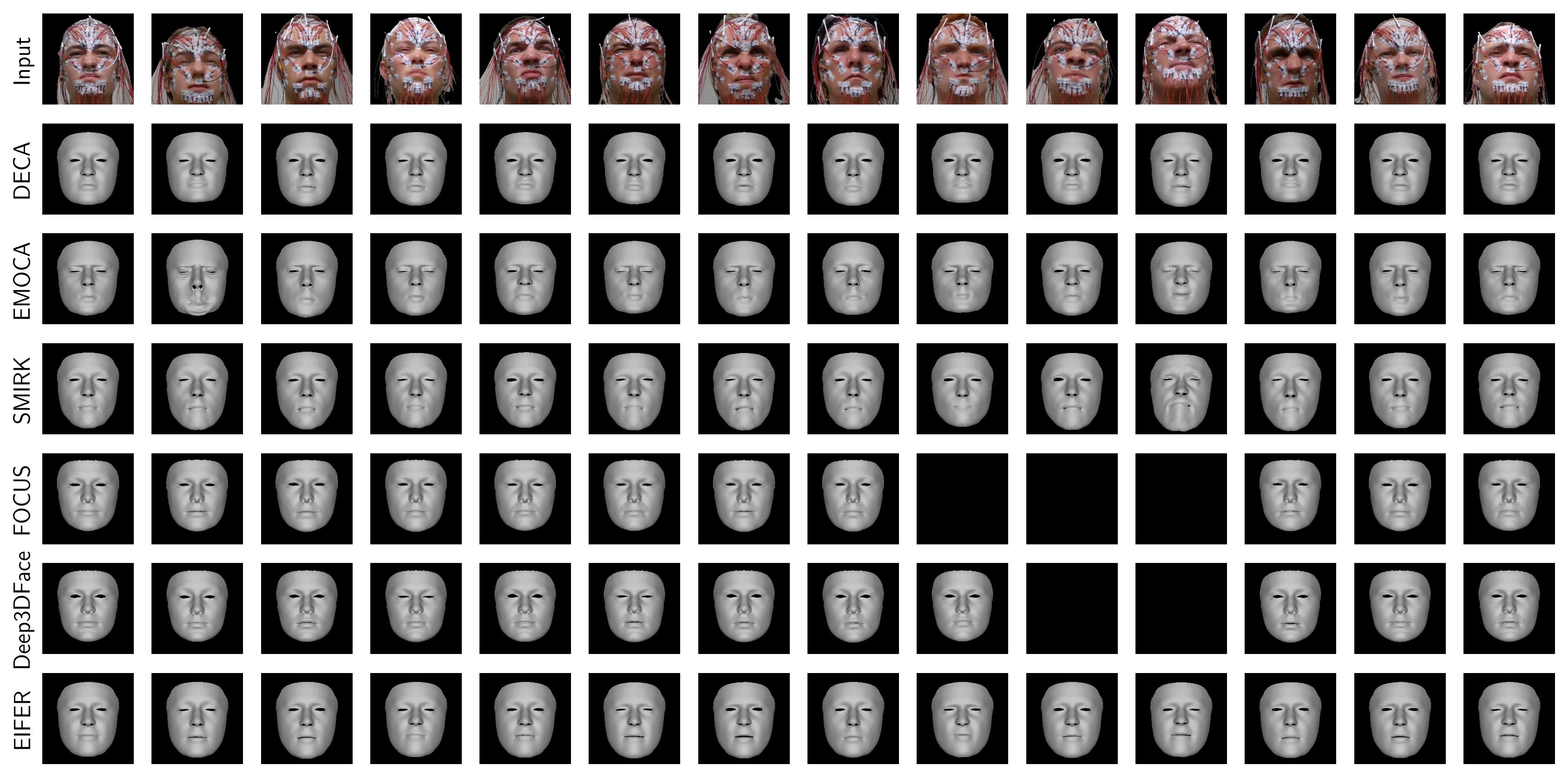}
    \caption{
        Isolated Facial Expression Reconstruction during \textbf{Nose-Wrinkler}
    }
    \label{fig:app:syn:emo:nose}
\end{figure*}

\begin{figure*}[h]
    \centering
    \includegraphics[width=\linewidth]{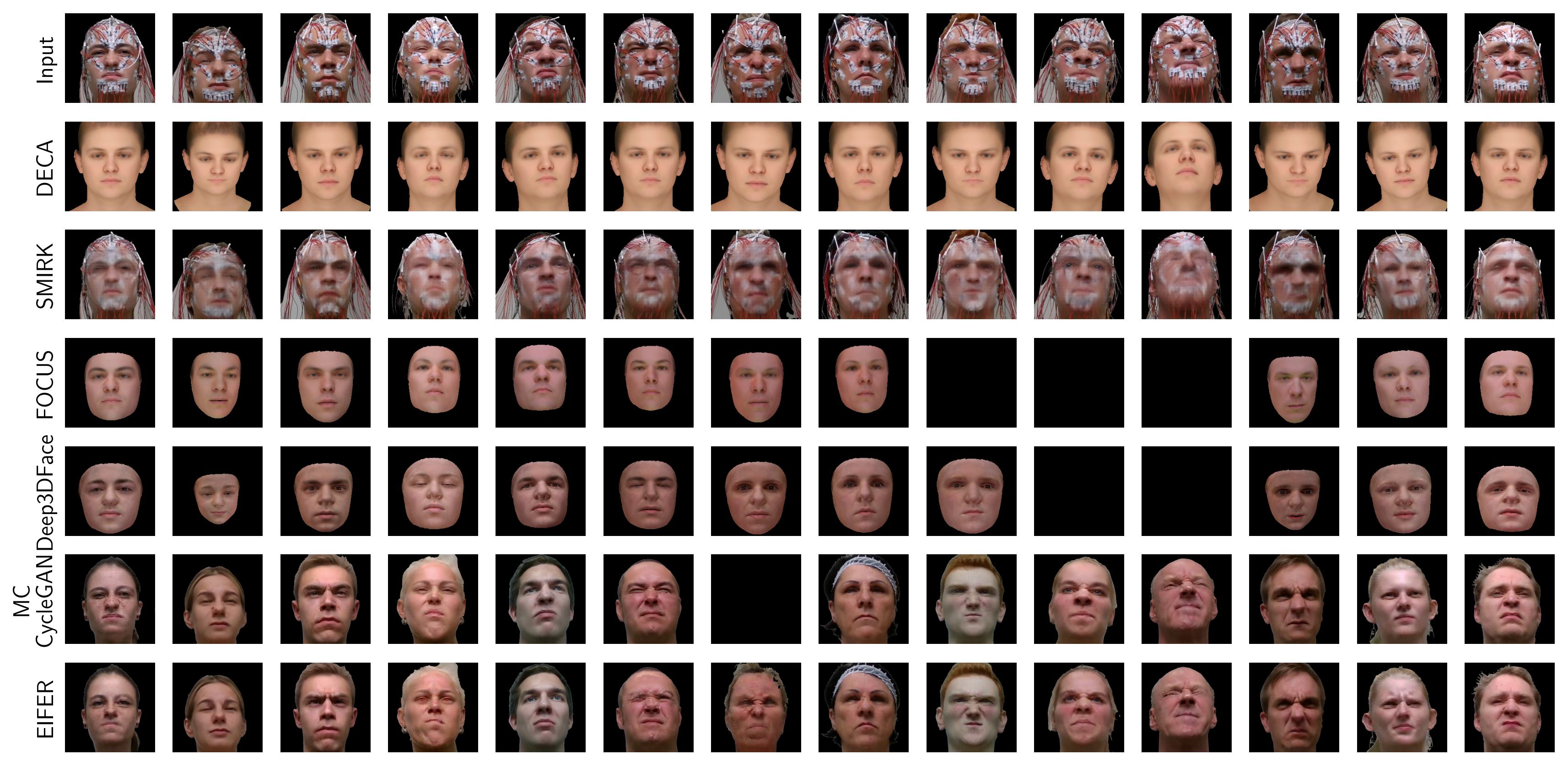}
    \caption{
        Facial Geometry Reconstruction during \textbf{Nose-Wrinkler}
    }
    \label{fig:app:syn:reco:nose}
\end{figure*}

We provide additional visual examples for facial geometry extraction and appearance reconstruction.
We also demonstrate the reconstruction using only expression parameters on a neutral face to evaluate the encoder's disentanglement ability during sEMG occlusion.
Examples include Face-At-Rest, Eye-Tight, Smile-Open, Snarl, and Nose-Wrinkler.
These are the following figures:
\begin{itemize}
    \item Face-At-Rest: 3D Geometry \Cref{fig:app:syn:shape:face-at-rest}
    \item Face-At-Rest: Isolated Expression \Cref{fig:app:syn:emo:face-at-rest}
    \item Face-At-Rest: Appearance Reconstruction \Cref{fig:app:syn:reco:face-at-rest}
    
    \item Eye-Tight: 3D Geometry \Cref{fig:app:syn:shape:eye-tight}
    \item Eye-Tight: Isolated Expression \Cref{fig:app:syn:emo:eye-tight}
    \item Eye-Tight: Appearance Reconstruction \Cref{fig:app:syn:reco:eye-tight}

    \item Smile-Open: 3D Geometry \Cref{fig:app:syn:shape:smile-open}
    \item Smile-Open: Isolated Expression \Cref{fig:app:syn:emo:smile-open}
    \item Smile-Open: Appearance Reconstruction \Cref{fig:app:syn:reco:smile-open}

    \item Snarl: 3D Geometry \Cref{fig:app:syn:shape:snarl}
    \item Snarl: Isolated Expression \Cref{fig:app:syn:emo:snarl}
    \item Snarl: Appearance Reconstruction \Cref{fig:app:syn:reco:snarl}

    \item Nose-Wrinkler: 3D Geometry \Cref{fig:app:syn:shape:nose}
    \item Nose-Wrinkler: Isolated Expression \Cref{fig:app:syn:emo:nose}
    \item Nose-Wrinkler: Appearance Reconstruction \Cref{fig:app:syn:reco:nose}
\end{itemize}

\subsection{EMG2Exp}
\begin{figure*}[ht]
    \centering
    \includegraphics[width=0.95\linewidth]{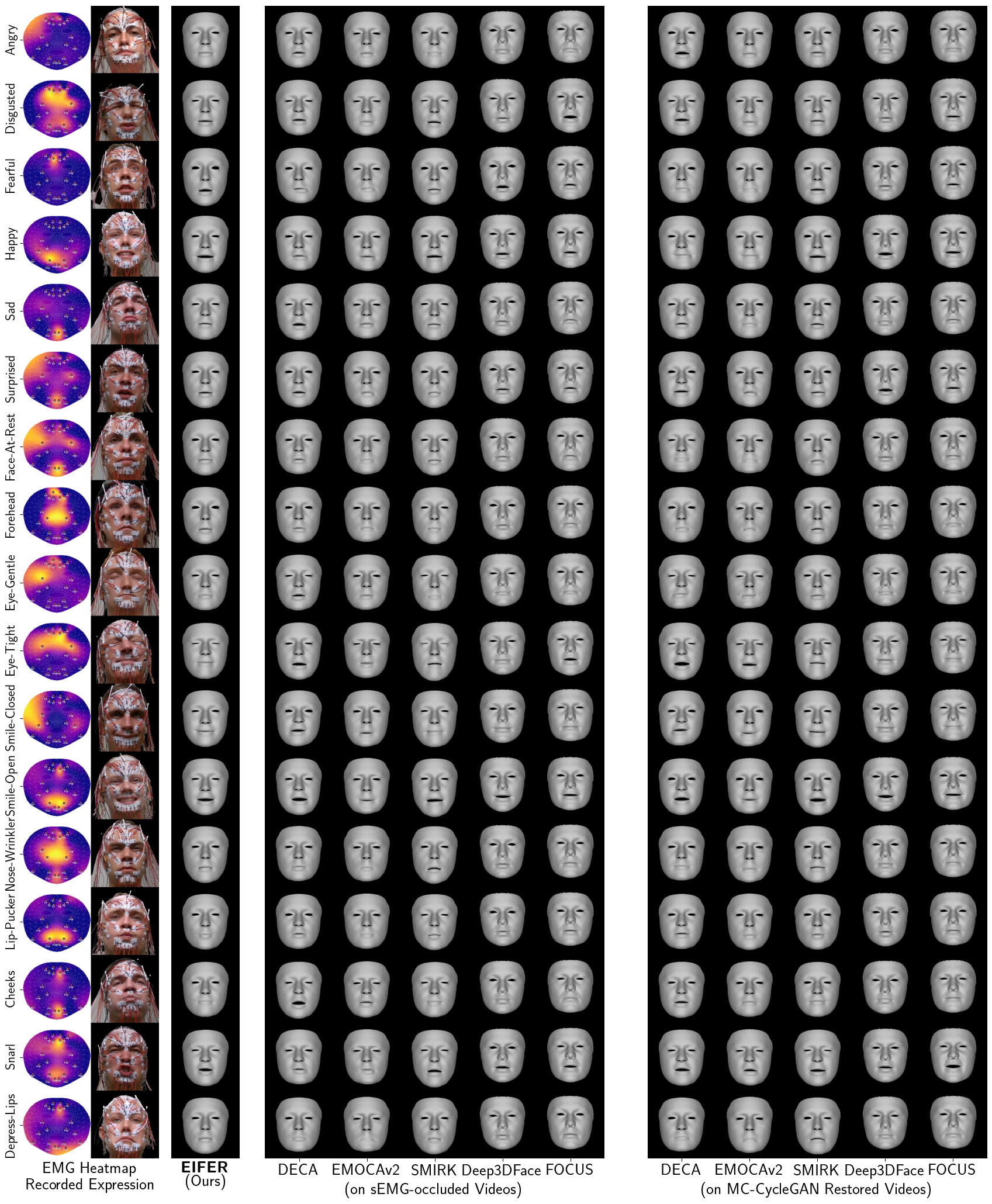}
    \caption{
        \textbf{Physiological-based Expression Synthesis via Muscle Activity:}
        We demonstrate synthesized facial expressions from recorded muscle activity.
        State-of-the-art methods, such as SMIRK and FOCUS, struggle to reconstruct expressions under sEMG occlusion.
        We see improved results on the MC-CycleGAN~\protect\citeSup{buchner2023let_supRef, buchner2023improved_supRef} restored faces, but only SMIRK performs well across all emotions.
        In contrast, our method, EIFER, achieves comparable synthesis quality directly from occluded images without needing electrode removal.
    }
    \label{fig:app:synthetic}
\end{figure*}

We provide additional visual examples of synthesized facial expressions based on muscle activity for all methods, including the six base emotions and eleven functional movements for more participants, as shown in \Cref{fig:app:synthetic}.
We also compare the results using MC-CycleGAN~\protect\citeSup{buchner2023let_supRef, buchner2023improved_supRef} restored recordings for a fair comparison.

Our method directly generates highly realistic faces from occluded faces, whereas other methods require occlusion-free faces.
This demonstrates the robustness of EIFER in handling sEMG occlusion. However, SMIRK~\protect\citeSup{smirk_supRef} is the only method to reconstruct the \textit{Depress-Lip} movement, demonstrating its ability to encode rare and subtle facial expressions.
In contrast, EIFER could not learn this movement, even under occlusion, highlighting a potential area for improvement.

We observe an interesting phenomenon where the model can synthesize the \textit{Eye-Tight} movement but not the \textit{Eye-Gentle} movement.
This suggests that the model can pick up on different muscular patterns depending on the strength of the same movement.
However, it remains unclear whether the differences between voluntary and enforced movements exhibit similar patterns.
Notably, EIFER is the only method that can restore the \textit{Lip-Pucker} movement.

We also observe that the jaw movement is challenging to learn, as the \textit{M. massester} muscle is only slightly active during jaw opening.
Although this task is easy to solve visually, the muscle activity appears insufficient.
Furthermore, we find that the performance of the two 3DMMs (FLAME~\protect\citeSup{FLAME_supRef} and BaselFaceModel~\protect\citeSup{bfm1_supRef, bfm2_supRef}) depends on the encoder model.
This suggests that a well-trained encoder model is more important than the capabilities of the 3DMM expression space.

This finding highlights the importance of disentanglement of shape and expression in 3DMMs, as well as the significance of the encoder model~\protect\citeSup{eggerIdentityExpressionAmbiguity3D2021_supRef, 3dmmpastpresentfuture_supRef}.
Although this task remains ill-posed, our results have implications for new research directions in medicine and psychology.

\subsection{Exp2EMG}
\begin{figure*}[h]
    \centering
    \includegraphics[width=1.0\linewidth]{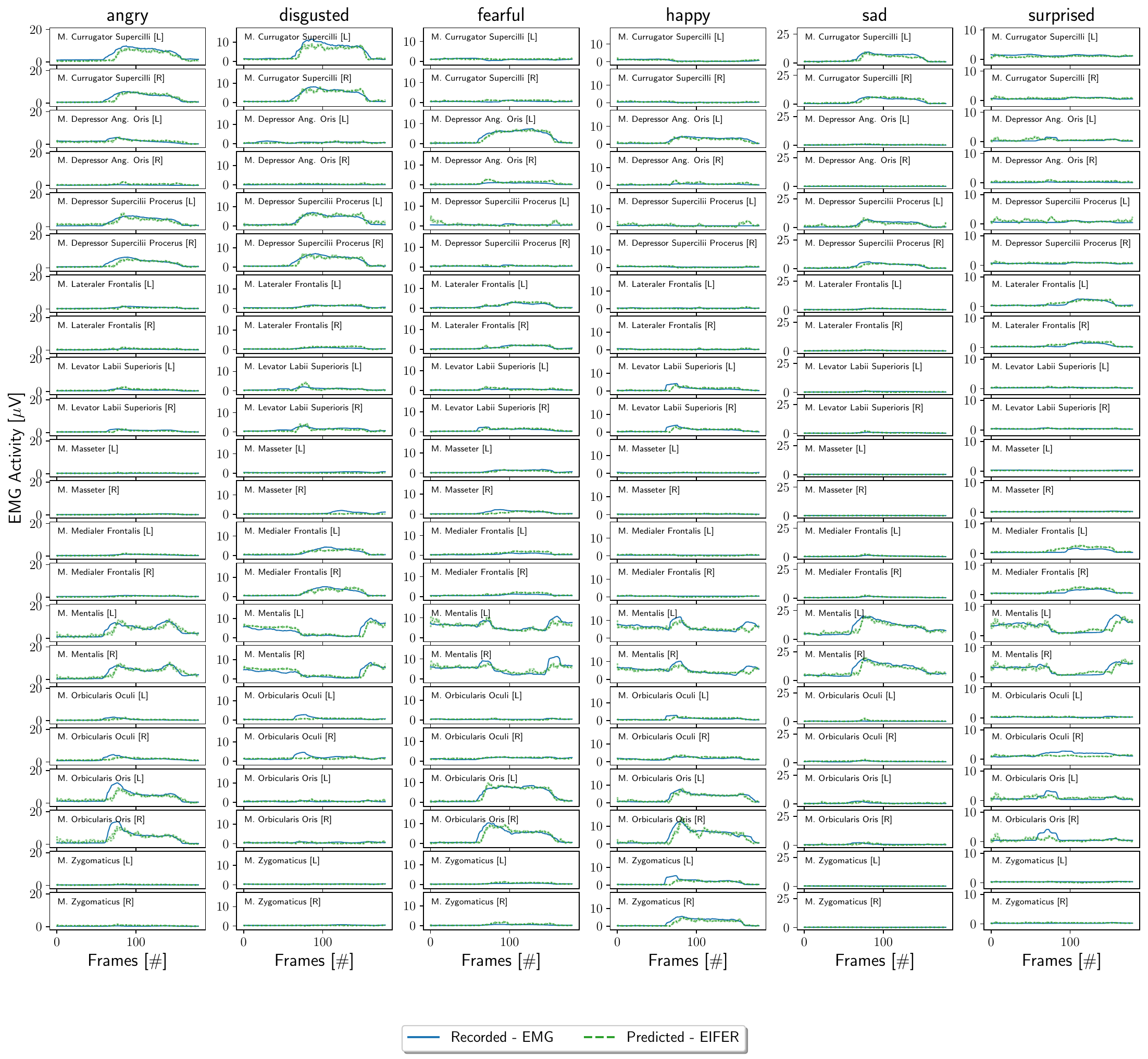}
    \caption{
        \textbf{Muscle Activity via Expression Parameters}
        We demonstrate the reconstruction of muscle activity from expression parameters, achieving fair results with minor amplitude signal issues.
        We visualize this capability for the six base emotions~\protect\citeSup{ekmanArgumentBasicEmotions1992_supRef}.
    }
    \label{fig:app:ana:emotion}
\end{figure*}

\begin{figure*}[h]
    \centering
    \includegraphics[width=\linewidth]{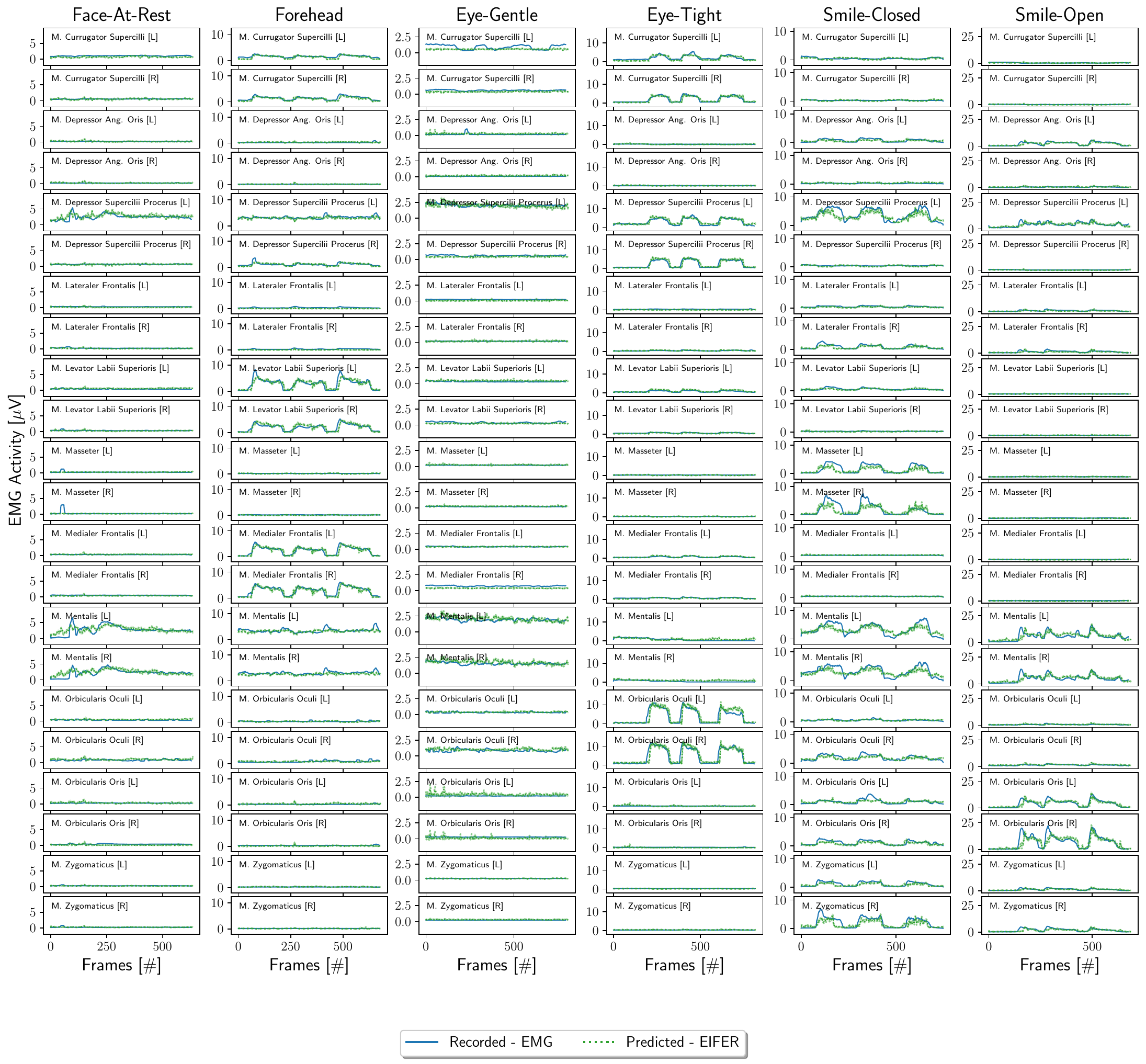}
    \caption{
        \textbf{Muscle Activity via Expression Parameters}
        We demonstrate the reconstruction of muscle activity from expression parameters, achieving fair results with minor amplitude signal issues.
        We visualize this capability for the six different functional movements~\protect\citeSup{schaede_supRef}.
    }
    \label{fig:app:ana:schaede1}
\end{figure*}

\begin{figure*}[h]
    \centering
    \includegraphics[width=\linewidth]{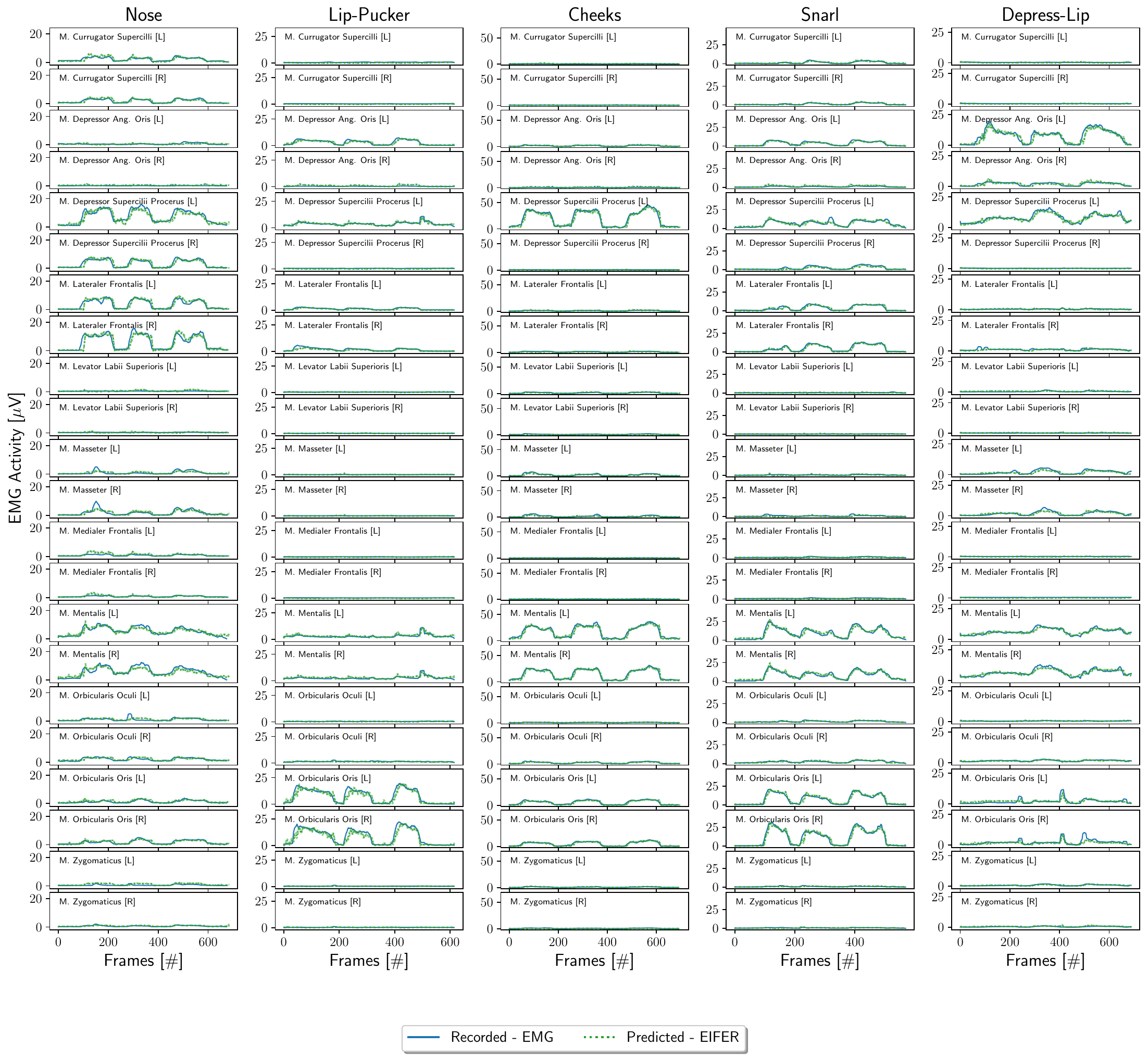}
    \caption{
    \textbf{Muscle Activity via Expression Parameters}
        We demonstrate the reconstruction of muscle activity from expression parameters, achieving fair results with minor amplitude signal issues.
        We visualize this capability for the remaining five functional movements~\protect\citeSup{schaede_supRef}.
    }
    \label{fig:app:ana:schaede2}
\end{figure*}

We provide additional examples of EIFER's muscle activity prediction beyond the single active and inactive muscle visualized in the main paper for the \textit{happy} expression.
These can be found in \Cref{fig:app:ana:emotion}, \Cref{fig:app:ana:schaede1}, and \Cref{fig:app:ana:schaede2}.
Certain muscles are typically active during specific facial expressions, while others remain inactive.
However, we also notice decreased activity in some muscles, accompanied by activation in others.
This phenomenon, which is not well-studied~\protect\citeSup{emg1_supRef}, suggests that facial muscles may be more interconnected than currently assumed~\protect\citeSup{ekmanArgumentBasicEmotions1992_supRef, ekmanFacialActionCoding1978_paper_supRef}, warranting further investigation.

EIFER can accurately predict the muscle activity envelope without requiring additional personal information
However, we refine the prediction by multiplying it by the participant's maximum observed activity (in $\mu V$), allowing us to estimate the relative activity and actual muscle strength.
Even without this refinement, EIFER remains a powerful tool for predicting muscle activity.

We observe that EIFER accurately fits the shape of the original signal in all reconstructions but occasionally struggles to estimate the signal amplitude correctly.
We attribute this to the per-participant normalization during training, which may cause the model to underestimate the general signal amplitude if participants require varying levels of muscle activity to evoke changes in facial mimicry.

Several potential reasons for this phenomenon deserve further exploration:
\begin{enumerate}
 \item Are there differences in voluntary and evoked expression patterns?
 \item Do participants exhibit unique muscle activity patterns for certain expressions due to pathological conditions?
 \item Are there learning effects between sessions, such as changes in reaction time, execution speed, or intensity?
\end{enumerate}

To drive progress in understanding and addressing these open questions, we are releasing our models \textit{EMG2Exp} and \textit{Exp2EMG} to the research community, inviting collaboration and exploration to uncover the underlying causes of these phenomena and push the boundaries of facial expression analysis.








\begin{figure*}
    \centering
    \includegraphics[width=0.94\linewidth]{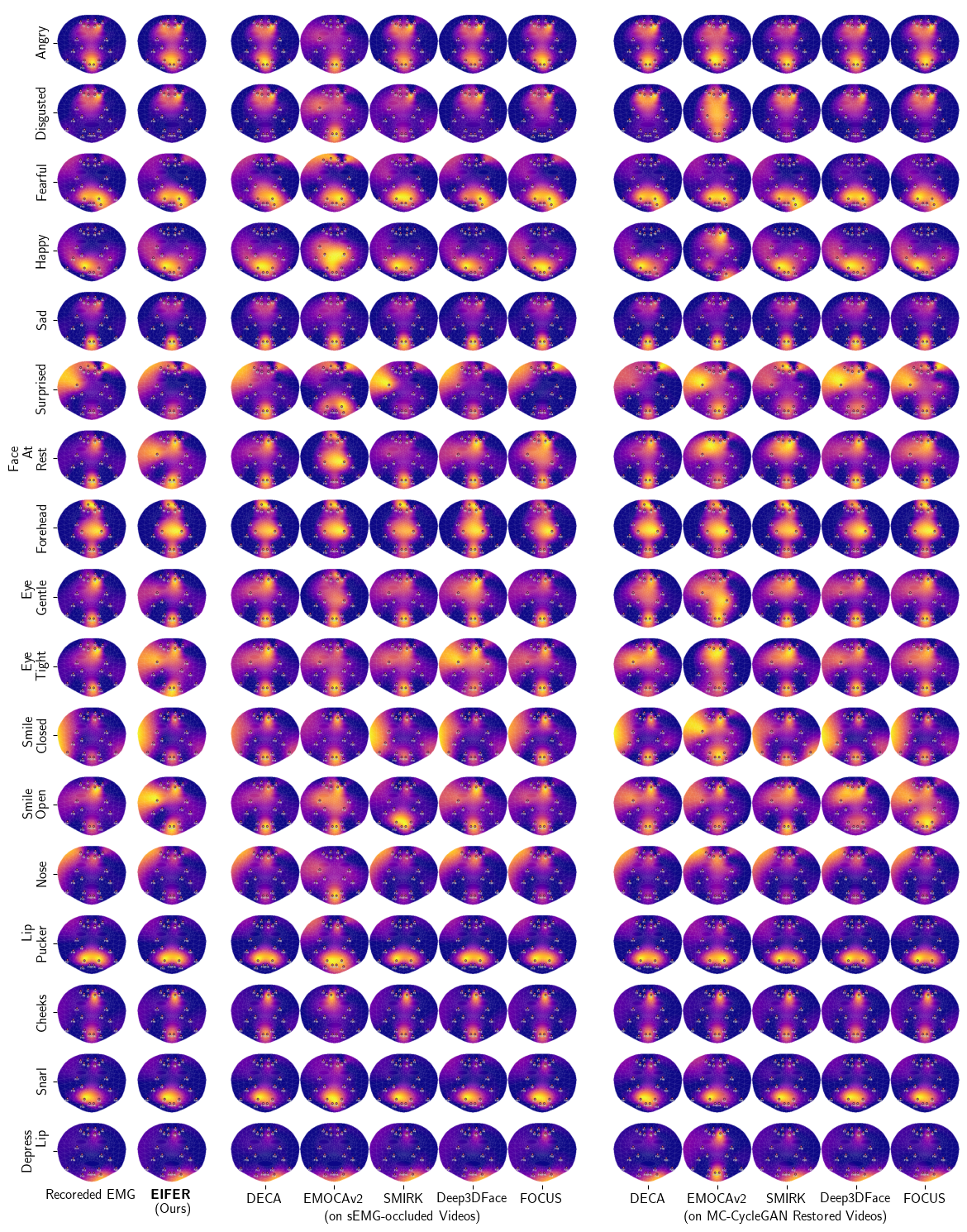}
    \caption{
        \textbf{Topological EMG Heatmaps:}
        We compare the topological heat maps during the peak muscle measurement of muscle activity for each movement.
        Further, we display the predicted muscle activity based on each method.
        SI units are committed for clarity but can be taken from the other muscle activity prediction figures (\Cref{fig:app:ana:emotion}, \Cref{fig:app:ana:schaede1}, \Cref{fig:app:ana:schaede2}).
    }
    \label{fig:app:ana:heatmap}
\end{figure*}

\FloatBarrier
\clearpage

\section{Ablation Studies}
In addition to the results presented in the main paper, we conduct further ablation studies to explore alternative applications of EIFER.
We investigate various downstream tasks to assess the model's versatility and potential uses beyond its original purpose.

\subsection{Convolutional Based Expression Classification}
\label{sup:sec:fer}
We leverage the six base emotions~\protect\citeSup{ekmanArgumentBasicEmotions1992_supRef} mimicked by our participants as reference annotations, serving as ground truth labels for image-based classification.
This way, we evaluate whether the appearance reconstruction accurately resembles the target facial expression, providing an additional objective criterion for assessing appearance quality.

We employ several convolution-based Facial Expression Recognition (FER) classifiers: Poster++~\protect\citeSup{mao2023poster_supRef}, ResidualMaskingNet~\protect\citeSup{luanresmaskingnet2020_supRef}, EmoNext~\protect\citeSup{elboudouriEmoNeXtAdaptedConvNeXt2023_supRef}, and SegmentationVGG~\protect\citeSup{segmetnationvgg_supRef}.
All required preprocessing steps are strongly followed, as outlined in the paper and corresponding code repositories.
While this model selection is not exhaustive, it gives a broad overview of existing classifiers trained on public datasets.
However, since we cannot directly assess the accuracy of the mimicked expression, we establish two baselines:
(1) an upper baseline using occlusion-free reference recordings, and
(2) a lower threshold using sEMG-occluded recordings.
Any model should perform better than the lower baseline.

We present the results in \Cref{tab:app:poster} to \Cref{tab:app:vgg}.
Notably, none of the reconstruction methods achieve the original upper limit on the occlusion-free videos. 
This discrepancy may be attributed to several factors: First, the methods may struggle with frames that differ from the training database's image quality and recording style, simulating a distribution shift or in-the-wild application scenario.
Second, the appearance reconstruction may introduce biases invisible to the human eye but affect the models' performance~\protect\citeSup{buchner2024power_supRef, buchner2024facing_supRef}.
Lastly, the reconstruction may not retain the essential facial features that the models rely on, indicating potential information loss during the reconstruction.
While the underlying cause is beyond the scope of this study, it is an essential area of research that can help uncover the black-box nature of FER classification models.

\begin{table}[h]
    \centering
    \resizebox{\linewidth}{!}{%
    \begin{tabular}{l|rrrrrr|r}
	\toprule
	                & Angry & Disgusted & Fearful & Happy & Sad   & Surprised & Average \\
	\midrule
	Upper Limit (N) & 65.97 & 82.29     & 53.12   & 94.08 & 75.00 & 70.49     & 73.49   \\
	Lower Limit (S) & 14.02 & 15.91     & 46.02   & 54.92 & 84.47 & 60.04     & 45.90   \\
	\midrule
	DECA            & 7.01  & 0.00      & 0.00    & 0.00  & 0.00  & 21.97     & 4.83    \\
	EMOCAv2         & 61.45 & 3.39      & 79.89   & 2.82  & 8.94  & 6.15      & 27.11   \\
	SMIRK           & 31.44 & 4.92      & 22.73   & 56.06 & 81.06 & 53.03     & 41.54   \\
	Deep3DFace      & 0.25  & 1.50      & 5.46    & 32.17 & 16.42 & 89.58     & 24.23   \\
	FOCUS           & 0.26  & 0.00      & 0.00    & 1.56  & 0.00  & 92.23     & 15.67   \\
	MCGAN           & 47.97 & 75.61     & 33.94   & 75.61 & 73.17 & 58.94     & 60.87   \\
	\midrule
	EIFER           & 48.67 & 61.55     & 27.84   & 71.78 & 67.23 & 56.44     & 55.59   \\
	\bottomrule
\end{tabular}%
    }
    \caption{
        \textbf{Emotion Classification Accuracy for Poster++\protect\citeSup{mao2023poster_supRef}:}
        We report the FER image-based classification results for the appearance reconstructions.
    }
    \label{tab:app:poster}
\end{table}

\begin{table}[h]
    \centering
    \resizebox{\linewidth}{!}{%
    \begin{tabular}{l|rrrrrr|r}
	\toprule
	                & Angry & Disgusted & Fearful & Happy & Sad  & Surprised & Average \\
	\midrule
	Upper Limit (N) & 56.60 & 72.57     & 36.81   & 84.67 & 8.33 & 57.29     & 52.71   \\
	Lower Limit (S) & 13.64 & 0.00      & 2.65    & 25.19 & 0.00 & 81.82     & 20.55   \\
	\midrule
	DECA            & 0.19  & 0.00      & 82.01   & 0.57  & 0.00 & 0.00      & 13.79   \\
	EMOCAv2         & 2.23  & 7.91      & 4.47    & 3.95  & 1.12 & 1.12      & 3.47    \\
	SMIRK           & 16.10 & 4.73      & 5.68    & 7.77  & 0.57 & 68.37     & 17.20   \\
	Deep3DFace      & 4.96  & 15.54     & 3.97    & 60.60 & 2.24 & 68.98     & 26.05   \\
	FOCUS           & 2.07  & 1.31      & 40.57   & 18.44 & 0.26 & 47.15     & 18.30   \\
	MCGAN           & 45.12 & 64.43     & 21.14   & 52.85 & 2.64 & 39.63     & 37.64   \\
	\midrule
	EIFER           & 48.67 & 62.69     & 14.20   & 55.11 & 3.22 & 39.39     & 37.22   \\
	\bottomrule
\end{tabular}%
    }
    \caption{
        \textbf{Emotion Classification Accuracy for ResidualMaskingNet\protect\citeSup{luanresmaskingnet2020_supRef}:}
        We report the FER image-based classification results for the appearance reconstructions.
    }
    \label{tab:app:rmn}
\end{table}

\begin{table}[h]
    \centering
    \resizebox{\linewidth}{!}{%
    \begin{tabular}{l|rrrrrr|r}
	\toprule
	                & Angry & Disgusted & Fearful & Happy & Sad   & Surprised & Average \\
	\midrule
	Upper Limit (N) & 48.96 & 0.00      & 14.24   & 97.21 & 21.88 & 69.44     & 41.95   \\
	Lower Limit (S) & 19.32 & 0.00      & 30.68   & 61.93 & 31.82 & 34.28     & 29.67   \\
	\midrule
	DECA            & 0.00  & 0.00      & 0.00    & 0.19  & 11.74 & 5.11      & 2.84    \\
	EMOCAv2         & 60.89 & 0.00      & 16.20   & 6.21  & 51.40 & 48.04     & 30.46   \\
	SMIRK           & 70.27 & 0.00      & 17.42   & 60.61 & 13.83 & 9.28      & 28.57   \\
	Deep3DFace      & 1.24  & 0.00      & 3.97    & 65.34 & 4.23  & 66.75     & 23.59   \\
	FOCUS           & 0.00  & 0.00      & 0.00    & 20.52 & 0.26  & 35.49     & 9.38    \\
	MCGAN           & 21.54 & 0.00      & 5.89    & 90.24 & 8.94  & 43.70     & 28.39   \\
	\midrule
	EIFER           & 48.11 & 0.00      & 2.65    & 94.70 & 8.52  & 35.80     & 31.63   \\
	\bottomrule
\end{tabular}%
    }
    \caption{
        \textbf{Emotion Classification Accuracy for EmoNextBase\protect\citeSup{elboudouriEmoNeXtAdaptedConvNeXt2023_supRef}:}
        We report the FER image-based classification results for the appearance reconstructions.
        Please note that the model has never predicted \textit{disgust} for any image.
    }
    \label{tab:app:emonext}
\end{table}

\begin{table}[h]
    \centering
    \resizebox{\linewidth}{!}{%
    \begin{tabular}{l|rrrrrr|r}
	\toprule
	                & Angry & Disgusted & Fearful & Happy & Sad   & Surprised & Average \\
	\midrule
	Upper Limit (N) & 21.53 & 0.00      & 11.46   & 78.75 & 69.10 & 1.74      & 30.43   \\
	Lower Limit (S) & 16.86 & 0.00      & 5.49    & 71.97 & 22.35 & 0.19      & 19.48   \\
	\midrule
	DECA            & 0.19  & 0.00      & 0.19    & 0.38  & 5.30  & 3.98      & 1.67    \\
	EMOCAv2         & 63.69 & 0.00      & 10.06   & 18.08 & 31.28 & 52.51     & 29.27   \\
	SMIRK           & 38.83 & 0.00      & 2.65    & 45.45 & 24.81 & 25.57     & 22.89   \\
	Deep3DFace      & 0.99  & 0.00      & 1.99    & 40.40 & 20.65 & 48.14     & 18.69   \\
	FOCUS           & 1.81  & 0.00      & 0.52    & 31.43 & 1.30  & 40.93     & 12.67   \\
	MCGAN           & 51.42 & 0.00      & 2.85    & 69.11 & 14.84 & 12.20     & 25.07   \\
	\midrule
	EIFER           & 61.74 & 0.00      & 1.70    & 72.54 & 10.04 & 15.72     & 26.96   \\
	\bottomrule
\end{tabular}%
    }
    \caption{
        \textbf{Emotion Classification Accuracy for SegmentationVGG19\protect\citeSup{segmetnationvgg_supRef}:}
        We report the FER image-based classification results for the appearance reconstructions.
        While SegmentationVGG19 performs well on the benchmark datasets, the application to our unseen data results in strong performance degradation.
    }
    \label{tab:app:vgg}
\end{table}

\subsection{Landmarks under Occlusions}
\begin{figure}[h]
    \centering
    \includegraphics[width=\linewidth]{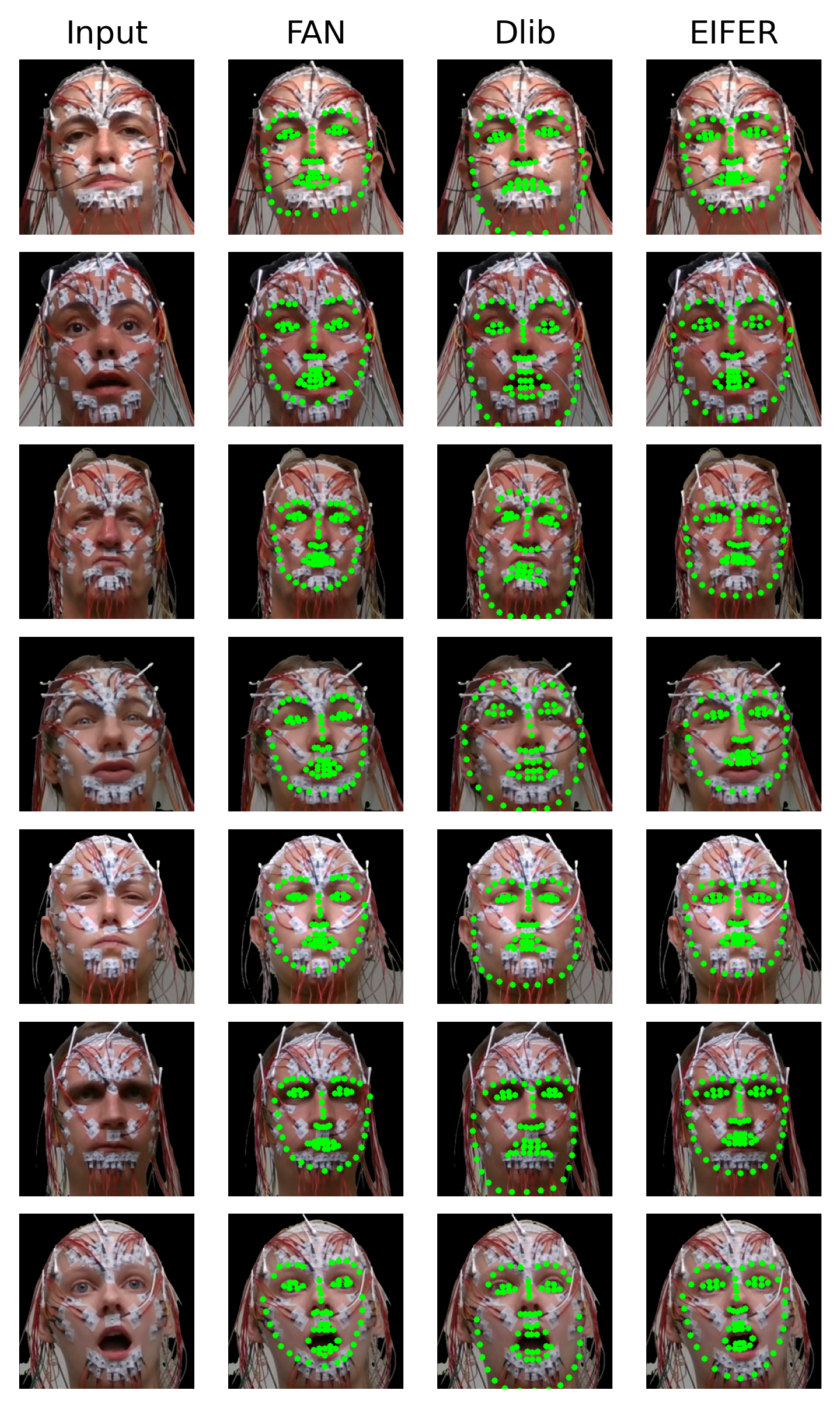}
    \caption{
        \textbf{Landmark Prediction under sEMG occlusion:}
        We see that EIFER can be used to predict the 2D facial landmarks under occlusion, whereas existing methods~\protect\citeSup{liuDenseFaceAlignment2017_supRef, dlib09_supRef} produce inaccurate predictions.
    }
    \label{fig:app:landmarks-new}
\end{figure}

We demonstrate in \Cref{fig:app:landmarks} that existing landmarking models struggle to predict landmarks accurately under sEMG occlusion.
However, EIFER, trained without landmark information, still aligns well with the facial geometry.
We leverage this alignment to predict landmarks, as defined on the FLAME model~\protect\citeSup{FLAME_supRef, deca_supRef, smirk_supRef}.
Although we lack groundtruth annotations for the landmarks, visual inspection reveals that EIFER's predictions outperform those of existing models~\protect\citeSup{bazarevsky2019blazeface_supRef, liuDenseFaceAlignment2017_supRef}, as shown in \Cref{fig:app:landmarks-new}. While EIFER shows improved alignment, there is still room for improvement.

\section{Extended Limitations Discussion}
\textbf{Facial Action Coding System:}
EIFER presents a novel data-driven approach for estimating muscle signals from facial expressions using electromyography.
Comparing this paradigm to traditional methods, such as the Facial Action Coding System (FACS)~\protect\citeSup{ekmanFacialActionCoding1978_paper_supRef}, is an open research direction that we leave for future work.
Existing FACS regression methods do work on our occluded images.
Instead, we rely on MC-CycleGAN recordings~\protect\citeSup{buchner2023let_supRef, buchner2023improved_supRef} to make a fair comparison.
However, as shown in \Cref{sup:sec:fer}, the appearance reconstructions of these models differ from occlusion-free reference recordings.
Further research is necessary to ensure the suitability of our dataset for a comprehensive comparison study.
\newline
\textbf{Generalization:}
Our results' generalizability is uncertain due to the limited sample size ($N=36$).
Additionally, our cohort is based in Germany, which may introduce cultural biases that could impact the results when applied to other populations.
We tested a wide range of standardized facial expressions~\protect\citeSup{schaede_supRef, ekmanArgumentBasicEmotions1992_supRef}, but participants did not perform them voluntarily.
This may affect the generalizability of spontaneous facial expressions, which might exhibit different muscle activity patterns.
However, our results still captured facial mimicry and muscle activity, suggesting that the learned correspondence remains valid.
Our study only includes healthy participants without pre-existing neurological diseases affecting the facial nerve. 
Therefore, conditions like facial palsy or Parkinson's disease may impact the predictions.
EIFER may not address facial asymmetry typical in facial palsy, as it may have learned a symmetry bias from our data~\protect\citeSup{buchner2024power_supRef, buchner2024facing_supRef}.
Furthermore, our models might not recover synkinetic effects (involuntary movements on the contralateral face side).
To address this, we currently record patients with unilateral synkinetic chronic facial palsy to validate our approach for medical use cases.
\newline
\textbf{Data Availability:}
Our dataset was recorded in Germany as part of a medical study, subject to strict data privacy regulations. Due to these regulations, we are restricted in sharing participant data and faces.
However, we will release our trained models, \textit{EMG2Exp} and \textit{Exp2EMG}, which do not contain person-identifiable information, ensuring compliance with data protection regulations~\protect\citeSup{eggerIdentityExpressionAmbiguity3D2021_supRef}.
\newline
\textbf{Disentanglement:}
Our approach relies on the disentanglement of shape and expression in 3D Morphable Models (3DMMs), specifically FLAME~\protect\citeSup{FLAME_supRef} and BaselFaceModel~\protect\citeSup{bfm1_supRef, bfm2_supRef}, as well as the face encoder's ability to establish this correspondence~\protect\citeSup{eggerIdentityExpressionAmbiguity3D2021_supRef, weihererApproximatingIntersectionsDifferences2024_supRef}.
The behavior of other 3DMMs, such as FaceScapes~\protect\citeSup{facescape1_supRef, facescape2_supRef}, ICT-FaceKIT~\protect\citeSup{ictfacekit_supRef}, or FaceWarehouse~\protect\citeSup{faceWarehouse_supRef}, is unclear and requires further investigation.
A necessary condition for exploring these models is the availability of well-pre-trained encoder models.
Without these, the correspondence between facial expressions and muscle activity might not be learnable.

\FloatBarrier
\clearpage
{
    \small
    \bibliographystyleSup{ieeenat_fullname}
    \bibliographySup{supp}
}

\end{document}